%% file: main.tex
\documentclass[10pt,twocolumn,letterpaper]{article}

\usepackage[pagenumbers]{cvpr} % arXiv version (copied from paper/ on 2026-11-24)

\input{preamble}
\usepackage{booktabs}
\usepackage{multirow}
\usepackage[table]{xcolor}
\usepackage{colortbl}
\usepackage{pifont}
\newcommand{\cmk}{\textcolor{green!45!black}{\ding{51}}}
\newcommand{\xmk}{\textcolor{hred}{\ding{55}}}
\newcommand{\pmk}{\textcolor{orange!80!black}{$\triangle$}}
\usepackage{amsmath,amssymb}
\usepackage{enumitem}
\usepackage{longtable}
\usepackage{tikz}
\usepackage{needspace}
\usepackage{tabularx}
\usepackage[skins]{tcolorbox}
\usetikzlibrary{positioning,arrows.meta}
\definecolor{hred}{RGB}{168,26,38}
\definecolor{hblue}{RGB}{26,62,148}
\newcommand{\ours}{VISOR}

\definecolor{cvprblue}{rgb}{0.21,0.49,0.74}
\usepackage[pagebackref,breaklinks,colorlinks,allcolors=cvprblue]{hyperref}

\def\confName{CVPR}
\def\confYear{2027}

\title{\ours{}: A Vision-Language Selector for Multimodal Time-Series Forecasting}

\author{
Seunghan Lee\quad
Jun Seo\quad
Junhyeok Kang\quad
Jaehoon Lee\quad
Sangjun Han\quad
Sungdong Yoo\\
Minjae Kim\quad
Tae Yoon Lim\quad
Dongwan Kang\quad
Hwanil Choi\quad
Soonyoung Lee\quad
Wonbin Ahn\\[4pt]
LG AI Research
}

\newif\ifdraftcolors \draftcolorsfalse
\definecolor{fixedgray}{HTML}{6B6B6B}
\newcommand{\pend}[1]{\ifdraftcolors{\color{black}#1}\else#1\fi}


\makeatletter
\def\@maketitle{
   \newpage
   \null
   \vskip .375in
   \begin{center}
      {\Large \bf \@title \par}
      \vspace*{\arxivTitleGap}{
        \large
        \lineskip .5em
        \begin{tabular}[t]{c}
            \@author
        \end{tabular}
        \par
      }
      \vskip .5em
      \vspace*{\arxivAffilGap}
   \end{center}
}
\makeatother
\newcommand{\arxivTitleGap}{18pt}
\newcommand{\arxivAffilGap}{-12pt}
\begin{document}
\ifdraftcolors\color{fixedgray}\fi
\maketitle
\input{fig/teaser}
\input{sec/0_abstract}
\input{sec/1_intro}
\input{sec/2_related}
\input{sec/3_method}
\input{sec/4_experiments}
\input{sec/5_analysis}
\input{sec/6_conclusion}
% LLM Usage Statement removed for the arXiv version
{
    \small
    \bibliographystyle{ieeenat_fullname}
    \bibliography{main}
}

\clearpage
\appendix
\renewcommand{\textfraction}{0.1} %< appendix only: tables share a column with text (2026-11-17)
\input{sec/X_qa}
\input{sec/X_suppl}

\end{document}

%% file: preamble.tex
\usepackage{cuted} %< for strip environment

\usepackage{currfile} %< for \currfilebase macro

\usepackage{caption} %< for "\captionof" commands (teaser)
\usepackage{placeins} %< appendix float barriers (2026-11-17)

%% file: fig/teaser.tex
\begin{strip}
\centering
\includegraphics[width=0.95\textwidth]{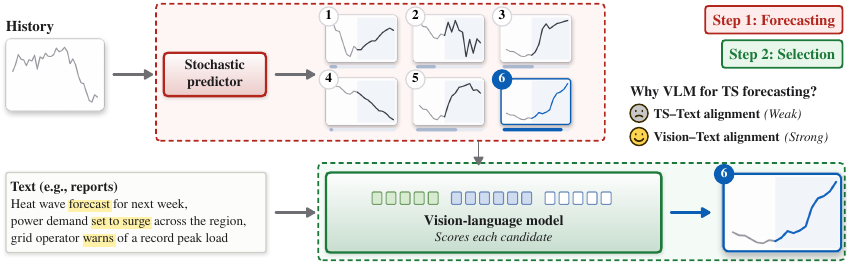}
\captionsetup{hypcap=false}
\captionof{figure}{\textbf{Overview of \ours{}.} \ours{} works in two steps. \textbf{Step 1 (Forecasting):} A predictor samples $K$ diverse candidate futures from the history. \textbf{Step 2 (Selection):} A VLM reads the candidates as numbered charts with the text and selects a single path as the forecast.}
\captionsetup{hypcap=true}
\label{fig:overview}%
\end{strip}

%% file: sec/0_abstract.tex
\begin{abstract}
Recent time-series foundation models (TSFMs) forecast a distribution over futures, and multimodal time-series (TS) forecasting further exploits other modalities to forecast more accurately. However, existing works have two limitations: 1) TSFMs reduce the predictive distribution to its median or mean, which flattens diverse futures into a calm line, and 2) multimodal TS forecasting models feed text into the forecaster through a TS-language alignment that is far weaker than vision-language alignment. To address both limitations, we propose the \textbf{VI}sion-language \textbf{S}elect\textbf{OR} (\textbf{\ours{}}), a multimodal TS forecasting framework that selects one plausible future instead of averaging, by using the alignment of vision-language models (VLMs) to choose among candidates with the text. \ours{} consists of 1) a \textbf{text-free stochastic predictor} that samples diverse candidate futures and 2) a \textbf{VLM selector} that reads the candidates as one chart image and picks one with the text. To avoid calm paths, we train the selector with a shape-aware target that scores each candidate by both its point error and its path shape. \pend{On three benchmarks covering ten datasets, \ours{} ranks first on average in point and volatility errors against TSFMs and multimodal TS forecasting models.} Code: \url{https://github.com/seunghan96/VISOR}.
\end{abstract}

%% file: sec/1_intro.tex
\section{Introduction}
\label{sec:intro}

\begin{figure*}[t]
\centering
\includegraphics[width=\textwidth]{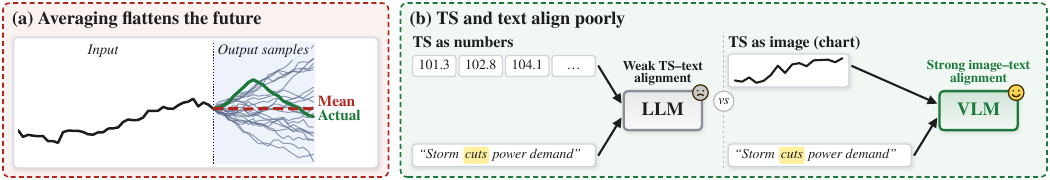}
\caption{\textbf{Two problems that motivate \ours{}.} (a) The mean of stochastic forecast samples is a flat line that \textit{matches neither the actual path nor any single sample}. (b) Feeding a TS as numbers to an LLM needs an alignment between TS and language that is still weak, while rendering it as a chart (TS as image) \textit{reuses the strong image-text alignment} of VLMs.}
\label{fig:motivation}
\end{figure*}

\begin{figure}[t]
\centering
\includegraphics[width=\columnwidth]{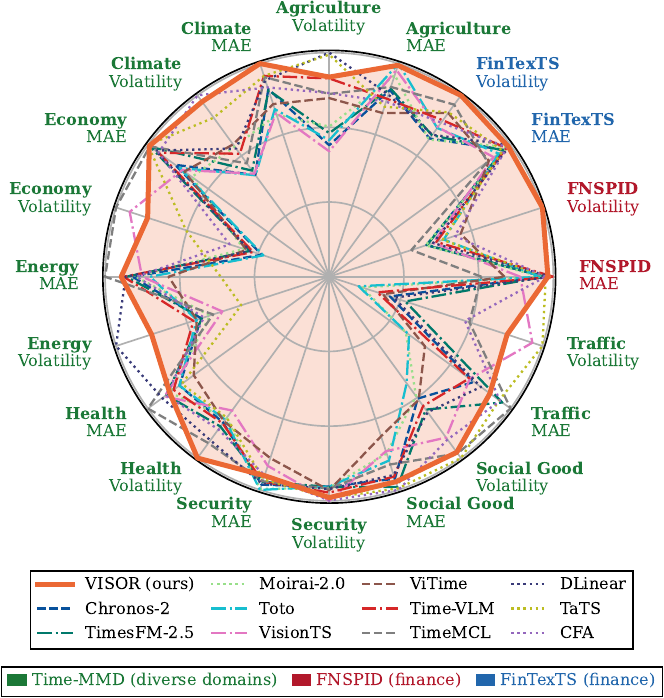}
\caption{\textbf{Comparison with various SoTA methods.} We compare with 1) TSFMs (Chronos-2, TimesFM-2.5, Moirai-2.0, Toto, and the vision-based VisionTS, ViTime), 2) task-specific (DLinear, TimeMCL), and 3) multimodal (Time-VLM, TaTS, CFA) models. Each axis is the best error over that of the method. All baselines are compared in Table~\ref{tab:main_all} and Figure~\ref{fig:rank_box}.}
\label{fig:radar}
\end{figure}

Time-series (TS) forecasting supports decisions in many domains, such as energy \citep{hong2016probabilistic}, demand \citep{lee2026exaone}, and finance \citep{lee2026finstar}. Recent TS foundation models (TSFMs) forecast with a predictive distribution, either by sampling future paths \citep{ansari2024chronos,woo2024moirai} or by predicting quantiles \citep{ansari2025chronos2,liu2025moirai2}. However, when a single forecast is required, the common practice reports the median of the samples or quantiles \citep{ansari2024chronos,woo2024moirai}, and benchmarks score this summary per step \citep{aksu2024gifteval}. When the possible futures branch, their median or mean is a flat line that matches none of them (Figure~\ref{fig:motivation}a), a problem also seen in TS \citep{liu2025sundial} and as blurry outputs in image \citep{isola2017pix2pix,ledig2017srgan} and video prediction \citep{mathieu2016deep}.

We call this effect the \textit{calm-path bias}: a summary of many noisy futures is calmer than the futures themselves, and the bias remains even when the summary is learned. For example, a model trained to pick the sampled future with the lowest point error chooses paths calmer than the most central sample at a cost in volatility (Figure~\ref{fig:calm}). This is not a flaw of a particular model but \textit{a property of the point error itself}, which rewards the mean or the median of the predictive distribution \citep{gneiting2011making}. As a result, a forecast with this bias underestimates the risk of volatile periods.

A second problem is how to bring text into such a forecast. Text such as news or reports often explains why a TS moves, which has made multimodal TS forecasting an active area \citep{liu2024timemmd,williams2025context,lee2026fintexts}. Most methods feed the text into the forecaster itself \citep{jin2024timellm,gruver2023llmtime}, which requires an alignment between TS and language. However, \textit{such alignment is hard to learn}, as shown in previous works: paired data are scarce \citep{xie2025chatts}, removing the language model barely hurts \citep{tan2024language}, and text adds little to strong forecasters \citep{zhang2025multimodality}. In contrast, vision-language models (VLMs) have a \textit{strong alignment between images and text}: they learn from large-scale image-text pairs \citep{radford2021clip,schuhmann2022laion}, answer chart questions \citep{masry2022chartqa,bai2025qwen25vl}, and reason better over a TS plot than over its numbers \citep{daswani2024plots} (Figure~\ref{fig:motivation}b).

To address both problems, we propose the \textbf{VI}sion-language \textbf{S}elect\textbf{OR} (\textbf{\ours{}}), which samples many futures without text and lets a VLM select one with the text (Figure~\ref{fig:overview}). Specifically, a stochastic predictor trained on the TS alone samples diverse candidate futures. A VLM then reads all candidates as one grid of numbered line charts with the text and returns the top-scored one. The text thus \textit{only guides this choice and never edits the predicted values}, following the view that text should enter a TS model in a constrained way \citep{lee2026cfa}. The VLM is fine-tuned to predict a shape-aware distance between each candidate and the truth, which reduces the calm-path bias. Note that we do not argue that images carry more information than numbers, but that text and TS should be combined \textit{where a strong alignment already exists}. Experiments on ten multimodal TS forecasting datasets from three benchmarks \citep{liu2024timemmd,lee2026fintexts,dong2024fnspid} show that \ours{} ranks first on average against TSFMs and multimodal TS forecasting models (Figure~\ref{fig:radar}).

\noindent The main contributions are summarized as follows:
\setlist[itemize]{leftmargin=0.3cm,itemsep=0pt,parsep=0pt,topsep=2pt,partopsep=0pt}
\begin{itemize}
\item We identify \textbf{two problems of multimodal TS forecasting}: 1) averaging or point-error training yields paths calmer than real ones (\textit{calm-path bias}), and 2) feeding text into the forecaster relies on weak TS-text alignment.
\item We propose \textbf{\ours{}}, which addresses both problems by 1) selecting one sampled future with a shape-aware target instead of averaging and 2) using the text only to choose among candidates read as a chart image by a VLM.
\item \pend{We show on ten datasets from three benchmarks that \ours{} 1) ranks first on average in point and volatility errors against TSFMs and multimodal models and 2) stays robust across predictors, VLM sizes, and test periods, and we analyze the roles of the chart, candidates, and text.}
\end{itemize}

%% file: sec/2_related.tex
\section{Related Work}
\label{sec:related}

\input{tables/tab_related}

\noindent\textbf{Probabilistic forecasting and its point summary.}
Probabilistic models and recent TSFMs sample future paths \citep{rasul2021timegrad,ansari2024chronos,woo2024moirai} and are scored with the point error of their median or mean \citep{gneiting2011making}. However, a point-error target smooths away sharp changes. This motivated shape-aware \citep{leguen2019dilate} and direction-aware losses \citep{lee2026direction} and diverse sampling \citep{leguen2020stripe} for TS, while we keep the samples apart and select one.

\noindent\textbf{Selecting one output from many samples.}
Returning the sample with the lowest expected loss against the other samples is known as minimum Bayes risk decoding \citep{kumar2004mbr}, and the medoid of a set of paths is its TS form. Vision models predict several hypotheses in ambiguous tasks \citep{guzmanrivera2012mcl,rupprecht2017learning}, which has been carried to TS forecasting \citep{cortes2025timemcl}. Closest to ours, a recent method lets an LLM rank the numeric continuations of a frozen TSFM given text \citep{nguyen2026lafp}. Different from it, our VLM reads the candidates as one chart image and returns one of them without merging. Its shape-aware target also decides the shape of the chosen path.

\noindent\textbf{Text for multimodal TS forecasting.}
Recent datasets pair text with TS of many domains \citep{liu2024timemmd,williams2025context}, including finance \citep{dong2024fnspid,lee2026fintexts}. Most methods feed the text into the forecaster through prompts \citep{gruver2023llmtime} or reprogrammed language models \citep{jin2024timellm}, or let LLMs reason over TS and text \citep{lee2026reasoncast,lee2026finstar}. However, the alignment between TS and language is weak in these methods: removing the language model keeps their accuracy \citep{tan2024language}, the learned alignment is shallow \citep{zheng2024understanding}, and tokenizers split numbers in ways that break their order and size \citep{spathis2023first}. Furthermore, text helps mostly weak unimodal models and adds little to a strong TSFM \citep{zhang2025multimodality}, which suggests that text should enter a TS model \textit{in a constrained way} \citep{lee2026cfa}. Unlike all of them, our predictor never sees text, and the text only guides the choice.

\noindent\textbf{Vision models for TS and charts.}
TS can be rendered as images for forecasting \citep{chen2024visionts,yang2025vitime}, and multimodal models reason better over plots than over the same TS written as numbers \citep{daswani2024plots,liu2025picture,zhou2025anomalies}. A VLM given a chart and its numbers can also forecast without training \citep{khezresmaeilzadeh2025vista}. VLMs learn a strong image-text alignment from large image-text data \citep{radford2021clip,bai2025qwen25vl}, while language models still struggle to reason about TS \citep{merrill2024struggle}. In contrast, our VLM never outputs numbers and only compares candidate futures plotted on a shared axis. We fine-tune it with low-rank adapters (LoRA) \citep{hu2022lora} on both its language model and its vision tower, as the vision tower has not seen such charts. Table~\ref{tab:related} compares \ours{} with prior work, and Appendix~\ref{app:related} in the supplementary material discusses further related work.

%% file: tables/tab_related.tex
\begin{table}[t]
\centering
\caption{\textbf{Comparison with prior work.} Sampling: samples diverse futures. Single path: returns one sample instead of a median or mean. Shape: trained to match the path shape. \pmk{}: partly.}
\label{tab:related}
\small
\setlength{\tabcolsep}{1.4pt}
\renewcommand{\arraystretch}{0.94}
\begin{tabular}{l|ccc|ccc}
\toprule
\multirow{2.5}{*}{Method} & \multicolumn{3}{c|}{Prediction} & \multicolumn{3}{c}{Multimodality} \\
\cmidrule(lr){2-4}\cmidrule(lr){5-7}
 & Sampling & Single path & Shape & Text & Image & VLM \\
\midrule
\rowcolor{gray!15}\multicolumn{7}{l}{\textit{Probabilistic forecasters}} \\
\midrule
$\vcenter{\hbox{\shortstack[l]{Chronos~\citep{ansari2024chronos}\\Moirai~\citep{woo2024moirai}\\Toto~\citep{cohen2025toto}\\LLMTime~\citep{gruver2023llmtime}}}}$ & \cmk & \xmk & \xmk & \xmk & \xmk & \xmk \\
\cmidrule(lr){1-7}
TimeMCL~\citep{cortes2025timemcl} & \cmk & \pmk & \xmk & \xmk & \xmk & \xmk \\
\midrule
\rowcolor{gray!15}\multicolumn{7}{l}{\textit{Text-guided forecasters}} \\
\midrule
$\vcenter{\hbox{\shortstack[l]{Time-LLM~\citep{jin2024timellm}\\MM-TSFlib~\citep{liu2024timemmd}}}}$ & \xmk & \xmk & \xmk & \cmk & \xmk & \xmk \\
\cmidrule(lr){1-7}
LAFP~\citep{nguyen2026lafp} & \cmk & \pmk & \xmk & \cmk & \xmk & \xmk \\
\midrule
\rowcolor{gray!15}\multicolumn{7}{l}{\textit{Vision-based forecasters}} \\
\midrule
VisionTS~\citep{chen2024visionts} & \xmk & \xmk & \xmk & \xmk & \cmk & \xmk \\
\cmidrule(lr){1-7}
ViTime~\citep{yang2025vitime} & \cmk & \xmk & \xmk & \xmk & \cmk & \xmk \\
\cmidrule(lr){1-7}
VISTA~\citep{khezresmaeilzadeh2025vista} & \xmk & \xmk & \xmk & \xmk & \cmk & \cmk \\
Time-VLM~\citep{zhong2025timevlm} & \xmk & \xmk & \xmk & \cmk & \cmk & \cmk \\
\midrule\midrule
\rowcolor{yellow!18} \ours{} & \cmk & \cmk & \cmk & \cmk & \cmk & \cmk \\
\bottomrule
\end{tabular}
\end{table}

%% file: sec/3_method.tex
\section{The \ours{} Framework}
\label{sec:method}

\ours{} forecasts in two steps (Figure~\ref{fig:overview}): 1) a \textbf{text-free predictor} samples candidates from the history, and 2) a \textbf{VLM selector} reads them as charts with the text and picks one.
% The method is domain-agnostic, and Section~\ref{sec:setup} gives the setting of each dataset. We define the task in Section~\ref{sec:problem}, and describe the predictor in Section~\ref{sec:gen}, the selector in Section~\ref{sec:sel}, and the target in Section~\ref{sec:target}.

\subsection{Problem Setup}
\label{sec:problem}

\noindent\textbf{Multimodal TS forecasting.}
Let $x_{t-L+1:t}$ be the last $L$ steps of a TS up to step $t$, and let $\mathcal{T}_{\le t}$ be the text available up to $t$, such as news or reports with a timestamp. Multimodal TS forecasting predicts the next $H$ steps $x_{t+1:t+H}$ from both inputs. Only the text published up to step $t$ is used, which prevents any leakage of future information.

\noindent\textbf{Path representation.}
TS differ in level and scale, and we compare all forecasts in one path space. We write the future as its change from the last observed value $x_t$, scaled per step, which gives the path $y \in \mathbb{R}^{H}$ with
\begin{equation}
y_h = \frac{x_{t+h} - x_t}{\sigma\sqrt{h}}, \quad h = 1, \dots, H,
\label{eq:path}
\end{equation}
where $\sigma$ is the spread of the one-step changes in the history (Appendix~\ref{app:norm}). This puts TS of different scales on one scale and keeps the spread roughly constant across steps. The truth, every sampled path, the selection target, and every metric live in this space. It fixes how forecasts are compared, not how the predictor produces them.

\noindent\textbf{Probabilistic forecasting.}
A probabilistic forecaster models the distribution $p(y \mid x_{t-L+1:t})$ and draws $K$ sample paths $\hat{y}_1, \dots, \hat{y}_K$ from it. When a single forecast is required, the per-step median or mean of these samples is commonly returned. This summary is not one of the samples and is calmer than the samples (Figure~\ref{fig:motivation}a).

\noindent\textbf{Our task.}
Instead of a summary, we return \textit{one of the candidate paths}, ${\hat{y} = \hat{y}_{k^\star}}$, where $k^\star$ is chosen with the help of $\mathcal{T}_{\le t}$. The predictor never sees the text, and the text only guides the choice of $k^\star$ without changing the sampled values. We evaluate $\hat{y}$ by its point error to $y$ and by how well its shape (\eg, volatility) matches $y$ (Section~\ref{sec:setup}).

\subsection{Step 1: Text-Free Stochastic Predictor}
\label{sec:gen}

\input{tables/tab_designs}
\noindent\textbf{Predictor designs.}
The predictor samples candidate paths from the history alone. Any stochastic forecaster can serve as the predictor, and we consider three designs (Table~\ref{tab:designs}): 1)~\textit{Frozen-TSFM} samples paths directly from a released TSFM without training, 2) \textit{TS-to-Chart} trains a TS model on numbers and draws its sampled paths as charts for the selector, and 3) \textit{Chart-Native} trains an image model that draws the future part of a chart of the history and reads its path from the drawn chart. Each design may read and sample in its own normalized form (Appendix~\ref{app:norm}). Appendix~\ref{app:impl} details the predictor used for each benchmark.

\noindent\textbf{Candidate set.}
For each window, the predictor $G$ of the chosen design draws $K$ candidate paths,
\begin{equation}
\hat{y}_k = G(x_{t-L+1:t}, \epsilon_k), \quad \epsilon_k \sim \mathcal{N}(\mathbf{0}, \mathbf{I}), \quad k = 1, \dots, K.
\label{eq:cand}
\end{equation}
The candidates cover a wide range of futures, and the best candidate in hindsight is far closer to the truth than the medoid, the candidate closest to all others (Appendix~\ref{app:impl}).

\subsection{Step 2: VLM Selector}
\label{sec:sel}

\begin{figure}[t]
\centering
\includegraphics[width=\linewidth]{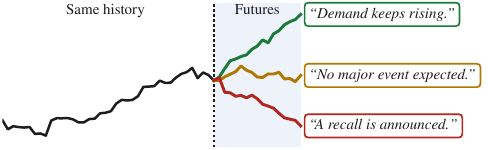}
\caption{\textbf{Same history, different futures.} One history can be followed by a rising, a flat, or a falling future. The text context tells which one is likely, which the history alone cannot.}
\label{fig:toy}
\end{figure}

\noindent\textbf{Why text matters.}
The same history can be followed by very different futures, and the history alone cannot tell them apart. The text often can, as it reports events that the history does not show (Figure~\ref{fig:toy}). The selector thus reads the text to pick one candidate, \textit{while the predictor never sees it}.

\noindent\textbf{Chart and prompt.}
The renderer $\mathcal{R}$ draws each candidate as a small line chart of the recent history followed by the candidate future on a shaded background. All $K$ charts are placed as numbered panels of one grid image $I = \mathcal{R}(x_{t-L+1:t}, \hat{y}_{1:K})$ on a shared value axis, which lets the VLM compare the candidates at a glance. The panel order is shuffled per window to prevent a position bias. The prompt gives the recent items of $\mathcal{T}_{\le t}$, followed by the grid image and one marker token per candidate (Appendix~\ref{app:prompt}).

\noindent\textbf{Scoring head.}
We use Qwen2.5-VL-3B \citep{bai2025qwen25vl} as the backbone and take the last hidden state at each marker token as a candidate state. A two-layer transformer over these $K$ states and the final prompt token scores each candidate,
\begin{equation}
(s_1, \dots, s_K) = f_\theta(I, \mathcal{T}_{\le t}), \quad k^\star = \arg\max_k s_k,
\label{eq:visor}
\end{equation}
where $f_\theta$ denotes the whole selector. The head has no positional encoding, which keeps the head free of any dependence on the panel order. We fine-tune the VLM with LoRA \citep{hu2022lora} on both its language model and its vision tower.

\noindent\textbf{Auxiliary chart reading.}
To help the vision tower read charts, each candidate state also predicts three values of its panel: 1) its centrality among the candidates, 2) its volatility, and 3) its total change. This auxiliary task needs no labels beyond the candidates and is used only in training. Further details are discussed in Appendix~\ref{app:aux}.

\subsection{Shape-Aware Selection Target}
\label{sec:target}

\noindent\textbf{Shape-aware distance.}
The selector regresses the distance of each candidate to the truth. The base distance is the mean absolute error (MAE) of the path, which is the target of a point-error selector. We add a shape term, the same error on four path properties: 1) volatility, 2) largest drop, 3) largest rise, and 4) the range within each step if given. This gives
\begin{equation}
d_k = \mathrm{MAE}(\hat{y}_k, y) + \lambda_{\text{shape}}\, \mathrm{MAE}\big(\tilde{\phi}(\hat{y}_k), \tilde{\phi}(y)\big),
\label{eq:target}
\end{equation}
where $\tilde{\phi}$ stacks the scaled properties (Appendix~\ref{app:target}).

\noindent\textbf{Regression loss.}
We train the scores to match the negative distances up to a constant per window,
\begin{equation}
\mathcal{L} = \frac{1}{K}\sum_{k=1}^{K} (s_k + d_k - \mu)^2, \quad \mu = \frac{1}{K}\sum_{j=1}^{K} (s_j + d_j),
\label{eq:loss}
\end{equation}
where $\mu$ absorbs the constant. The negative score thus estimates the expected distance, and the pick minimizes it.

\noindent\textbf{Why a shape term.}
With the point error alone, the candidate with the lowest expected error is a calm one, as it stays close to the center of the candidates. The shape term makes a calm candidate pay for missing the size of the move. This reduces the calm-path bias \textit{without predicting the direction}.
\makeatletter\typeout{SEC3END page \thepage\space col \if@firstcolumn 1\else 2\fi\space total \the\pagetotal\space goal \the\pagegoal}\makeatother

%% file: tables/tab_designs.tex
\begin{table}[t]
\centering
\caption{\textbf{Predictor designs.} $F$ is a frozen TSFM and $G_\theta$ is trained on each dataset. Chart-Native samples a chart $\hat{I}_k$ of the future, whose path $\hat{y}_k$ is read and redrawn by $\mathcal{R}$ (Section~\ref{sec:sel}).}
\label{tab:designs}
\small
\setlength{\tabcolsep}{2.4pt}
\begin{tabular}{l|l|c}
\toprule
Design & Output of predictor & Input for selector \\
\midrule
Frozen-TSFM & Path $\hat{y}_k = F(x_{t-L+1:t}, \epsilon_k)$ & \multirow{3}{*}{$\mathcal{R}(x_{t-L+1:t}, \hat{y}_k)$} \\
TS-to-Chart & Path $\hat{y}_k = G_\theta(x_{t-L+1:t}, \epsilon_k)$ & \\
Chart-Native & Chart $\hat{I}_k = G_\theta(x_{t-L+1:t}, \epsilon_k)$ & \\
\bottomrule
\end{tabular}
\end{table}

%% file: sec/4_experiments.tex
\section{Experiments}
\label{sec:exp}
\makeatletter\typeout{SEC4PROBE page \thepage\space col \if@firstcolumn 1\else 2\fi\space total \the\pagetotal}\makeatother

\subsection{Experimental Setup}
\label{sec:setup}

\noindent\textbf{Datasets.}
We use three benchmarks that pair TS with dated text: eight datasets of Time-MMD \citep{liu2024timemmd} from domains such as health and traffic, each a separate task, FinTexTS \citep{lee2026fintexts}, which pairs stocks with LLM summaries of news, and FNSPID \citep{dong2024fnspid}, which pairs stocks with news. We follow the protocol of each benchmark, split each dataset in time, and train a predictor and a selector on each dataset. Table~\ref{tab:data} summarizes the splits, and Appendix~\ref{app:data} gives the details.

\input{tables/tab_data}

\noindent\textbf{Metrics.}
MAE and MSE measure the point error, and Endpoint (End.) the error of the total change. Volatility (Vol.) compares the volatility of the chosen path with that of the truth, and FD is the Fr\'echet distance between the feature distributions of the chosen and the true paths \citep{heusel2017fid}, with the features in Appendix~\ref{app:impl}. On FNSPID, whose steps also have an upper and a lower value, HL and Range score these two paths and their gap (Appendix~\ref{app:data}). We assess significance with paired $t$-statistics clustered by time, which accounts for the common move at each step.

\noindent\textbf{Baselines.}
We compare with nine TSFMs, five task-specific and eight multimodal models (Table~\ref{tab:baselines}, Appendix~\ref{app:impl}). All methods forecast from the same history. On Time-MMD, the prior that the benchmark releases for each horizon step is read by the baselines built for it and, on some datasets, by our predictor (Appendix~\ref{app:tmmd}). On our candidates, we compare with a random candidate and the medoid, the candidate closest to the others \citep{kumar2004mbr}. Point forecasts are not paths and serve as references for the point error (Appendix~\ref{app:wtl}).

\input{tables/tab_baselines}

\input{tables/tab_main_all}

\noindent\textbf{Implementation.}
We use Qwen2.5-VL-3B, $K{=}32$ candidates, and $\lambda_{\text{shape}}{=}0.1$ for \ours{}. The selectors of FinTexTS and Time-MMD are initialized from the FNSPID selector. All settings and checkpoints are chosen on validation, and we report three-seed means (Appendix~\ref{app:seeds}).

\begin{figure}[t]
\centering
\includegraphics[width=\columnwidth]{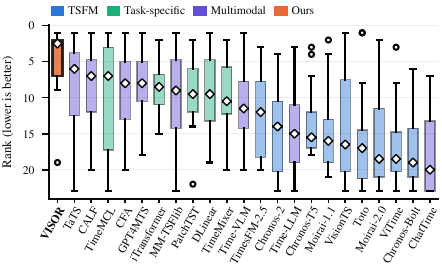}
\caption{\textbf{Ranks across benchmarks.} Rank of each method in each column of Table~\ref{tab:main_all}, sorted by the median rank (diamond).}
\label{fig:rank_box}
\end{figure}

\noindent\textbf{No test leakage.}
\label{sec:leak}
Every dataset is split in time, and every choice of \ours{} is made on validation. The predictor is trained only on pre-test data and never reads text, and the VLM does not recall test outcomes (Appendix~\ref{app:leak}).

\subsection{Main Results}
\label{sec:main}

\noindent\textbf{Results on all benchmarks.}
\pend{We score the default point forecast of each released TSFM, the classic and multimodal models trained on each dataset, and \ours{} on the same rows. As shown in Table~\ref{tab:main_all}, \ours{} beats every TSFM in volatility error on most datasets, and on average also their single sampled or medoid paths (Appendix~\ref{app:single}). As no forecaster is best on every dataset \citep{qiu2024tfb,li2025tsfmbench}, Figure~\ref{fig:rank_box} ranks the methods in each column of Table~\ref{tab:main_all}. \ours{} has the lowest median rank and the best average rank on the MAE columns alone. The strongest baselines lead some columns but trail on others, while \ours{} wins more paired tests than it loses against each (Appendix~\ref{app:wtl}). On stock prices, which move close to a random walk, no method beats the no-change forecast in point error, and the shape of the path separates the methods there (Appendix~\ref{app:wtl}). It also ranks first while training few parameters on TS (Figure~\ref{fig:eff_share}). Table~\ref{tab:main_all} tunes each dataset on validation, and one default setting on every dataset still ranks first on average (Appendix~\ref{app:tmmd}).}

\noindent\textbf{Effect of VLM selection.}
\pend{To ask whether the VLM beats training-free rules, we compare \ours{} with two rules that pick from the same candidates without the VLM, a random candidate and the medoid. In Table~\ref{tab:medoid}, \ours{} beats both in every metric on FNSPID and gives the lowest MAE and endpoint error on all three benchmarks. The VLM thus picks a candidate \textit{closer to the truth}, with significant gains marked by $\dagger$. Our text-free predictor already places its medoid near the top of Table~\ref{tab:main_all}, and the selector lowers the MAE further on every benchmark. The medoid keeps a lower volatility error on Time-MMD, where its candidates spread the least (Appendix~\ref{app:diversity}).}

%% file: tables/tab_data.tex
\begin{table}[t]
\centering
\caption{\textbf{Datasets.} Input length $L$, horizon $H$, paired text, and windows per split. Summ.\ denotes LLM summaries of news.}
\label{tab:data}
\small
\setlength{\tabcolsep}{1.1pt}
\begin{tabular}{l|cccc|ccc}
\toprule
Dataset & Freq. & $L$ & $H$ & Text & Train & Validation & Test \\
\midrule
\multicolumn{8}{l}{\textbf{Time-MMD} \citep{liu2024timemmd}} \\
Agriculture & Month & 24 & 12 & Report & 312 & 39 & 88 \\
Climate & Month & 24 & 12 & Report & 312 & 39 & 88 \\
Economy & Month & 24 & 12 & Report & 261 & 32 & 73 \\
Energy & Week & 24 & 12 & Report & 1{,}000 & 138 & 284 \\
Health & Week & 24 & 12 & Report & 937 & 129 & 266 \\
Security & Month & 24 & 12 & Report & 172 & 20 & 48 \\
Social Good & Month & 24 & 12 & Report & 595 & 79 & 169 \\
Traffic & Month & 24 & 12 & Report & 336 & 43 & 95 \\
\midrule
\textbf{FNSPID} \citep{dong2024fnspid} & Day & 90 & 10 & News & 320{,}812 & 31{,}457 & 10{,}000 \\
\midrule
\textbf{FinTexTS} \citep{lee2026fintexts} & Day & 64 & 3 & Summ. & 71{,}800 & 25{,}800 & 25{,}800 \\
\bottomrule
\end{tabular}
\end{table}

%% file: tables/tab_baselines.tex
\begin{table}[t]
\centering
\caption{\textbf{Baselines.} TSFMs are used as released, apart from the input window of ViTime, with their point forecast. Other models, except the zero-shot ChatTime, are trained on each dataset, with their released pretrained backbones if any (Appendix~\ref{app:impl}).}
\label{tab:baselines}
\small
\setlength{\tabcolsep}{3.5pt}
\begin{tabular}{>{\raggedright\arraybackslash}m{180pt}|l}
\toprule
Models & Output \\
\midrule
\rowcolor{gray!15}\multicolumn{2}{l}{\textit{TSFMs}} \\
\midrule
Chronos-T5 \citep{ansari2024chronos}, Moirai-1.1 \citep{woo2024moirai}, Toto \citep{cohen2025toto}, ViTime \citep{yang2025vitime} & Samples \\
\cmidrule(lr){1-2}
Chronos-Bolt \citep{ansari2024chronos}, Chronos-2 \citep{ansari2025chronos2}, TimesFM-2.5 \citep{das2024timesfm}, Moirai-2.0 \citep{liu2025moirai2} & Quantiles \\
\cmidrule(lr){1-2}
VisionTS \citep{chen2024visionts} & Point \\
\midrule
\rowcolor{gray!15}\multicolumn{2}{l}{\textit{Task-specific TS models}} \\
\midrule
DLinear \citep{zeng2023dlinear}, PatchTST \citep{nie2023patchtst}, iTransformer \citep{liu2024itransformer}, TimeMixer \citep{wang2024timemixer} & Point \\
\cmidrule(lr){1-2}
TimeMCL \citep{cortes2025timemcl} & Hypotheses \\
\midrule
\rowcolor{gray!15}\multicolumn{2}{l}{\textit{Multimodal TS models}} \\
\midrule
MM-TSFlib$^{*}$ \citep{liu2024timemmd}, TaTS$^{*}$ \citep{li2026tats}, CFA$^{*}$ \citep{lee2026cfa}, Time-VLM \citep{zhong2025timevlm}, GPT4MTS \citep{jia2024gpt4mts}, Time-LLM \citep{jin2024timellm}, CALF \citep{liu2025calf}, ChatTime \citep{wang2025chattime} & Point \\
\bottomrule
\end{tabular}
\par\vspace{3pt}
{\footnotesize\raggedright $^{*}$Frameworks that a TS forecaster plugs into. For each of them, we report the best of the five TS backbones we run (Appendix~\ref{app:mm_bb}).\par}
\end{table}

%% file: tables/tab_main_all.tex
\begin{table*}[t]
\centering
\caption{\textbf{Results on multimodal TS benchmarks.} Test MAE and volatility error of each method. On average over the ten datasets, \ours{} ranks first in both metrics and has the best mean rank in MAE (Appendix~\ref{app:qa}), while no method wins every dataset.}
\label{tab:main_all}
\scriptsize
\setlength{\tabcolsep}{1.15pt}
\begin{tabular}{l|cc|cc|cc|cc|cc|cc|cc|cc|cc|cc||cc}
\toprule
\multirow{4}{*}{Method} & \multicolumn{2}{c|}{\multirow{3}{*}{\textbf{FNSPID}~\citep{dong2024fnspid}}} & \multicolumn{2}{c|}{\multirow{3}{*}{\textbf{FinTexTS}~\citep{lee2026fintexts}}} & \multicolumn{16}{c||}{\textbf{Time-MMD}~\citep{liu2024timemmd}} & \multicolumn{2}{c}{\multirow{3}{*}{\textbf{Average}}} \\
\cmidrule(lr){6-21}
 & \multicolumn{2}{c|}{} & \multicolumn{2}{c|}{} & \multicolumn{2}{c|}{Agriculture} & \multicolumn{2}{c|}{Climate} & \multicolumn{2}{c|}{Economy} & \multicolumn{2}{c|}{Energy} & \multicolumn{2}{c|}{Health} & \multicolumn{2}{c|}{Security} & \multicolumn{2}{c|}{Social Good} & \multicolumn{2}{c||}{Traffic} & \multicolumn{2}{c}{} \\
\cmidrule(lr){2-3} \cmidrule(lr){4-5} \cmidrule(lr){6-7} \cmidrule(lr){8-9} \cmidrule(lr){10-11} \cmidrule(lr){12-13} \cmidrule(lr){14-15} \cmidrule(lr){16-17} \cmidrule(lr){18-19} \cmidrule(lr){20-21} \cmidrule(lr){22-23}
 & MAE & Vol. & MAE & Vol. & MAE & Vol. & MAE & Vol. & MAE & Vol. & MAE & Vol. & MAE & Vol. & MAE & Vol. & MAE & Vol. & MAE & Vol. & MAE & Vol. \\
\midrule
\rowcolor{gray!15}\multicolumn{23}{l}{\textit{TSFMs}} \\
Chronos-T5 \citep{ansari2024chronos} & \cellcolor[RGB]{250,223,229}0.851 & \cellcolor[RGB]{254,249,250}0.712 & \cellcolor[RGB]{251,230,234}0.876 & \cellcolor[RGB]{227,246,240}0.516 & \cellcolor[RGB]{181,230,216}1.009 & \cellcolor[RGB]{251,230,234}0.694 & \cellcolor[RGB]{252,236,239}0.336 & \cellcolor[RGB]{250,223,229}0.729 & \cellcolor[RGB]{254,249,250}0.448 & \cellcolor[RGB]{253,242,244}0.732 & \cellcolor[RGB]{248,217,223}2.127 & \cellcolor[RGB]{209,240,230}0.653 & \cellcolor[RGB]{250,223,229}2.082 & \cellcolor[RGB]{227,246,240}1.118 & \cellcolor[RGB]{172,227,211}0.700 & \cellcolor[RGB]{250,223,229}3.739 & \cellcolor[RGB]{252,236,239}0.806 & \cellcolor[RGB]{251,230,234}0.991 & \cellcolor[RGB]{250,223,229}0.309 & \cellcolor[RGB]{248,217,223}0.522 & 0.954\hspace{1pt}{\tiny \#16} & 1.041\hspace{1pt}{\tiny \#16} \\
Chronos-Bolt \citep{ansari2024chronos} & \cellcolor[RGB]{209,240,230}0.796 & \cellcolor[RGB]{245,198,208}0.782 & \cellcolor[RGB]{209,240,230}0.831 & \cellcolor[RGB]{246,204,213}0.599 & \cellcolor[RGB]{199,236,226}1.084 & \cellcolor[RGB]{244,191,202}0.900 & \cellcolor[RGB]{250,223,229}0.337 & \cellcolor[RGB]{246,204,213}0.821 & \cellcolor[RGB]{251,230,234}0.454 & \cellcolor[RGB]{244,191,202}0.934 & \cellcolor[RGB]{255,255,255}1.992 & \cellcolor[RGB]{252,236,239}0.754 & \cellcolor[RGB]{248,217,223}2.090 & \cellcolor[RGB]{246,204,213}1.240 & \cellcolor[RGB]{199,236,226}0.708 & \cellcolor[RGB]{245,198,208}3.779 & \cellcolor[RGB]{245,198,208}0.891 & \cellcolor[RGB]{243,185,197}1.297 & \cellcolor[RGB]{248,217,223}0.395 & \cellcolor[RGB]{244,191,202}0.848 & 0.958\hspace{1pt}{\tiny \#17} & 1.195\hspace{1pt}{\tiny \#23} \\
Chronos-2 \citep{ansari2025chronos2} & \cellcolor[RGB]{255,255,255}0.801 & \cellcolor[RGB]{246,204,213}0.779 & \cellcolor[RGB]{218,243,235}0.832 & \cellcolor[RGB]{245,198,208}0.610 & \cellcolor[RGB]{227,246,240}1.112 & \cellcolor[RGB]{245,198,208}0.890 & \cellcolor[RGB]{253,242,244}0.334 & \cellcolor[RGB]{244,191,202}0.838 & \cellcolor[RGB]{253,242,244}0.452 & \cellcolor[RGB]{245,198,208}0.925 & \cellcolor[RGB]{199,236,226}1.913 & \cellcolor[RGB]{246,252,250}0.724 & \cellcolor[RGB]{251,230,234}2.073 & \cellcolor[RGB]{251,230,234}1.203 & \cellcolor[RGB]{181,230,216}0.702 & \cellcolor[RGB]{243,185,197}3.780 & \cellcolor[RGB]{218,243,235}0.747 & \cellcolor[RGB]{246,204,213}1.085 & \cellcolor[RGB]{254,249,250}0.263 & \cellcolor[RGB]{253,242,244}0.425 & 0.923\hspace{1pt}{\tiny \#9} & 1.126\hspace{1pt}{\tiny \#21} \\
TimesFM-2.5 \citep{das2024timesfm} & \cellcolor[RGB]{236,249,245}0.800 & \cellcolor[RGB]{247,210,218}0.774 & \cellcolor[RGB]{236,249,245}0.835 & \cellcolor[RGB]{247,210,218}0.596 & \cellcolor[RGB]{218,243,235}1.102 & \cellcolor[RGB]{247,210,218}0.815 & \cellcolor[RGB]{252,236,239}0.336 & \cellcolor[RGB]{247,210,218}0.803 & \cellcolor[RGB]{181,230,216}0.390 & \cellcolor[RGB]{248,217,223}0.788 & \cellcolor[RGB]{190,233,221}1.856 & \cellcolor[RGB]{255,255,255}0.729 & \cellcolor[RGB]{218,243,235}1.906 & \cellcolor[RGB]{253,242,244}1.172 & \cellcolor[RGB]{209,240,230}0.712 & \cellcolor[RGB]{246,204,213}3.773 & \cellcolor[RGB]{181,230,216}0.718 & \cellcolor[RGB]{250,223,229}0.995 & \cellcolor[RGB]{172,227,211}0.220 & \cellcolor[RGB]{255,255,255}0.339 & \cellcolor[RGB]{237,203,176}0.888\hspace{1pt}{\tiny \#3} & 1.078\hspace{1pt}{\tiny \#19} \\
Moirai-1.1 \citep{woo2024moirai} & \cellcolor[RGB]{248,217,223}0.866 & \cellcolor[RGB]{162,224,206}0.552 & \cellcolor[RGB]{247,210,218}0.895 & \cellcolor[RGB]{253,242,244}0.555 & \cellcolor[RGB]{254,249,250}1.197 & \cellcolor[RGB]{246,252,250}0.613 & \cellcolor[RGB]{247,210,218}0.369 & \cellcolor[RGB]{251,230,234}0.707 & \cellcolor[RGB]{252,236,239}0.453 & \cellcolor[RGB]{250,223,229}0.786 & \cellcolor[RGB]{246,204,213}2.226 & \cellcolor[RGB]{247,210,218}0.814 & \cellcolor[RGB]{245,198,208}2.158 & \cellcolor[RGB]{181,230,216}1.048 & \cellcolor[RGB]{255,255,255}0.722 & \cellcolor[RGB]{252,236,239}3.713 & \cellcolor[RGB]{251,230,234}0.819 & \cellcolor[RGB]{254,249,250}0.962 & \cellcolor[RGB]{245,198,208}0.418 & \cellcolor[RGB]{246,204,213}0.699 & 1.013\hspace{1pt}{\tiny \#20} & 1.045\hspace{1pt}{\tiny \#17} \\
Moirai-2.0 \citep{liu2025moirai2} & \cellcolor[RGB]{252,236,239}0.812 & \cellcolor[RGB]{244,191,202}0.785 & \cellcolor[RGB]{172,227,211}0.814 & \cellcolor[RGB]{244,191,202}0.615 & \cellcolor[RGB]{190,233,221}1.030 & \cellcolor[RGB]{248,217,223}0.779 & \cellcolor[RGB]{243,185,197}0.387 & \cellcolor[RGB]{255,255,255}0.655 & \cellcolor[RGB]{245,198,208}0.507 & \cellcolor[RGB]{254,249,250}0.731 & \cellcolor[RGB]{227,246,240}1.944 & \cellcolor[RGB]{251,230,234}0.778 & \cellcolor[RGB]{236,249,245}1.947 & \cellcolor[RGB]{245,198,208}1.243 & \cellcolor[RGB]{162,224,206}0.688 & \cellcolor[RGB]{247,210,218}3.757 & \cellcolor[RGB]{247,210,218}0.854 & \cellcolor[RGB]{245,198,208}1.107 & \cellcolor[RGB]{243,185,197}0.497 & \cellcolor[RGB]{245,198,208}0.781 & 0.948\hspace{1pt}{\tiny \#14} & 1.123\hspace{1pt}{\tiny \#20} \\
Toto \citep{cohen2025toto} & \cellcolor[RGB]{251,230,234}0.815 & \cellcolor[RGB]{218,243,235}0.661 & \cellcolor[RGB]{254,249,250}0.857 & \cellcolor[RGB]{250,223,229}0.567 & \cellcolor[RGB]{153,221,201}\textbf{0.974} & \cellcolor[RGB]{246,204,213}0.857 & \cellcolor[RGB]{246,204,213}0.374 & \cellcolor[RGB]{243,185,197}0.840 & \cellcolor[RGB]{250,223,229}0.465 & \cellcolor[RGB]{243,185,197}0.949 & \cellcolor[RGB]{251,230,234}2.099 & \cellcolor[RGB]{227,246,240}0.700 & \cellcolor[RGB]{252,236,239}2.071 & \cellcolor[RGB]{247,210,218}1.227 & \cellcolor[RGB]{153,221,201}\textbf{0.685} & \cellcolor[RGB]{245,198,208}3.779 & \cellcolor[RGB]{250,223,229}0.821 & \cellcolor[RGB]{244,191,202}1.241 & \cellcolor[RGB]{244,191,202}0.477 & \cellcolor[RGB]{243,185,197}0.876 & 0.964\hspace{1pt}{\tiny \#18} & 1.170\hspace{1pt}{\tiny \#22} \\
VisionTS \citep{chen2024visionts} & \cellcolor[RGB]{246,204,213}0.901 & \cellcolor[RGB]{251,230,234}0.728 & \cellcolor[RGB]{250,223,229}0.883 & \cellcolor[RGB]{251,230,234}0.564 & \cellcolor[RGB]{172,227,211}1.004 & \cellcolor[RGB]{243,185,197}0.933 & \cellcolor[RGB]{244,191,202}0.382 & \cellcolor[RGB]{245,198,208}0.827 & \cellcolor[RGB]{248,217,223}0.474 & \cellcolor[RGB]{162,224,206}0.313 & \cellcolor[RGB]{253,242,244}2.010 & \cellcolor[RGB]{246,204,213}0.868 & \cellcolor[RGB]{199,236,226}1.883 & \cellcolor[RGB]{243,185,197}1.312 & \cellcolor[RGB]{245,198,208}0.772 & \cellcolor[RGB]{153,221,201}\textbf{3.538} & \cellcolor[RGB]{246,204,213}0.865 & \cellcolor[RGB]{218,243,235}0.822 & \cellcolor[RGB]{253,242,244}0.271 & \cellcolor[RGB]{162,224,206}0.127 & 0.945\hspace{1pt}{\tiny \#13} & 1.003\hspace{1pt}{\tiny \#12} \\
ViTime \citep{yang2025vitime} & \cellcolor[RGB]{245,198,208}0.982 & \cellcolor[RGB]{172,227,211}0.576 & \cellcolor[RGB]{246,204,213}0.910 & \cellcolor[RGB]{218,243,235}0.515 & \cellcolor[RGB]{247,210,218}1.268 & \cellcolor[RGB]{252,236,239}0.655 & \cellcolor[RGB]{248,217,223}0.363 & \cellcolor[RGB]{246,252,250}0.644 & \cellcolor[RGB]{247,210,218}0.481 & \cellcolor[RGB]{251,230,234}0.756 & \cellcolor[RGB]{245,198,208}2.343 & \cellcolor[RGB]{227,246,240}0.700 & \cellcolor[RGB]{244,191,202}2.291 & \cellcolor[RGB]{250,223,229}1.213 & \cellcolor[RGB]{243,185,197}0.805 & \cellcolor[RGB]{253,242,244}3.712 & \cellcolor[RGB]{244,191,202}0.946 & \cellcolor[RGB]{247,210,218}1.074 & \cellcolor[RGB]{246,204,213}0.398 & \cellcolor[RGB]{250,223,229}0.480 & 1.079\hspace{1pt}{\tiny \#22} & 1.033\hspace{1pt}{\tiny \#15} \\
\midrule
\rowcolor{gray!15}\multicolumn{23}{l}{\textit{Task-specific TS models}} \\
DLinear \citep{zeng2023dlinear} & \cellcolor[RGB]{162,224,206}0.778 & \cellcolor[RGB]{250,223,229}0.729 & \cellcolor[RGB]{153,221,201}\textbf{0.801} & \cellcolor[RGB]{248,217,223}0.594 & \cellcolor[RGB]{236,249,245}1.140 & \cellcolor[RGB]{190,233,221}0.522 & \cellcolor[RGB]{199,236,226}0.319 & \cellcolor[RGB]{254,249,250}0.665 & \cellcolor[RGB]{227,246,240}0.401 & \cellcolor[RGB]{247,210,218}0.791 & \cellcolor[RGB]{181,230,216}1.848 & \cellcolor[RGB]{153,221,201}\textbf{0.435} & \cellcolor[RGB]{172,227,211}1.859 & \cellcolor[RGB]{199,236,226}1.073 & \cellcolor[RGB]{190,233,221}0.705 & \cellcolor[RGB]{251,230,234}3.735 & \cellcolor[RGB]{246,252,250}0.761 & \cellcolor[RGB]{253,242,244}0.976 & \cellcolor[RGB]{255,255,255}0.251 & \cellcolor[RGB]{254,249,250}0.399 & \cellcolor[RGB]{208,214,222}0.886\hspace{1pt}{\tiny \#2} & 0.992\hspace{1pt}{\tiny \#11} \\
PatchTST \citep{nie2023patchtst} & \cellcolor[RGB]{181,230,216}0.786 & \cellcolor[RGB]{236,249,245}0.685 & \cellcolor[RGB]{255,255,255}0.853 & \cellcolor[RGB]{236,249,245}0.523 & \cellcolor[RGB]{244,191,202}1.343 & \cellcolor[RGB]{181,230,216}0.521 & \cellcolor[RGB]{218,243,235}0.320 & \cellcolor[RGB]{227,246,240}0.617 & \cellcolor[RGB]{199,236,226}0.396 & \cellcolor[RGB]{236,249,245}0.679 & \cellcolor[RGB]{218,243,235}1.942 & \cellcolor[RGB]{181,230,216}0.572 & \cellcolor[RGB]{254,249,250}2.002 & \cellcolor[RGB]{255,255,255}1.146 & \cellcolor[RGB]{253,242,244}0.723 & \cellcolor[RGB]{227,246,240}3.597 & \cellcolor[RGB]{255,255,255}0.767 & \cellcolor[RGB]{199,236,226}0.809 & \cellcolor[RGB]{162,224,206}0.218 & \cellcolor[RGB]{236,249,245}0.204 & 0.935\hspace{1pt}{\tiny \#12} & 0.935\hspace{1pt}{\tiny \#5} \\
iTransformer \citep{liu2024itransformer} & \cellcolor[RGB]{253,242,244}0.810 & \cellcolor[RGB]{246,252,250}0.701 & \cellcolor[RGB]{253,242,244}0.861 & \cellcolor[RGB]{162,224,206}0.489 & \cellcolor[RGB]{252,236,239}1.216 & \cellcolor[RGB]{209,240,230}0.548 & \cellcolor[RGB]{218,243,235}0.320 & \cellcolor[RGB]{209,240,230}0.569 & \cellcolor[RGB]{218,243,235}0.399 & \cellcolor[RGB]{209,240,230}0.613 & \cellcolor[RGB]{252,236,239}2.065 & \cellcolor[RGB]{181,230,216}0.572 & \cellcolor[RGB]{246,252,250}1.958 & \cellcolor[RGB]{246,252,250}1.125 & \cellcolor[RGB]{236,249,245}0.717 & \cellcolor[RGB]{181,230,216}3.573 & \cellcolor[RGB]{190,233,221}0.722 & \cellcolor[RGB]{227,246,240}0.852 & \cellcolor[RGB]{227,246,240}0.237 & \cellcolor[RGB]{199,236,226}0.154 & 0.931\hspace{1pt}{\tiny \#10} & \cellcolor[RGB]{237,203,176}0.920\hspace{1pt}{\tiny \#3} \\
TimeMixer \citep{wang2024timemixer} & \cellcolor[RGB]{236,249,245}0.800 & \cellcolor[RGB]{181,230,216}0.588 & \cellcolor[RGB]{199,236,226}0.822 & \cellcolor[RGB]{246,252,250}0.527 & \cellcolor[RGB]{248,217,223}1.238 & \cellcolor[RGB]{162,224,206}0.414 & \cellcolor[RGB]{218,243,235}0.320 & \cellcolor[RGB]{236,249,245}0.628 & \cellcolor[RGB]{246,204,213}0.489 & \cellcolor[RGB]{190,233,221}0.477 & \cellcolor[RGB]{246,252,250}1.974 & \cellcolor[RGB]{172,227,211}0.550 & \cellcolor[RGB]{247,210,218}2.100 & \cellcolor[RGB]{236,249,245}1.120 & \cellcolor[RGB]{250,223,229}0.728 & \cellcolor[RGB]{190,233,221}3.576 & \cellcolor[RGB]{254,249,250}0.781 & \cellcolor[RGB]{255,255,255}0.941 & \cellcolor[RGB]{246,252,250}0.250 & \cellcolor[RGB]{246,252,250}0.211 & 0.950\hspace{1pt}{\tiny \#15} & \cellcolor[RGB]{208,214,222}0.903\hspace{1pt}{\tiny \#2} \\
TimeMCL \citep{cortes2025timemcl} & \cellcolor[RGB]{244,191,202}1.131 & \cellcolor[RGB]{243,185,197}0.921 & \cellcolor[RGB]{245,198,208}0.918 & \cellcolor[RGB]{172,227,211}0.490 & \cellcolor[RGB]{209,240,230}1.094 & \cellcolor[RGB]{253,242,244}0.640 & \cellcolor[RGB]{181,230,216}0.314 & \cellcolor[RGB]{248,217,223}0.752 & \cellcolor[RGB]{172,227,211}0.388 & \cellcolor[RGB]{153,221,201}\textbf{0.293} & \cellcolor[RGB]{153,221,201}\textbf{1.679} & \cellcolor[RGB]{250,223,229}0.799 & \cellcolor[RGB]{153,221,201}\textbf{1.712} & \cellcolor[RGB]{190,233,221}1.055 & \cellcolor[RGB]{227,246,240}0.714 & \cellcolor[RGB]{248,217,223}3.742 & \cellcolor[RGB]{253,242,244}0.804 & \cellcolor[RGB]{172,227,211}0.748 & \cellcolor[RGB]{153,221,201}\textbf{0.210} & \cellcolor[RGB]{209,240,230}0.184 & 0.896\hspace{1pt}{\tiny \#4} & 0.962\hspace{1pt}{\tiny \#10} \\
\midrule
\rowcolor{gray!15}\multicolumn{23}{l}{\textit{Multimodal TS models}} \\
MM-TSFlib \citep{liu2024timemmd} & \cellcolor[RGB]{254,249,250}0.802 & \cellcolor[RGB]{253,242,244}0.714 & \cellcolor[RGB]{248,217,223}0.890 & \cellcolor[RGB]{252,236,239}0.556 & \cellcolor[RGB]{253,242,244}1.200 & \cellcolor[RGB]{153,221,201}\textbf{0.380} & \cellcolor[RGB]{218,243,235}0.320 & \cellcolor[RGB]{190,233,221}0.537 & \cellcolor[RGB]{255,255,255}0.416 & \cellcolor[RGB]{199,236,226}0.517 & \cellcolor[RGB]{250,223,229}2.114 & \cellcolor[RGB]{243,185,197}1.128 & \cellcolor[RGB]{181,230,216}1.867 & \cellcolor[RGB]{162,224,206}1.002 & \cellcolor[RGB]{251,230,234}0.726 & \cellcolor[RGB]{172,227,211}3.570 & \cellcolor[RGB]{227,246,240}0.750 & \cellcolor[RGB]{181,230,216}0.750 & \cellcolor[RGB]{209,240,230}0.226 & \cellcolor[RGB]{227,246,240}0.192 & 0.931\hspace{1pt}{\tiny \#10} & 0.935\hspace{1pt}{\tiny \#5} \\
TaTS \citep{li2026tats} & \cellcolor[RGB]{199,236,226}0.795 & \cellcolor[RGB]{227,246,240}0.667 & \cellcolor[RGB]{246,252,250}0.837 & \cellcolor[RGB]{209,240,230}0.513 & \cellcolor[RGB]{250,223,229}1.221 & \cellcolor[RGB]{199,236,226}0.527 & \cellcolor[RGB]{162,224,206}0.312 & \cellcolor[RGB]{181,230,216}0.536 & \cellcolor[RGB]{153,221,201}\textbf{0.384} & \cellcolor[RGB]{181,230,216}0.436 & \cellcolor[RGB]{243,185,197}3.467 & \cellcolor[RGB]{244,191,202}1.054 & \cellcolor[RGB]{246,204,213}2.106 & \cellcolor[RGB]{244,191,202}1.245 & \cellcolor[RGB]{246,252,250}0.721 & \cellcolor[RGB]{199,236,226}3.584 & \cellcolor[RGB]{172,227,211}0.713 & \cellcolor[RGB]{162,224,206}0.729 & \cellcolor[RGB]{199,236,226}0.224 & \cellcolor[RGB]{153,221,201}\textbf{0.121} & 1.078\hspace{1pt}{\tiny \#21} & 0.941\hspace{1pt}{\tiny \#8} \\
CFA \citep{lee2026cfa} & \cellcolor[RGB]{209,240,230}0.796 & \cellcolor[RGB]{190,233,221}0.602 & \cellcolor[RGB]{181,230,216}0.817 & \cellcolor[RGB]{254,249,250}0.547 & \cellcolor[RGB]{255,255,255}1.186 & \cellcolor[RGB]{254,249,250}0.637 & \cellcolor[RGB]{254,249,250}0.331 & \cellcolor[RGB]{153,221,201}\textbf{0.470} & \cellcolor[RGB]{236,249,245}0.408 & \cellcolor[RGB]{218,243,235}0.627 & \cellcolor[RGB]{236,249,245}1.951 & \cellcolor[RGB]{254,249,250}0.740 & \cellcolor[RGB]{209,240,230}1.903 & \cellcolor[RGB]{254,249,250}1.150 & \cellcolor[RGB]{246,204,213}0.735 & \cellcolor[RGB]{162,224,206}3.545 & \cellcolor[RGB]{162,224,206}0.708 & \cellcolor[RGB]{209,240,230}0.812 & \cellcolor[RGB]{190,233,221}0.223 & \cellcolor[RGB]{218,243,235}0.190 & 0.906\hspace{1pt}{\tiny \#7} & 0.932\hspace{1pt}{\tiny \#4} \\
ChatTime \citep{wang2025chattime} & \cellcolor[RGB]{243,185,197}1.247 & \cellcolor[RGB]{252,236,239}0.725 & \cellcolor[RGB]{243,185,197}1.423 & \cellcolor[RGB]{243,185,197}0.627 & \cellcolor[RGB]{243,185,197}1.952 & \cellcolor[RGB]{250,223,229}0.706 & \cellcolor[RGB]{245,198,208}0.378 & \cellcolor[RGB]{218,243,235}0.612 & \cellcolor[RGB]{244,191,202}0.513 & \cellcolor[RGB]{246,252,250}0.698 & \cellcolor[RGB]{244,191,202}2.887 & \cellcolor[RGB]{253,242,244}0.752 & \cellcolor[RGB]{243,185,197}2.535 & \cellcolor[RGB]{209,240,230}1.088 & \cellcolor[RGB]{244,191,202}0.787 & \cellcolor[RGB]{246,252,250}3.658 & \cellcolor[RGB]{243,185,197}1.060 & \cellcolor[RGB]{246,252,250}0.935 & \cellcolor[RGB]{247,210,218}0.396 & \cellcolor[RGB]{251,230,234}0.469 & 1.318\hspace{1pt}{\tiny \#23} & 1.027\hspace{1pt}{\tiny \#14} \\
Time-VLM \citep{zhong2025timevlm} & \cellcolor[RGB]{153,221,201}\textbf{0.768} & \cellcolor[RGB]{255,255,255}0.706 & \cellcolor[RGB]{190,233,221}0.820 & \cellcolor[RGB]{255,255,255}0.539 & \cellcolor[RGB]{246,252,250}1.167 & \cellcolor[RGB]{227,246,240}0.590 & \cellcolor[RGB]{162,224,206}0.312 & \cellcolor[RGB]{253,242,244}0.693 & \cellcolor[RGB]{236,249,245}0.408 & \cellcolor[RGB]{246,204,213}0.822 & \cellcolor[RGB]{172,227,211}1.830 & \cellcolor[RGB]{218,243,235}0.683 & \cellcolor[RGB]{255,255,255}1.995 & \cellcolor[RGB]{252,236,239}1.197 & \cellcolor[RGB]{209,240,230}0.712 & \cellcolor[RGB]{254,249,250}3.674 & \cellcolor[RGB]{236,249,245}0.755 & \cellcolor[RGB]{248,217,223}1.023 & \cellcolor[RGB]{252,236,239}0.272 & \cellcolor[RGB]{247,210,218}0.537 & 0.904\hspace{1pt}{\tiny \#6} & 1.046\hspace{1pt}{\tiny \#18} \\
GPT4MTS \citep{jia2024gpt4mts} & \cellcolor[RGB]{227,246,240}0.799 & \cellcolor[RGB]{199,236,226}0.635 & \cellcolor[RGB]{227,246,240}0.834 & \cellcolor[RGB]{190,233,221}0.495 & \cellcolor[RGB]{251,230,234}1.218 & \cellcolor[RGB]{236,249,245}0.604 & \cellcolor[RGB]{255,255,255}0.328 & \cellcolor[RGB]{172,227,211}0.489 & \cellcolor[RGB]{199,236,226}0.396 & \cellcolor[RGB]{227,246,240}0.678 & \cellcolor[RGB]{209,240,230}1.927 & \cellcolor[RGB]{248,217,223}0.806 & \cellcolor[RGB]{190,233,221}1.881 & \cellcolor[RGB]{248,217,223}1.219 & \cellcolor[RGB]{248,217,223}0.731 & \cellcolor[RGB]{236,249,245}3.601 & \cellcolor[RGB]{190,233,221}0.722 & \cellcolor[RGB]{153,221,201}\textbf{0.712} & \cellcolor[RGB]{181,230,216}0.221 & \cellcolor[RGB]{172,227,211}0.142 & 0.906\hspace{1pt}{\tiny \#7} & 0.938\hspace{1pt}{\tiny \#7} \\
Time-LLM \citep{jin2024timellm} & \cellcolor[RGB]{247,210,218}0.894 & \cellcolor[RGB]{199,236,226}0.635 & \cellcolor[RGB]{244,191,202}0.941 & \cellcolor[RGB]{190,233,221}0.495 & \cellcolor[RGB]{245,198,208}1.275 & \cellcolor[RGB]{172,227,211}0.474 & \cellcolor[RGB]{199,236,226}0.319 & \cellcolor[RGB]{252,236,239}0.705 & \cellcolor[RGB]{243,185,197}0.529 & \cellcolor[RGB]{252,236,239}0.749 & \cellcolor[RGB]{247,210,218}2.139 & \cellcolor[RGB]{245,198,208}0.878 & \cellcolor[RGB]{253,242,244}2.003 & \cellcolor[RGB]{218,243,235}1.110 & \cellcolor[RGB]{255,255,255}0.722 & \cellcolor[RGB]{255,255,255}3.665 & \cellcolor[RGB]{248,217,223}0.845 & \cellcolor[RGB]{252,236,239}0.979 & \cellcolor[RGB]{251,230,234}0.280 & \cellcolor[RGB]{252,236,239}0.444 & 0.995\hspace{1pt}{\tiny \#19} & 1.013\hspace{1pt}{\tiny \#13} \\
CALF \citep{liu2025calf} & \cellcolor[RGB]{172,227,211}0.780 & \cellcolor[RGB]{248,217,223}0.750 & \cellcolor[RGB]{252,236,239}0.868 & \cellcolor[RGB]{181,230,216}0.494 & \cellcolor[RGB]{246,204,213}1.272 & \cellcolor[RGB]{255,255,255}0.632 & \cellcolor[RGB]{190,233,221}0.318 & \cellcolor[RGB]{199,236,226}0.557 & \cellcolor[RGB]{190,233,221}0.391 & \cellcolor[RGB]{255,255,255}0.721 & \cellcolor[RGB]{255,255,255}1.992 & \cellcolor[RGB]{199,236,226}0.609 & \cellcolor[RGB]{162,224,206}1.732 & \cellcolor[RGB]{172,227,211}1.013 & \cellcolor[RGB]{252,236,239}0.724 & \cellcolor[RGB]{218,243,235}3.591 & \cellcolor[RGB]{153,221,201}\textbf{0.668} & \cellcolor[RGB]{236,249,245}0.929 & \cellcolor[RGB]{218,243,235}0.233 & \cellcolor[RGB]{190,233,221}0.149 & 0.898\hspace{1pt}{\tiny \#5} & 0.945\hspace{1pt}{\tiny \#9} \\
\midrule\midrule
\ours{} & \cellcolor[RGB]{190,233,221}0.787 & \cellcolor[RGB]{153,221,201}\textbf{0.355} & \cellcolor[RGB]{162,224,206}0.811 & \cellcolor[RGB]{153,221,201}\textbf{0.465} & \cellcolor[RGB]{162,224,206}0.982 & \cellcolor[RGB]{218,243,235}0.586 & \cellcolor[RGB]{153,221,201}\textbf{0.294} & \cellcolor[RGB]{162,224,206}0.487 & \cellcolor[RGB]{162,224,206}0.387 & \cellcolor[RGB]{172,227,211}0.344 & \cellcolor[RGB]{162,224,206}1.809 & \cellcolor[RGB]{162,224,206}0.522 & \cellcolor[RGB]{227,246,240}1.946 & \cellcolor[RGB]{153,221,201}\textbf{0.968} & \cellcolor[RGB]{247,210,218}0.734 & \cellcolor[RGB]{209,240,230}3.590 & \cellcolor[RGB]{209,240,230}0.736 & \cellcolor[RGB]{190,233,221}0.753 & \cellcolor[RGB]{227,246,240}0.237 & \cellcolor[RGB]{172,227,211}0.142 & \cellcolor[RGB]{250,222,137}\textbf{0.872\hspace{1pt}{\tiny \#1}} & \cellcolor[RGB]{250,222,137}\textbf{0.821\hspace{1pt}{\tiny \#1}} \\
\bottomrule
\end{tabular}
\end{table*}

%% file: sec/5_analysis.tex
\section{Analysis}
\makeatletter\typeout{SEC5PROBE page \thepage\space col \if@firstcolumn 1\else 2\fi}\makeatother
\label{sec:analysis}

\subsection{Selection vs.\ Averaging}
\label{sec:calm}

% Table 6 sits right before Table 7 and Figure 6 right before Figure 7, so that Section 5 starts in the left column of page 6 and Figure 6 lands in the left column of page 7 (user 2026-11-17)

\noindent\textbf{Effect of the selection target.}
\pend{To validate that the calm-path bias comes from the point-error target, we train the same VLM with four targets that differ only in the terms of Eq.~\eqref{eq:target} (Appendix~\ref{app:target}) and average the change over the MAE-only target on all ten datasets. As shown in Table~\ref{tab:objective_all}, the shape term is the only one that lowers the error in every metric, with the largest gains in volatility and FD. The endpoint term, like the point error, rewards paths ending near the center of the candidates, and adding it partly undoes the shift toward realistic movement. This indicates that the bias is \textit{a property of the target} rather than of the model.}

\input{tables/tab_medoid} %< after Figure 6 (cited first), both after their paragraph so that they share the right column and Section 5 starts in the left (user 2026-11-17)
\input{tables/tab_objective_all}

\begin{figure}[t]
\centering
\includegraphics[width=\columnwidth]{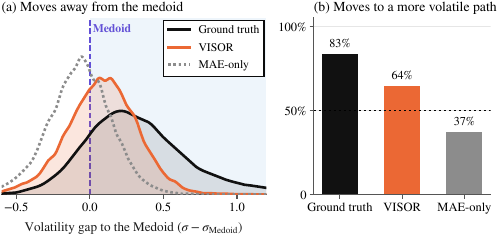}
\caption{\textbf{Calm-path bias.} (a) Volatility gap of the chosen path to the medoid, for the truth and each selector on rows where it leaves the medoid. (b) Share of these rows more volatile than the medoid.}
\label{fig:calm}
\end{figure}

\noindent\textbf{Calm-path bias.}
A single path cannot have both the lowest point error and a realistic shape. The point error is minimized by the median of each step, which is calm by construction, while any real path moves. To see this bias on FNSPID, the benchmark with the most training rows, we compare the volatility of each chosen path with that of the medoid. As shown in Figure~\ref{fig:calm}, the true path is more volatile than the medoid on most rows. When the MAE-only selector leaves the medoid, it mostly moves to a calmer path, away from the truth. \ours{} mostly moves to a more volatile path, toward the truth (Figure~\ref{fig:calm}b). We thus compare \ours{} mainly with the medoid, which is also a single path, and report point forecasts only as references.

\begin{figure}[t]
\centering
\includegraphics[width=\linewidth]{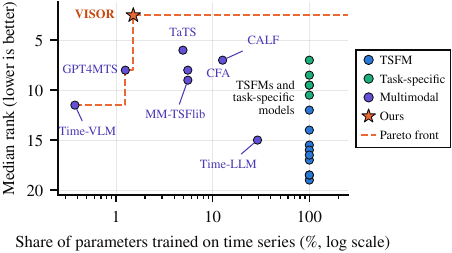}
\caption{\textbf{Training efficiency vs.\ rank.} Frozen backbones are not trained on TS. ChatTime is left out, its share unreported.}
\label{fig:eff_share}
\end{figure}

\begin{figure}[t]
\centering
\includegraphics[width=\columnwidth]{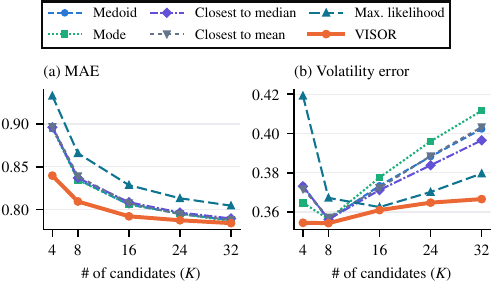}
\caption{\textbf{Number of candidates.} Retrained \ours{} (Appendix~\ref{app:more}) and five rules pick from the first $K$ paths of the two FNSPID candidate sets of Appendix~\ref{app:impl}.}
\label{fig:numk}
\end{figure}

\noindent\textbf{Number of candidates.}
\label{sec:numk}
To ask how the selector depends on the size of the candidate set, we evaluate the selector retrained on more years (Appendix~\ref{app:more}) with the first 4, 8, 16, 24, and 32 candidates of each FNSPID test row and compare it with five training-free rules that pick one of the same candidates (Appendix~\ref{app:impl}). As shown in Figure~\ref{fig:numk}, \ours{} has the lowest MAE and volatility error at every $K$. Its gain in MAE is largest with few candidates, while the rules drift toward calmer central paths as $K$ grows and their volatility error rises. \ours{} keeps a nearly constant volatility error, and its choice \textit{does not flatten} as $K$ grows.

% Figure 8 is placed here so that it lands at the top of the page that holds its paragraph (Section 5.3)
\begin{figure*}[t]
\centering
\includegraphics[width=\textwidth]{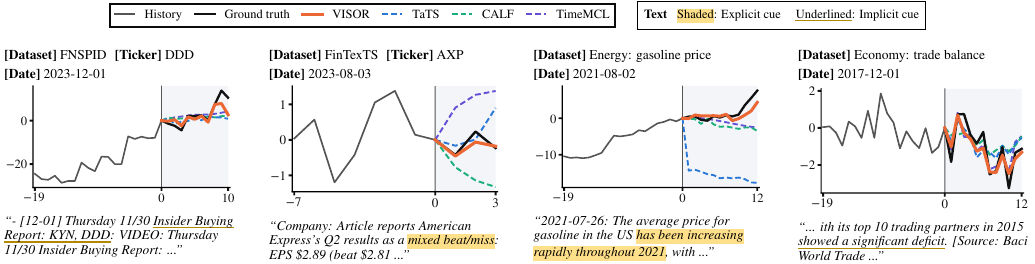}
\caption{\textbf{Selected test forecasts with their text.} Rows where \ours{} picks a path close to the truth, with the forecasts of three baselines and a quote from the text the selector saw. Paths are in units of the recent step size (\% change on FNSPID).}
\label{fig:examples}
\end{figure*}

\subsection{Chart vs.\ Number Input}
\label{sec:why_chart}

\noindent\textbf{Chart input vs.\ number input.}
\pend{To ask whether the chart matters, we train the same selector with the history and candidates written as numbers in the prompt. As shown in Table~\ref{tab:chart_num}, on the full FNSPID test set the chart selector matches the number selector in MAE and MSE and gives a lower volatility, range, and FD error. The gains in volatility and range are significant in a paired test, and both inputs also lower the volatility error of the medoid. Thus, on FNSPID, the VLM reads candidate shapes \textit{better from a chart}.}

\input{tables/tab_chart_num}

\noindent\textbf{Ablation of the chart reader.}
\pend{To validate that the selector reads the charts, we remove one chart-reading component at a time and rank the three settings on the test set of each dataset. As shown in Table~\ref{tab:ablation}, the full \ours{} has the best average rank in every metric. Removing the LoRA on the vision tower gives the worst rank in MAE, MSE, and volatility, indicating that the vision tower must adapt to these charts. Removing the chart auxiliary loss also worsens the rank in every metric, and thus every chart-reading part helps the choice of the selector.}

\input{tables/tab_ablation}

\subsection{Qualitative Examples}
\label{sec:news}

\pend{To show what a good pick looks like, we select test rows on which \ours{} picks a path close to the truth and read the text available before the forecast. As shown in Figure~\ref{fig:examples}, the text in these rows is consistent with the chosen path, and the chosen path agrees with the cue whether it is explicit, such as mixed earnings before a flat stock, or implicit, such as insider buying before a rising stock.}

\subsection{Efficiency Analysis}
\label{sec:robust}

Figure~\ref{fig:eff_share} places each method by the share of its parameters trained on TS and its median rank in Table~\ref{tab:main_all}. \ours{} keeps the pretrained weights of the VLM frozen and trains only the LoRA adapters, the head, and the predictor, a smaller share than most multimodal models and far below the TSFMs and task-specific models. It is the only method on the upper step of the Pareto front, and no method that trains a smaller share ranks as high. One forecast of \ours{} takes under two seconds on one GPU, longer than the TSFMs and about as long as the zero-shot LLM forecaster (Appendix~\ref{app:eff}), small against daily or weekly intervals.

\subsection{Other Analyses}
\label{sec:other}

\noindent The appendix reports further analyses, summarized below.
\begin{itemize}
\item \textbf{Selectors without a VLM.} Gradient boosting on path features has a higher MAE (Appendix~\ref{app:nonvlm}).
\item \textbf{Zero-shot LLM ranker.} An LLM ranker without training has a higher MAE on every dataset (Appendix~\ref{app:nonvlm}).
\item \textbf{Single paths of TSFMs.} Sampled paths and medoids of TSFMs have a larger MAE on average (Appendix~\ref{app:single}).
\item \textbf{Proper score.} \ours{} has the lowest average variogram score of all methods, a proper path score (Appendix~\ref{app:wtl}).
\item \textbf{Scale-free error.} Divided by the no-change error, \ours{} has the lowest mean MAE (Appendix~\ref{app:wtl}).
\item \textbf{Text use.} Without text, the MAE rises on most datasets, and no control text lowers it where tested (Appendix~\ref{app:news}).
\item \textbf{Oracle text.} With a text stating the true path, the MAE of the selector falls on most datasets (Appendix~\ref{app:news}).
\item \textbf{Predictor design.} A trained predictor gives lower errors than the candidates of a frozen TSFM (Appendix~\ref{app:design_cmp}).
\item \textbf{Hidden identity.} Hiding the ticker and the company name leaves the test errors unchanged (Appendix~\ref{app:prompt}).
\item \textbf{Robustness.} \ours{} keeps a lower volatility error than the medoid on earlier years, on both FNSPID candidate sets, and after retraining (Appendix~\ref{app:more}).
\item \textbf{Seed spread.} The spread over three seeds stays within about two percent of the MAE (Appendix~\ref{app:seeds}).
\item \textbf{Shape weight.} A larger weight lowers volatility and FD errors where candidates are diverse (Appendix~\ref{app:shape}).
\item \textbf{VLM size.} The 4B and 8B models move the point errors of the 3B model, which we use, by about one percent or less (Appendix~\ref{app:size}).
\item \textbf{Baseline backbones.} Multimodal baselines run with several backbones, and we report the best (Appendix~\ref{app:mm_bb}).
\end{itemize}

%% file: tables/tab_medoid.tex
\begin{table}[t]
\centering
\caption{\textbf{Effect of VLM selection.} Sample is a random candidate, Medoid the one closest to the others, and Time-MMD a dataset mean. $^\dagger$ marks a significant gain of \ours{} over Medoid.}
\label{tab:medoid}
\small
\setlength{\tabcolsep}{1.25pt}
\resizebox{\columnwidth}{!}{\begin{tabular}{l|ccc|ccc|ccc}
\toprule
\multirow{2.5}{*}{Rule} & \multicolumn{3}{c|}{FNSPID} & \multicolumn{3}{c|}{FinTexTS} & \multicolumn{3}{c}{Time-MMD} \\
\cmidrule(lr){2-4} \cmidrule(lr){5-7} \cmidrule(lr){8-10}
 & MAE & End. & Volatility & MAE & End. & Volatility & MAE & End. & Volatility \\
\midrule
Sample & 1.069 & 1.082 & 0.471 & 1.158 & 1.157 & 0.545 & 1.002 & 1.038 & 0.938 \\
Medoid & 0.789 & 0.797 & 0.401 & 0.825 & 0.843 & \textbf{\color{hred}0.465} & 0.899 & 0.941 & \textbf{\color{hred}0.906} \\
\midrule
\rowcolor{yellow!18}\ours{} & \textbf{\color{hred}0.787} & \textbf{\color{hred}0.793} & \textbf{\color{hred}0.355}$^\dagger$ & \textbf{\color{hred}0.811}$^\dagger$ & \textbf{\color{hred}0.826}$^\dagger$ & \textbf{\color{hred}0.465} & \textbf{\color{hred}0.891} & \textbf{\color{hred}0.933} & 0.924 \\
\bottomrule
\end{tabular}}
\end{table}

%% file: tables/tab_objective_all.tex
\begin{table}[t]
\centering
\caption{\textbf{Effect of the selection target.} Mean test error change (\%) over the ten datasets when each term is added to the MAE-only target of the selector, on the same candidates.}
\label{tab:objective_all}
\small
\setlength{\tabcolsep}{4pt}
\begin{tabular}{l|cccc}
\toprule
Target & MAE & MSE & Volatility & FD \\
\midrule
+ Endpoint & +0.28 & +0.10 & +0.41 & +3.21 \\
\rowcolor{yellow!18}+ Shape (\ours{}) & \textbf{\color{hred}$-$0.32} & \textbf{\color{hred}$-$0.36} & \textbf{\color{hred}$-$1.57} & \textbf{\color{hred}$-$2.56} \\
+ Endpoint + Shape & \underline{\color{hblue}+0.08} & \underline{\color{hblue}$-$0.06} & \underline{\color{hblue}$-$0.92} & \underline{\color{hblue}$-$1.34} \\
\bottomrule
\end{tabular}
\end{table}

%% file: tables/tab_chart_num.tex
\begin{table}[t]
\centering
\caption{\textbf{Chart input vs.\ number input.} Test errors of the medoid and the selector reading the candidates as numbers or a chart. $^\dagger$ marks a significant gain of the chart in a test clustered by time.}
\label{tab:chart_num}
\small
\setlength{\tabcolsep}{3pt}
\begin{tabular}{l|ccccc}
\toprule
Input & MAE & MSE & Volatility & Range & FD \\
\midrule
Medoid & 0.789 & 1.172 & 0.401 & 0.322 & 5.91 \\
\midrule
Number & \textbf{\color{hred}0.787} & \textbf{\color{hred}1.163} & \underline{\color{hblue}0.368} & \underline{\color{hblue}0.309} & \underline{\color{hblue}5.42} \\
\rowcolor{yellow!18}Chart (\ours{}) & \textbf{\color{hred}0.787} & \textbf{\color{hred}1.163} & \textbf{\color{hred}0.355}$^\dagger$ & \textbf{\color{hred}0.306}$^\dagger$ & \textbf{\color{hred}5.25} \\
\bottomrule
\end{tabular}
\end{table}

%% file: tables/tab_ablation.tex
\begin{table}[t]
\centering
\caption{\textbf{Ablation of the chart reader.} Each row removes one part of \ours{}. Rank of each setting among the three on each test set (1 is best), averaged over all datasets.}
\label{tab:ablation}
\small
\setlength{\tabcolsep}{3.5pt}
\begin{tabular}{l|cccc}
\toprule
Setting & MAE & MSE & Volatility & FD \\
\midrule
\rowcolor{yellow!18}\ours{} & \textbf{\color{hred}1.50} & \textbf{\color{hred}1.50} & \textbf{\color{hred}1.70} & \textbf{\color{hred}1.80} \\
w/o LoRA on the vision tower & 2.50 & 2.60 & 2.50 & \underline{\color{hblue}2.00} \\
w/o chart auxiliary loss & \underline{\color{hblue}2.00} & \underline{\color{hblue}1.90} & \underline{\color{hblue}1.80} & 2.20 \\
\bottomrule
\end{tabular}
\end{table}

%% file: sec/6_conclusion.tex
\section{Conclusion}
\label{sec:conclusion}

We studied how to turn the many sampled futures of a stochastic forecaster into a single forecast using text. We showed that both central summaries and point-error selectors suffer from the calm-path bias, returning paths calmer than real ones. We proposed \ours{}, where a text-free predictor samples candidate futures and a fine-tuned VLM selects one by reading them as a chart with the text, trained with a shape-aware target. \pend{On ten datasets, \ours{} ranks first on average among all methods and keeps the point error of the medoid while it moves toward more volatile paths, as the truth does on most rows.}

\noindent\textbf{Limitations and future work.}
\label{sec:limits}
\ours{} returns only sampled paths, and a rare spike outside the candidates stays out of reach (Appendix~\ref{app:security}), which heavy-tailed predictors and more TS pretraining data could address. Although the text helps the choice, the gap to an oracle text stays large (Appendix~\ref{app:news}), reflecting the inherent difficulty of multimodal TS forecasting, where a real text reveals only part of the future. We plan to narrow this gap and extend \ours{} to longer horizons and multivariate targets.
\makeatletter\typeout{ENDPROBE page \thepage\space col \if@firstcolumn 1\else 2\fi\space total \the\pagetotal\space goal \the\pagegoal}\makeatother

%% file: sec/X_qa.tex
\onecolumn
\section{Anticipated Questions}
\label{app:qa}

We answer questions that a reviewer may raise and point to where the paper answers each of them.

{\small
\setlength{\tabcolsep}{5pt}
\renewcommand{\arraystretch}{1.15}
\begin{longtable}{@{}p{0.045\textwidth}p{0.925\textwidth}@{}}
\caption{\textbf{Anticipated questions and answers.}}\label{tab:qa}\\
\toprule
\endfirsthead
\toprule
\endhead
\bottomrule
\endfoot
\endlastfoot
\rowcolor{gray!12}\textbf{Q1} & \textbf{Why read the TS as a chart instead of the raw values?} \\*
\textbf{A1} & \leavevmode VLMs inherit a vision-language alignment from large-scale image-text pretraining, while the alignment between TS and language is far weaker (Section~\ref{sec:related}). On FNSPID, a selector with the same VLM, target, and training reads the same history and candidates as numbers, and only the input format differs. It matches the chart selector in point error but has significantly higher volatility and range errors and a higher FD error (Section~\ref{sec:why_chart}, Table~\ref{tab:chart_num}). \\
\midrule
\rowcolor{gray!12}\textbf{Q2} & \textbf{Why select one candidate instead of averaging them?} \\*
\textbf{A2} & \leavevmode An average of diverse futures is a calm line that no future follows (Figure~\ref{fig:motivation}). The point forecasts of Table~\ref{tab:main_all} have a higher average volatility error than \ours{}, and Section~\ref{sec:calm} shows that the point-error target itself pushes any selector toward calm paths. \\
\midrule
\rowcolor{gray!12}\textbf{Q3} & \textbf{Does the VLM do more than the medoid?} \\*
\textbf{A3} & \leavevmode \pend{Yes. \ours{} lowers the MAE and endpoint error against the medoid on all three benchmarks, and the volatility error significantly on FNSPID (Table~\ref{tab:medoid}). The gain in MAE grows when the candidate set is small (Figure~\ref{fig:numk}). A gradient-boosting selector that regresses the same target from hand-made path features has a higher MAE than \ours{} on eight of the ten datasets and is never significantly better (Appendix~\ref{app:nonvlm}). Such a selector also needs hand-made features and cannot read the text, while the VLM learns the choice end to end from the chart and the text of each window.} \\
\midrule
\rowcolor{gray!12}\textbf{Q4} & \textbf{Do the gains come from the predictor or from the selector?} \\*
\textbf{A4} & \leavevmode \pend{From both. The medoid of our candidates is already a strong path, which shows the value of a text-free predictor trained on the TS (Table~\ref{tab:medoid}). The selector lowers the MAE and endpoint error of the medoid on all three benchmarks and its volatility error on FNSPID (Table~\ref{tab:medoid}), and the trained predictor designs give a lower MAE than frozen TSFM candidates on average (Table~\ref{tab:design_cmp}).} \\
\midrule
\rowcolor{gray!12}\textbf{Q5} & \textbf{Does the selector use the text?} \\*
\textbf{A5} & \leavevmode \pend{Yes, when the text describes the path. We use the two datasets where the text has the largest effect, measured by the gain of the multimodal baselines over their unimodal backbones. On them, replacing the real text of the reported \ours{} by a flipped, unrelated, empty, or shuffled text never lowers its MAE, and an empty or a flipped text raises it significantly on one of them (Table~\ref{tab:ctrl}). The choice also changes on the examples of Figure~\ref{fig:textswap_app}. A probe shows that the text carries a signal on the volatility and range of the path (Table~\ref{tab:news_info}). The same selector given an oracle text that describes the path lowers its MAE far below that with the real text on five datasets and also lowers its volatility error (Table~\ref{tab:oracle}). The same selector trained without text has a higher MAE on seven of the ten datasets and never a significantly lower one (Table~\ref{tab:notext}). The selectors that ignore the text have a higher MAE than \ours{} on most datasets. On FNSPID, the training-free rules have a higher MAE and volatility error at every number of candidates (Figure~\ref{fig:numk}), and a gradient-boosting selector trained on the same target has a higher MAE on eight of the ten datasets (Table~\ref{tab:gbm_mae}). The text never enters the predictor, which keeps noisy text from changing the sampled values (Section~\ref{sec:problem}).} \\
\midrule
\rowcolor{gray!12}\textbf{Q6} & \textbf{How does \ours{} compare over the datasets?} \\*
\textbf{A6} & \leavevmode \pend{\ours{} has a lower average MAE and volatility error over the ten datasets than every baseline, and \ours{} has the lowest median rank (Table~\ref{tab:main_all}, Figure~\ref{fig:rank_box}). In paired tests clustered by forecast time on each dataset, \ours{} has more significant wins than losses against every baseline in both MAE and volatility error (Appendix~\ref{app:wtl}, Table~\ref{tab:wtl}).} \\
\midrule
\rowcolor{gray!12}\textbf{Q7} & \textbf{Why is \ours{} not first on every dataset?} \\*
\textbf{A7} & \leavevmode No forecaster is best on every dataset \citep{qiu2024tfb,li2025tsfmbench}. On the dataset of disaster grants, the truth often jumps in spikes that no candidate of the predictor contains, and the selector then keeps a calm path near the no-change forecast (Appendix~\ref{app:security}). This limit lies in the predictor, which rarely samples such spikes, rather than in the VLM. \\
\midrule
\rowcolor{gray!12}\textbf{Q8} & \textbf{Are the baselines tuned as carefully as \ours{}?} \\*
\textbf{A8} & \leavevmode \pend{Each baseline uses its released settings, with the fixes of Appendix~\ref{app:impl}. For each multimodal framework we report the best of five TS backbones by the mean rank of the test MAE. This is an oracle choice that can only help the baselines, as each framework gets its best backbone in hindsight, while every choice for \ours{} is made on validation (Appendix~\ref{app:mm_bb}). The settings of \ours{} that differ per dataset are chosen on validation only, with a correction for the number of changes tried (Appendix~\ref{app:tmmd}). They matter little. With one default setting on every dataset, \ours{} still ranks first on average in both metrics (Appendix~\ref{app:tmmd}, Table~\ref{tab:onesetting}). Neither setting is better on every dataset. The default is slightly ahead on average, as the validation splits of Time-MMD are small and a setting chosen on them does not always carry over to the test year. We still report the row chosen on validation as the main result, as our protocol selects on validation only, and report the default in the appendix.} \\
\midrule
\rowcolor{gray!12}\textbf{Q9} & \textbf{Why does \ours{} have different values in different tables?} \\*
\textbf{A9} & \leavevmode \pend{Table~\ref{tab:main_all} uses the setting chosen on validation per dataset. The design comparison uses one default setting everywhere (Table~\ref{tab:design_cmp}). The FNSPID analyses of Tables~\ref{tab:robust}, \ref{tab:objective}, and~\ref{tab:size} retrain without checkpoint selection, the last with the MAE-only target.} \\
\midrule
\rowcolor{gray!12}\textbf{Q10} & \textbf{Why are the errors computed on normalized paths?} \\*
\textbf{A10} & \leavevmode The datasets differ in scale by orders of magnitude, and the paths of Eq.~\eqref{eq:path} put them on one scale. Every baseline is trained with its released normalization, and its forecast is mapped to the same paths and scored by the same evaluator on the same test rows (Appendix~\ref{app:norm}). The mapping is a fixed function of the history, and it gives no method an advantage. Without the division by $\sqrt{h}$, \ours{} still has the lowest average MAE and the best mean rank in MAE. The ranking also holds on the scale-free mean, which divides the error of each method by that of the no-change forecast in the same space (Table~\ref{tab:relerr}). \\
\midrule
\rowcolor{gray!12}\textbf{Q11} & \textbf{Does a reference path among the candidates do the work of the selector?} \\*
\textbf{A11} & \leavevmode \pend{No. On one Time-MMD dataset, validation favors adding the prior that the benchmark releases as one more candidate (Table~\ref{tab:tmmd_pick}), the same prior that the baselines built for the benchmark read as input. Training-free rules see the same candidate set and do not benefit, as the medoid stays with the sampled paths. The selector learns from the training target when the reference path is the better choice, which is the role of a learned selector over a mixed candidate set.} \\
\midrule
\rowcolor{gray!12}\textbf{Q12} & \textbf{Are the single paths of the TSFMs real samples?} \\*
\textbf{A12} & \leavevmode For the TSFMs that sample, yes, with the default sampler of their released pipeline. Quantile TSFMs return only the quantiles of each step, and their paths keep one quantile level across the horizon, which makes them smoother than real samples (Appendix~\ref{app:impl}). Table~\ref{tab:main_all} thus scores the default point forecast of each TSFM, and single paths appear only in Appendix~\ref{app:single}. \\
\midrule
\rowcolor{gray!12}\textbf{Q13} & \textbf{Can test information leak into the forecasts?} \\*
\textbf{A13} & \leavevmode No. Every dataset is split in time, and every choice of \ours{} is made on validation (Appendix~\ref{app:data}, Appendix~\ref{app:impl}). The predictor is trained from scratch on data that precedes the test period and never reads text, and the selector only chooses among its candidates (Section~\ref{sec:leak}). The selector also does not rely on recalling test outcomes from the pretraining of the VLM. Such recall would bring it close to the best candidate in hindsight, whose error is far below that of \ours{} (Table~\ref{tab:cand}). Hiding the ticker and the company name at test time leaves every error unchanged (Table~\ref{tab:anon}). The FNSPID selector that initializes the other selectors learned only to rank candidate paths of stocks, and it saw no series or report of Time-MMD and no FinTexTS test day (Appendix~\ref{app:data}). The benchmark prior that some Time-MMD settings read is built from earlier values of the same series only. \\
\midrule
\rowcolor{gray!12}\textbf{Q14} & \textbf{How reliable are the results across seeds?} \\*
\textbf{A14} & \leavevmode \pend{We train three seeds for the main results of \ours{} and report their mean, with the same rule for every dataset (Appendix~\ref{app:seeds}). Each trained baseline runs once with the recipe and seed of its release and stops early on validation, and each TSFM runs zero-shot (Appendix~\ref{app:impl}). The standard deviation over seeds stays within about two percent of the MAE (Appendix~\ref{app:seeds}). Paired tests clustered by forecast time appear in Table~\ref{tab:medoid} and in Appendix~\ref{app:more}.} \\
\midrule
\rowcolor{gray!12}\textbf{Q15} & \textbf{Are the volatility and FD errors fair to point forecasters?} \\*
\textbf{A15} & \leavevmode \ours{} keeps the best average rank when the methods are ranked on the MAE columns of Table~\ref{tab:main_all} alone, without the volatility error (averaged over the ten datasets, ahead of every other method). Table~\ref{tab:main_all} reports the MAE of every method on every dataset. The shape errors are meant for paths, and we compare \ours{} on them mainly with other single paths, the medoid and sampled TSFM paths (Table~\ref{tab:medoid}, Table~\ref{tab:single_paths}). A larger shape weight trades point error for these errors on datasets with diverse candidates (Appendix~\ref{app:shape}). On the variogram score, a proper score of the output path, \ours{} has the lowest average and the best mean rank of all methods (Appendix~\ref{app:wtl}). On the CRPS, a proper score of the whole predictive distribution, the candidates of \ours{} score better than the output paths of every TSFM on average (Appendix~\ref{app:wtl}). \\
\midrule
\rowcolor{gray!12}\textbf{Q16} & \textbf{Does \ours{} depend on one predictor?} \\*
\textbf{A16} & \leavevmode The selector runs on all three predictor designs (Table~\ref{tab:designs}). The trained predictor gives a lower MAE and volatility error than the candidates of each of three frozen TSFMs (Table~\ref{tab:design_cmp}). On FNSPID, the volatility gains over the medoid also hold on a second candidate set and in two earlier test years of the same market (Section~\ref{sec:other}, Tables~\ref{tab:rolling} and~\ref{tab:robust}). \\
\midrule
\rowcolor{gray!12}\textbf{Q17} & \textbf{How costly is \ours{}?} \\*
\textbf{A17} & \leavevmode \ours{} keeps the pretrained weights of the VLM frozen and trains only the LoRA adapters, the head, and the predictor, and no method that trains a smaller share of its parameters ranks as high (Figure~\ref{fig:eff_share}). One forecast takes about as long as that of the zero-shot LLM forecaster and, like the baselines, fits on one GPU. \\
\midrule
\rowcolor{gray!12}\textbf{Q18} & \textbf{Does a larger VLM help?} \\*
\textbf{A18} & \leavevmode No. On FNSPID, the larger VLMs do not lower the point errors of the 3B model, which we therefore use, and on FinTexTS and Time-MMD they change each error by about one percent or less (Appendix~\ref{app:size}). \\
\midrule
\rowcolor{gray!12}\textbf{Q19} & \textbf{Is the method specific to finance?} \\*
\textbf{A19} & \leavevmode No. The method uses no property of prices, and eight of the ten datasets come from Time-MMD, which spans domains beyond finance, such as health, energy, and traffic (Section~\ref{sec:setup}, Table~\ref{tab:data}). \\
\midrule
\rowcolor{gray!12}\textbf{Q20} & \textbf{What is new over re-ranking sampled outputs?} \\*
\textbf{A20} & \leavevmode Minimum Bayes risk decoding returns the medoid, verifiers and reward models rank text answers, and the closest method lets an LLM rank the numeric continuations of a frozen TSFM given text (Section~\ref{sec:related}). \ours{} reads all candidates in one chart through the vision-language alignment and trains the selector with a shape-aware target against the calm-path bias (Table~\ref{tab:related}, Section~\ref{sec:target}). The number-input selector of Section~\ref{sec:why_chart} is the closest controlled test of ranking numeric candidates (Table~\ref{tab:chart_num}). \ours{} also has a lower MAE than an LLM that ranks the same numeric candidates without training on every dataset (Table~\ref{tab:lafp}). Other recent methods let an LLM revise the forecast of a TSFM \citep{liu2026posttime} or turn the text into scenarios that guide a diffusion forecaster \citep{tran2026scenariodiff}. Both let the text change the forecast values, while \ours{} keeps the sampled values and uses the text only to choose among them. \\
\midrule
\rowcolor{gray!12}\textbf{Q21} & \textbf{The no-change forecast has a low MAE on FNSPID. Why forecast a path?} \\*
\textbf{A21} & \leavevmode A no-change forecast is a flat line and tells nothing about how much the series moves. Averaged over the ten datasets, it has a higher MAE than \ours{} and a higher volatility error than every method in Table~\ref{tab:main_all}. Relative to the no-change forecast on each dataset, \ours{} has the lowest scale-free MAE of all methods, and its two rows the lowest volatility errors (Appendix~\ref{app:wtl}). \\
\midrule
\rowcolor{gray!12}\textbf{Q22} & \textbf{What does the oracle text show, and what does it not show?} \\*
\textbf{A22} & \leavevmode It shows that the selector can read a text and ground it in the chart: with a text that describes the path, its MAE falls on seven of the ten datasets, by a large margin on five (Table~\ref{tab:oracle}). It is an upper bound and not the gain of the real text. How much a real text helps depends on what it tells about the future path, and a probe finds a signal on its volatility and range (Table~\ref{tab:news_info}). \ours{} keeps the text out of the predictor, and a text with little signal thus cannot harm the sampled values (Section~\ref{sec:problem}). \\
\midrule
\rowcolor{gray!12}\textbf{Q23} & \textbf{Where does the selector gain most over the medoid?} \\*
\textbf{A23} & \leavevmode In the shape of the path. The medoid is already a strong point forecast, and the selector keeps or lowers its MAE and endpoint error on all three benchmarks while lowering the volatility error significantly on FNSPID (Table~\ref{tab:medoid}). This also holds in two earlier test years, on a second candidate set, and after retraining (Table~\ref{tab:rolling}, Table~\ref{tab:robust}). An MAE-only selector moves the other way and raises the volatility error significantly in all of these settings (Tables~\ref{tab:rolling} and~\ref{tab:robust}), which is why the shape term is needed (Table~\ref{tab:objective_all}). \\
\midrule
\rowcolor{gray!12}\textbf{Q24} & \textbf{Why these shape properties and not the direction of the path?} \\*
\textbf{A24} & \leavevmode The direction and size of the move are the endpoint of the path, and adding an endpoint term to the target raises every error, while the shape term lowers every error (Table~\ref{tab:objective_all}). The shape properties are the ones that the point error flattens, the spread of the steps and the largest drop and rise (Section~\ref{sec:target}, Appendix~\ref{app:target}). \\
\midrule
\rowcolor{gray!12}\textbf{Q25} & \textbf{How does the medoid differ from the median of each step?} \\*
\textbf{A25} & \leavevmode The medoid is one whole candidate, the one closest to all others, and keeps the moves of a sampled path. The median of each step takes each step from a different candidate, so the moves of the candidates cancel and the result is a calm path that no candidate follows (Section~\ref{sec:calm}). The candidate closest to this median is one of the rules of Figure~\ref{fig:numk}. \\
\bottomrule
\end{longtable}
}
\twocolumn

%% file: sec/X_suppl.tex
\raggedbottom
\FloatBarrier
\section{Additional Related Work}
\label{app:related}

\noindent\textbf{Probabilistic forecasters.}
Probabilistic models sample future paths with autoregressive diffusion \citep{rasul2021timegrad}, diffusion with adaptive noise schedules \citep{lee2024ant}, or score-based models \citep{tashiro2021csdi}. Adversarial training also generates synthetic TS \citep{yoon2019timegan}. Recent TSFMs model a TS as a token sequence \citep{ansari2024chronos}, as patches with a decoder \citep{das2024timesfm,cohen2025toto}, as patches with quantile outputs \citep{ansari2025chronos2,liu2025moirai2}, or with a mixture output \citep{woo2024moirai}. Domain-specific models tokenize prices \citep{shi2025kronos} or generate paths with a diffusion transformer \citep{kit2026}. Besides the point error, they are scored with distribution scores \citep{gneiting2007scoring}. Stochastic models are also used for video \citep{mathieu2016deep,denton2018svg} and weather \citep{price2025gencast}, where a single mean prediction blurs the fine structure of the future.

\noindent\textbf{Selection in language models.}
Language models sample many answers and pick one with a trained verifier \citep{cobbe2021verifiers}, and they are also trained with a learned reward \citep{ouyang2022instructgpt}. \ours{} follows the same idea for TS, with candidate paths in place of answers and a VLM that reads them as a chart in place of the verifier that scores each answer.

\noindent\textbf{Text-conditioned forecasting.}
Text has long been used to forecast TS, from news events \citep{ding2015event} and tweets paired with prices \citep{xu2018stocknet} to the tone of financial media \citep{tetlock2007giving}.

\noindent\textbf{Vision models for TS.}
TS can be rendered as images for classification \citep{wang2015imaging,li2023vitst}. Networks trained on chart images also forecast future moves \citep{jiang2023reimagining}, although numeric inputs with the same scaling perform on par in that setting. VLMs also answer questions about charts \citep{masry2022chartqa}.

\FloatBarrier
\section{Experimental Setup}
\label{app:setup}

\subsection{Dataset Details}
\label{app:data}

\noindent\textbf{FNSPID.}
We use the daily prices and news of FNSPID \citep{dong2024fnspid}, which pairs US-listed stocks with dated news. The history is the last 90 daily candles, the text is the news about the stock over the last 5 trading days, and the forecast is the high, low, and close path of the next 10 trading days. Each article belongs to the first trading day whose close comes after its publication time in New York, and an article with a date but no time belongs to the next trading day. Thus, \textit{no text of the input is published after the last close of the history}. The point error and the shape terms use the close, while the range uses the high and low. The predictor is a diffusion transformer \citep{peebles2023dit} that follows the KiT architecture \citep{kit2026} and is trained on prices up to 2017. The selector is trained on 2018 to 2021, validated on 2022, and scored on 10{,}000 stock-days of 2023 sampled with a fixed seed. We also build a second candidate set from a predictor trained up to 2021 to check whether the selector transfers to unseen candidates from a different predictor.

\noindent\textbf{FinTexTS.}
FinTexTS \citep{lee2026fintexts} pairs 100 US stocks with daily LLM summaries of company, related-company, sector, and macro news. We follow its protocol with a 64-day input, a 3-day horizon, and the years 2019 to 2021, 2022, and 2023 for training, validation, and test.

\noindent\textbf{Time-MMD.}
We use eight Time-MMD datasets \citep{liu2024timemmd} as separate tasks and follow the protocol of MM-TSFlib: each dataset is split 70/10/20 in time, the input has 24 steps, and the text is the report of the last input step. We leave out Environment, the only daily dataset of Time-MMD, to keep all eight on monthly or weekly steps under one protocol. The horizon is 12 steps, an official horizon of every dataset. The benchmark prior that some predictors of Appendix~\ref{app:tmmd} read is the historical average that Time-MMD releases for each future step, built from earlier values of the same series such as the same month of previous years, and thus reveals no future value. The MM-TSFlib and TaTS baselines read the same prior as an input. A Time-MMD column in a table is the mean over these eight datasets.

\noindent\textbf{Setup across datasets.}
On FNSPID, $x$ is the log price. On FinTexTS and Time-MMD, the predictor is a flow-matching MLP, and the selector is fine-tuned from the FNSPID selector. Every method is scored on the paths of Eq.~\eqref{eq:path}, and Appendix~\ref{app:norm} details the normalization.

\subsection{Input and Output Normalization}
\label{app:norm}

Each predictor reads the history in its own normalized form and samples the future in its own unit. Both are then mapped to the paths of Eq.~\eqref{eq:path}, which the selector and every metric share. Table~\ref{tab:norm} compares the two groups of datasets.

\begin{table}[t]
\centering
\caption{\textbf{Normalization of each predictor.} Predictor input and output for each group of datasets, and the map to Eq.~\eqref{eq:path}.}
\label{tab:norm}
\small
\setlength{\tabcolsep}{2.5pt}
\begin{tabular}{l|cc}
\toprule
 & FinTexTS, Time-MMD & FNSPID \\
\midrule
Value $x$ & Raw value & Log price \\
Predictor & Flow-matching MLP & KiT transformer \\
Input & $(x_{t-j}-x_t)/\sigma$ & Candle chart \\
$\sigma$ & Std.\ of one-step changes & RMS of log returns \\
Output & $(x_{t+h}-x_t)/\sigma$ & Bins of Eq.~\eqref{eq:path} \\
To Eq.~\eqref{eq:path} & Divide by $\sqrt{h}$ & Bin to value \\
\bottomrule
\end{tabular}
\end{table}

\noindent\textbf{FinTexTS and Time-MMD.}
Let $\sigma$ be the standard deviation of the one-step changes within the input, floored at $10^{-3}$. The input is the history relative to its last value, $(x_{t-j}-x_t)/\sigma$ for each input step $j$. The predictor models the future in the same unit, $(x_{t+h}-x_t)/\sigma$, and clips the training targets at $\pm 30$. This instance scaling removes the level of the series, and for datasets whose level matters the predictor also reads the last value, mean, and standard deviation of the history, each standardized over the training split. Dividing each sample by $\sqrt{h}$ gives the path of Eq.~\eqref{eq:path}.

\input{tables/tab_baseline_variants}
\input{tables/tab_val_sweep}

\noindent\textbf{FNSPID.}
Here $x$ is the log price, and the predictor follows KiT \citep{kit2026}. Its input is a chart image of the last 90 daily candles with a volume panel and a 20-day moving average, not a sequence of numbers. The price axis of the chart spans the lowest low to the highest high of the window, which removes the price level and the size of the moves, and the log of this span is added as an embedding to keep the size. The output is a canvas with one row per future day and 64 value bins that cover $\pm 4$ units of $\sigma\sqrt{h}$, where $\sigma$ is the root mean square of the daily log returns in the input. One half of the canvas marks the bin of the close, and the other half marks the bins between the low and the high. A sampled canvas is read back as the center of the close bin, and as a random point in the top and bottom marked bins for the high and the low. The bins are already in the unit of Eq.~\eqref{eq:path}.

\FloatBarrier
\subsection{Implementation Details}
\label{app:impl}

\noindent\textbf{Selection without training.}
The rules of Figure~\ref{fig:numk} pick one of the same candidates without training. The medoid is the candidate closest to the others, and the mode the candidate with the highest kernel density over whole paths. The closest to median and closest to mean are the candidates nearest to the per-step median and mean, and the maximum likelihood is the candidate with the highest kernel density of its values summed over all steps of the horizon.

\noindent\textbf{Predictor.}
On FinTexTS and Time-MMD, the predictor is the TS-to-Chart flow-matching MLP of Appendix~\ref{app:norm}, and on FNSPID it is the Chart-Native predictor, each chosen on validation. On six Time-MMD datasets, it is first pretrained on FNSPID prices at the dataset frequency, cut before validation. The Chart-Native predictor follows the KiT diffusion transformer \citep{kit2026} with a width of 256 and 6 layers. It is trained from scratch for 30{,}000 steps with a batch of 128 and a learning rate of $2\times10^{-4}$ on FNSPID daily prices up to 2017. It conditions on 90 daily candles and samples 10 future candles with 32 flow steps. The second set uses the same recipe up to 2021, and Appendix~\ref{app:diversity} reports the quality of both sets. Sample in Table~\ref{tab:medoid} averages all candidates.

\noindent\textbf{Selector.}
The selector uses Qwen2.5-VL-3B-Instruct with LoRA of rank 16 and dropout 0.05 on all attention and MLP projections of the language model and the vision tower. The scoring head is a two-layer transformer of width 512 with 8 heads, which reads the marker states of the candidates and the last prompt token. The auxiliary loss has a weight of $0.5$. We train for 6{,}000 steps with AdamW, 8 rows per step, 16 random candidates per row, and a learning rate of $10^{-4}$ for the adapters and $10^{-3}$ for the head. Evaluation uses all 32 candidates. We evaluate every 1{,}000 steps on 3{,}000 validation rows and keep the step whose chosen candidates have the lowest target of Eq.~\eqref{eq:target} on these rows. Each panel is $224\times112$ pixels in a four-column grid and shows the recent history and the horizon. On FinTexTS and each Time-MMD dataset, the selector is initialized from the FNSPID selector and keeps its chart, prompt, losses, and hyperparameters, apart from the settings of Appendix~\ref{app:tmmd}.

\noindent\textbf{Selection target on validation.}
We tune one factor at a time on the 2022 FNSPID validation year, starting from the MAE-only selector. Table~\ref{tab:val_sweep} lists the final validation metrics of every setting against the medoid on the same rows. A shape weight of $0.1$ \textit{barely changes the point errors}, such as the MAE and the HL error, but \textit{clearly lowers the shape errors}, such as the volatility, range, and FD errors.

\noindent\textbf{Baselines.}
We use the base checkpoints of Chronos-T5, Chronos-Bolt, and Moirai-1.1 and the small checkpoint of Moirai-2.0. TSFMs that sample paths draw them with the default sampler of their released pipeline. Quantile TSFMs give only the quantiles of each step, and each of their paths keeps one quantile level across the horizon, with the levels evenly spaced. These paths are smoother than real samples. MM-TSFlib and TaTS read the text through a frozen BERT and GPT-2 embedding, respectively. Each trained baseline runs once with the recipe and seed of its release and stops early on our validation split, and each TSFM runs zero-shot. TimeMCL keeps its released recipe with one fix, as its decoder does not reset the LSTM state between hypotheses. ViTime reads either the full input window, as its documentation describes, or only the last steps, as its code does by default, chosen per dataset on validation. Table~\ref{tab:baseline_variants} compares our runs with both models exactly as released, and our runs give the lower MAE on most datasets. The released scaler of TimeMCL sets the scale of a constant input to one, which inflates its error on the few FNSPID rows with a flat price throughout the whole input window.

\noindent\textbf{Metrics.}
Let $r_h$ be the one-step changes of a path in units of $\sigma$. The volatility error is the absolute gap between the standard deviations of $r_h$ of the chosen path and of the truth, and Appendix~\ref{app:metrics} gives the formula of every metric. FD fits a Gaussian to seven features of the chosen paths and one to those of the true paths, each feature divided by its spread over the true paths, and reports the Fr\'echet distance between the two. The features are the volatility, the mean absolute change, the largest drop, the largest rise, the total change, the lag-one autocorrelation of the changes, and the range. A paired $t$-statistic is computed on the per-row differences averaged within each forecast time, and we call a difference significant when its absolute value exceeds two. On Time-MMD, each dataset is one series with one row per forecast time, and the test reduces to a paired $t$-test over overlapping windows, which can overstate significance.

\subsection{No Test Leakage}
\label{app:leak}

Every dataset is split in time, and \textit{every choice of \ours{} is made on validation}. The predictor is trained only on data that precedes the test period and never reads text, and the selector only chooses among its candidates. The selector does not recall outcomes from the pretraining of the VLM either, as it stays far from the best candidate in hindsight (Appendix~\ref{app:diversity}), which such recall would approach. Hiding the ticker and the company name from the prompt and the news leaves its test errors unchanged (Appendix~\ref{app:prompt}).

\subsection{Selection Target}
\label{app:target}

\noindent\textbf{Full target.}
Eq.~\eqref{eq:target} with the endpoint term is
\begin{equation}
\begin{aligned}
d_k = {}& \tfrac{1}{H}\textstyle\sum_{h=1}^{H} |\hat{y}_{k,h} - y_h| + \lambda_{\text{end}}\,|\hat{y}_{k,H} - y_H| \\
& + \lambda_{\text{shape}}\,\tfrac{1}{4}\textstyle\sum_{j=1}^{4} |\phi_j(\hat{y}_k) - \phi_j(y)|/s_j ,
\end{aligned}
\label{eq:target_full}
\end{equation}
where $\phi_j$ is the $j$-th shape property and $s_j$ is a fixed scale close to the spread of $\phi_j(y)$ over true paths. The range term is zero where the data give no range. \ours{} uses $\lambda_{\text{shape}}{=}0.1$ and $\lambda_{\text{end}}{=}0$, the endpoint targets $\lambda_{\text{end}}{=}0.5$.

\noindent\textbf{Shape properties.}
The properties are computed in units of $\sigma$ (Section~\ref{sec:problem}). They are the volatility, the largest fall from a running peak, the largest rise from a running trough, and the mean high-low range on FNSPID. Appendix~\ref{app:shapeprop} gives the formula of each of these four properties.

\FloatBarrier
\subsection{Prompt}
\input{tables/tab_anon}
\label{app:prompt}

\noindent\textbf{Template.}
The FNSPID prompt reads as follows, where the news block lists up to 8 items of the last 5 trading days with the date, title, and summary of the two newest items.
\begin{tcolorbox}[enhanced,colback=cvprblue!6,frame hidden,boxrule=0pt,arc=0pt,left=6pt,right=4pt,top=3pt,bottom=3pt,borderline west={2pt}{0pt}{cvprblue},before skip=4pt,after skip=4pt]
\small\ttfamily\raggedright
Stock \{ticker\}. News of the last days, newest first:\\
- [03-23] \{title\}: \{summary\}\\
\dots\\[2pt]
The image has 32 numbered panels on the same price scale. Each shows the same last 40 trading days and a different possible path for the next 10 days (shaded).\\
\{image\}\\
Candidate 1\{marker\}\\
\dots\\
Candidate 32\{marker\}\\
Which candidate is the most plausible future given the chart and the news?
\end{tcolorbox}
\noindent For FinTexTS and Time-MMD, the first line names the TS, and the news block becomes the text of the last step.

\noindent\textbf{Named vs.\ anonymized prompt.}
To ask whether the selector leans on what the pretrained VLM knows about a company, we evaluate the same selectors on the FNSPID test set with the ticker and the company name replaced by a neutral phrase in the prompt and in the news. As shown in Table~\ref{tab:anon}, every error stays within a small fraction of a percent of that with the named prompt. This indicates that the selector \textit{does not rely on recognizing the identity of the stock}.

\FloatBarrier
\section{Results in Detail}
\label{app:detail}

\subsection{Paired Significance Tests}
\label{app:wtl}
\input{tables/tab_wtl}
\input{tables/tab_relerr}
\input{tables/tab_crps}
\input{tables/tab_vs}

Table~\ref{tab:wtl} compares \ours{} with each baseline of Table~\ref{tab:main_all} on each dataset with a paired $t$-test clustered by forecast time, where $|t|>2$ counts as significant. Total adds the two metrics, and red bold marks the largest count in each row. Against every baseline, \ours{} has more significant wins than losses in both MAE and volatility error, and against most TSFMs it never loses. Its lead in Table~\ref{tab:main_all} thus \textit{does not rest on averaging datasets of different scales}.

\needspace{3\baselineskip}
Table~\ref{tab:relerr} removes the scale of each dataset by dividing every error by that of the no-change forecast on the same rows. In both the arithmetic and the geometric mean of this ratio, \ours{} has the lowest MAE and the lowest volatility error among all methods of Table~\ref{tab:main_all}.

The volatility error scores a single path. For a score that rewards the whole predictive distribution, Table~\ref{tab:crps} compares the CRPS of the candidate set of \ours{} with that of the output paths of each TSFM. The candidates of \ours{} have the lowest CRPS on average, showing that \textit{its predictor gives a better distribution to choose from}.

The single output path can also be scored by a proper score. The variogram score compares how every pair of steps of a path moves relative to each other with the same pairs of the truth. A calm line and a path that moves at the wrong steps both score badly \citep{scheuerer2015variogram}. In Table~\ref{tab:vs}, Rank is the mean rank over all datasets. \ours{} has the lowest variogram score on average and the best mean rank over the ten datasets against every baseline, and its gain is thus \textit{not tied to the volatility error}. In paired tests on each dataset, clustered by time, \ours{} has more significant wins than losses in the variogram score against every baseline except TimeMCL, with which it ties.

\subsection{More Analyses on FNSPID}
\input{tables/tab_rolling}
\input{tables/tab_robust}
\input{tables/tab_objective}
\input{tables/tab_oracle}
\input{tables/tab_news_info}
\input{tables/tab_main}
\label{app:more}

\noindent\textbf{Robustness over time.}
To ask whether the results hold in other periods, we move the whole FNSPID protocol back by one and two years, evaluating on 2022 and 2021 with the same predictor and the configurations fixed from the main split. As shown in Table~\ref{tab:rolling}, \ours{} keeps lower volatility, range, and FD errors than the medoid in both folds, with a point error no worse than the medoid. Its point errors, the MAE, endpoint, and HL errors, \textit{stay close to those of the MAE-only selector}, while its shape errors, the volatility, range, and FD errors, \textit{drop by a wide margin}. Training on MAE alone raises the volatility error above the medoid in both folds, as in the main split of 2023.

\noindent\textbf{Retraining without checkpoint selection.}
To validate that the gains do not depend on a lucky checkpoint, we retrain each FNSPID configuration on 2018 to 2022, which merges the training and validation years, and evaluate the final step on 2023. As shown in Table~\ref{tab:robust}, the retrained \ours{} reproduces the gains of the original runs on both candidate sets. Its point errors \textit{stay close to those of the MAE-only selector}, while its volatility, range, and FD errors \textit{are far lower}. The MAE-only selector also keeps its pattern on both candidate sets, a lower MAE with a significantly worse volatility. This indicates that the trade-off between the two targets is \textit{stable rather than lucky} across retraining.

\noindent\textbf{Selection target on two candidate sets.}
Table~\ref{tab:objective} compares the four selection targets on both FNSPID candidate sets with the full set of metrics, including the endpoint, upper and lower path and range errors. On both sets, adding the shape term to the MAE target \textit{barely changes the point errors}, the MAE, endpoint, and HL errors, while it \textit{clearly lowers the shape errors}, the volatility, range, and FD errors.

\subsection{Single Paths of TSFMs}
\label{app:single}

\noindent\textbf{Single TSFM paths vs.\ \ours{}.}
A TSFM that samples paths can give one path by drawing a sample or by taking the medoid of its samples. As shown in Table~\ref{tab:single_paths}, both a sampled path and the medoid of samples have a larger point error than \ours{} on nine of the ten datasets and on average. Each cell uses the best of Chronos-T5, Moirai-1.1, and Toto, the TSFMs that sample paths on every dataset, chosen per dataset and path type on test MAE. This choice favors the TSFMs, as the opponent changes to the strongest one on each dataset, and \ours{} still leads on average in both errors. Where a TSFM path has the lower volatility error, its MAE is clearly higher, except on Security. On Security, the three are close, as every candidate stays near the recent level and leaves little to choose from (Appendix~\ref{app:security}).

\section{Further Analyses}
\label{app:further}

\subsection{Role of the Text}
\label{app:news}

\noindent\textbf{Evidence of text use.}
\pend{Three results support the role of the text in the choice. 1) With an oracle text that describes the path, the MAE and volatility error of the same selector fall on most datasets (Table~\ref{tab:oracle}), which shows that the selector reads the text and grounds it in the chart. 2) The real text carries a signal on the volatility and range of the future path (Table~\ref{tab:news_info}). 3) With the text of another row, the selector changes its choice on the examples of Figure~\ref{fig:textswap_app}.}

\input{tables/tab_ctrl}
\input{tables/tab_notext}
\noindent\textbf{Real vs.\ oracle text.}
\pend{To separate the selector from the information in the text, we train the same selector with an oracle text that states the true total change, extremes, and typical step size of each row. As shown in Table~\ref{tab:oracle}, the oracle text lowers the MAE on seven of the ten datasets, by a large margin on five of them, and the volatility error with it. The four datasets with almost no change belong to Time-MMD, whose candidates spread the least (Appendix~\ref{app:diversity}), where the candidates are too close to each other for any text to separate them. This indicates that the selector \textit{grounds the text in the chart} when the text describes the path. The small gain from the real text thus reflects \textit{how little it says about the future path} rather than a limit in the ability of the VLM to select among the candidates.}

\noindent\textbf{Future signal in the text.}
\pend{Table~\ref{tab:news_info} reports a probe that asks whether the text carries information on the future path. A ridge regression predicts the five properties of the true path from features of the candidates alone or together with TF-IDF features of the text. A lower error with the text means that the text helps the probe predict the future path.}

\noindent\textbf{Corrupted text.}
To check whether each method uses the content of the text, we keep each trained model and only replace the text of each test row by a corrupted text. Flipped reverses every direction word, Unrelated is a report of another series, Empty says no report exists, and Shuffled is another row's text. We use the two Time-MMD datasets where the text matters most, measured without \ours{} as the median MAE gain of MM-TSFlib, CFA, and TaTS over the same TS backbone (Table~\ref{tab:main_all}, Appendix~\ref{app:mm_bb}). Each baseline is retrained with its setting of Table~\ref{tab:main_all}. In Table~\ref{tab:ctrl}, a positive value means that the broken text makes the forecast worse, which shows that the method relies on the real text. For \ours{}, every corrupted text raises the MAE on both datasets, significantly for the empty and the flipped text. The baselines barely react, and some even improve with a broken text. The selector thus \textit{reads the direction of the real text}, while the baselines make little use of its content.

\input{tables/tab_shape}
\input{tables/tab_mm_bb}
\input{tables/tab_gbm_mae}
\noindent\textbf{Training without text.}
To ask whether the selector gains from the text when it is trained with it, we train the selector of \ours{} in its default setting with an empty text and score it the same way. Removing the text changes the choice on a large share of the rows, and on the other rows the text cannot matter. We thus compare the MAE on the rows whose choice changes. As shown in Table~\ref{tab:notext}, removing the text raises it on every benchmark. Thus, the text helps the trained selector, while its effect is bounded by how much the text says about the path (Table~\ref{tab:oracle}).

\subsection{Weight of the Shape Term}
\label{app:shape}

\noindent\textbf{Weight sweep.}
\pend{To show how the shape term trades point error for realism, we vary its weight on every dataset with all other settings at their defaults. As shown in Table~\ref{tab:shape}, a larger weight lowers the volatility error on most datasets, most strongly where the candidates are diverse, while the MAE rises with the weight. The selector thus \textit{trades a small point error for a more realistic path}, and a small weight keeps the MAE close to its lowest value. We keep the default weight for all datasets rather than tuning it per dataset, since the weight changes the validation target itself and its best value differs from one metric to another.}

\subsection{Multimodal Baselines with Other \mbox{Backbones}}
\label{app:mm_bb}

\setcounter{dbltopnumber}{3}% three wide floats on this page (Tables 30, 31, 32)
\input{tables/tab_lafp}
\input{tables/tab_seeds}
\input{tables/tab_design_cmp}

MM-TSFlib and TaTS add the text to a TS backbone. We run MM-TSFlib with the four classic models and Transformer, TaTS with five backbones that mix channels, as TaTS feeds the text as extra input channels, and CFA with five backbones. As shown in Table~\ref{tab:mm_bb}, the backbone changes the error of each model. Table~\ref{tab:main_all} reports the backbone with the lowest mean rank of MAE for each model, which is PatchTST for MM-TSFlib, iTransformer for TaTS, and PatchTST for CFA. ChatTime is used zero-shot from its released checkpoint, while the other models are trained on each dataset with their released settings. Time-VLM builds on CLIP, and GPT4MTS, Time-LLM, and CALF build on GPT-2, with Time-LLM using GPT-2 in place of LLaMA to fit the compute budget. CALF has no text input in its released form, and we prepend the text tokens to the input of both its temporal and textual branches.

\subsection{Selectors without a VLM}
\label{app:nonvlm}

\pend{To ask whether a VLM is needed to choose among the candidates, we train a gradient-boosting selector that sees no chart and no text and regresses the selection target of \ours{} from path, centrality, and history features, on the same candidates and rows. As shown in Table~\ref{tab:gbm_mae}, \ours{} has a lower MAE on eight of the ten datasets, significantly on six, and is not significantly worse on any. \ours{} is also lower on the selection target that both regress on the same eight datasets, significantly on five. A feature-based selector needs features designed for the target and cannot read the text, while the VLM learns the choice \textit{directly from the chart and the text of each row}.}

The closest prior method, LAFP \citep{nguyen2026lafp}, lets an LLM rank the numeric continuations of a TSFM given text. We evaluate this ranking step without training on the same candidates, history, text, and test rows, with the same VLM used as a text-only LLM that returns the index of one candidate. As shown in Table~\ref{tab:lafp}, \ours{} has a lower MAE than this ranker on every dataset, significantly on eight. This indicates that \textit{ranking the candidates with an LLM is not enough} without the chart input and the trained selector of \ours{}.

% --- commented out (user 2026-11-16): not a required table ---
% \noindent\textbf{Gain over the medoid per dataset.}
% \pend{Table~\ref{tab:medoid_mae} splits the comparison with the medoid of Table~\ref{tab:medoid} by dataset. \ours{} has a lower MAE than the medoid of the same candidates on eight of the ten datasets. The gain is significant on three datasets, and the MAE of \ours{} is not significantly above that of the medoid on any dataset.}

% \input{tables/tab_medoid_mae}

\subsection{Seed Spread}
\label{app:seeds}

\pend{Table~\ref{tab:seeds} reports the spread of \ours{} over its three seeds. On every dataset, the standard deviation of the test MAE stays within about two percent of the MAE.}

\subsection{Trained vs.\ Frozen Predictor}
\label{app:design_cmp}

Table~\ref{tab:design_cmp} asks whether training our own predictor matters. It pairs the same selector in one default setting with the candidates of each of three frozen TSFMs that return real samples, Chronos-T5, Moirai-1.1, and Toto, and with those of the trained predictor of \ours{}. The trained predictor gives the lowest MAE on every dataset except Security and the lowest volatility error on most datasets. Where a frozen TSFM has the lower volatility error, its MAE is higher, except on Security (Appendix~\ref{app:security}). Thus, \textit{the selector gains from a predictor trained on the target series} rather than a released TSFM. Both rows return one sampled path and are thus not comparable with the point forecasts of the same TSFMs in Table~\ref{tab:main_all}, which average over samples and are thus calmer by the calm-path bias of Section~\ref{sec:calm}.

\begin{figure*}[t]
\centering
\includegraphics[width=\textwidth]{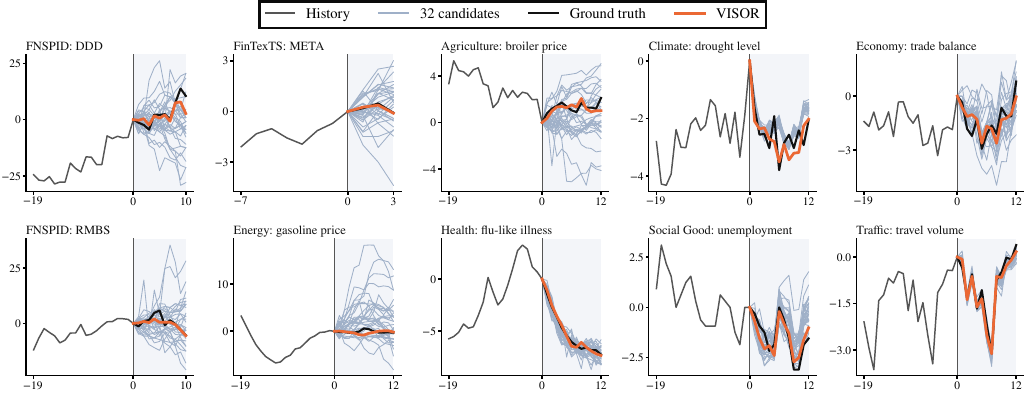}
\caption{\textbf{Diversity of the candidates.} All predictor candidates on test rows of every benchmark, with the truth and the pick of \ours{}.}
\label{fig:diversity}
\end{figure*}

\subsection{Diversity of the Candidates}
\setcounter{dbltopnumber}{2}%
\label{app:diversity}
\input{tables/tab_diversity}
\input{tables/tab_cand}

\noindent\textbf{Candidate sets.}
The predictor samples candidates that differ in level, trend, and volatility, and the selector chooses one of them. Figure~\ref{fig:diversity} shows all candidates of test rows from every benchmark with the truth and the path chosen by \ours{}. The candidates spread around the truth, and in these rows \ours{} picks one that follows its turns rather than the center of the candidate set.

\noindent\textbf{Spread and headroom.}
Table~\ref{tab:diversity} measures the spread by the mean distance between two candidates and the headroom by the error of the best candidate in hindsight, both relative to the error of the medoid. On every benchmark the best candidate lies far closer to the truth than the medoid, which leaves room for selection. The room is largest on the stock benchmarks, whose candidates differ the most.

\noindent\textbf{Candidate quality.}
On FNSPID, Table~\ref{tab:cand} scores one random draw, the medoid, and the best candidate in hindsight among the candidates of each test row. The best candidate in hindsight has a far lower error than the medoid, which shows that \textit{the candidates often contain a path close to the truth}. The medoid has a low point error but a high FD, as it always returns a calm central path, while a random candidate keeps the spread of the true paths but has a high point error. The two candidate sets give similar errors for every choice. A gain on the second set thus reflects the selector rather than easier candidates to choose from.

\begin{figure}[t]
\centering
\includegraphics[width=\columnwidth]{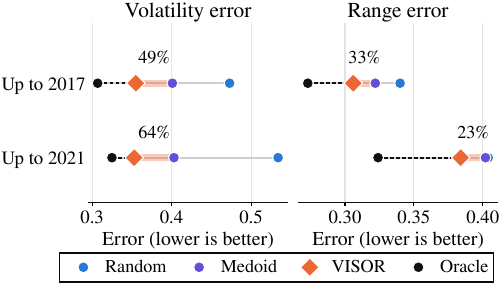}
\caption{\textbf{Selection headroom on FNSPID.} Volatility and range errors of each choice on the candidates of the predictors trained up to 2017 and 2021. The percentage is the share of the gap from the medoid to the Oracle that \ours{} closes.}
\label{fig:headroom}
\end{figure}

\noindent\textbf{Headroom in volatility and range.}
On financial TS, the volatility and the range of the path carry the risk. We thus examine the headroom in these two errors. Figure~\ref{fig:headroom} compares a random candidate, the medoid, \ours{}, and the candidate with the lowest MAE in hindsight (Oracle) on both FNSPID candidate sets. First, the Oracle lies far below the random candidate and the medoid, which shows that a path close to the truth is often already among the candidates. This highlights \textit{the importance of choosing well beyond predicting well}. Second, \ours{} lies below the random candidate and the medoid in all four cases and closes a large part of the gap from the medoid to the Oracle. Third, a clear gap to the Oracle remains, especially in the range, which leaves room for better selectors in future work.

\subsection{VLM Size}
\input{tables/tab_size}
\label{app:size}

To ask whether a larger VLM chooses better, we compare the Qwen2.5-VL-3B selector with Qwen3-VL-4B and Qwen3-VL-8B selectors on FNSPID, all trained with the MAE-only target. As shown in Table~\ref{tab:size}, the 4B and 8B models do not lower the point errors of the 3B model, which has the largest MAE gain over the medoid. We attribute this to the task being bounded by \textit{what the charts and the text reveal} rather than by the capacity of the VLM. A larger VLM may also need more training pairs than the few multimodal TS datasets offer, which leaves its extra capacity unused. We therefore use the 3B model in all other experiments.

With the shape-aware target and one default setting, Table~\ref{tab:size_shape} replaces the 3B selector by larger ones on FinTexTS and Time-MMD. Every error moves by about one percent or less, and the 3B selector also loses little to larger VLMs on these benchmarks, which supports the 3B choice.

\input{tables/tab_size_shape}

\subsection{Rendering Ablations}
\input{tables/tab_ablation_detail}
\label{app:abl_detail}

We vary the rendering of \ours{} one at a time on the FNSPID validation set. As shown in Table~\ref{tab:ablation_detail}, larger panels and a band of all candidates behind a path do not beat the default line chart, which we keep on every dataset. Larger panels barely change the errors, as the default panel already shows the shape of each path. The band raises every error, as the other candidates crowd each panel and hide its path.

\subsection{Tuning on Time-MMD}
\label{app:tmmd}

\noindent\textbf{Tuning per dataset.}
\pend{To ask whether \ours{} needs tuning per dataset, we try several changes on each dataset with three seeds.%
\input{tables/tab_tmmd_pick} They cover the text window, the number of candidates, the history length, the optimizer, and the sampling temperature of the predictor. They also cover mixed pools that sample half of the candidates at a lower temperature or half at a lower and half at a higher one,%
\input{tables/tab_onesetting}\begin{figure*}[t]
\centering
\includegraphics[width=\textwidth]{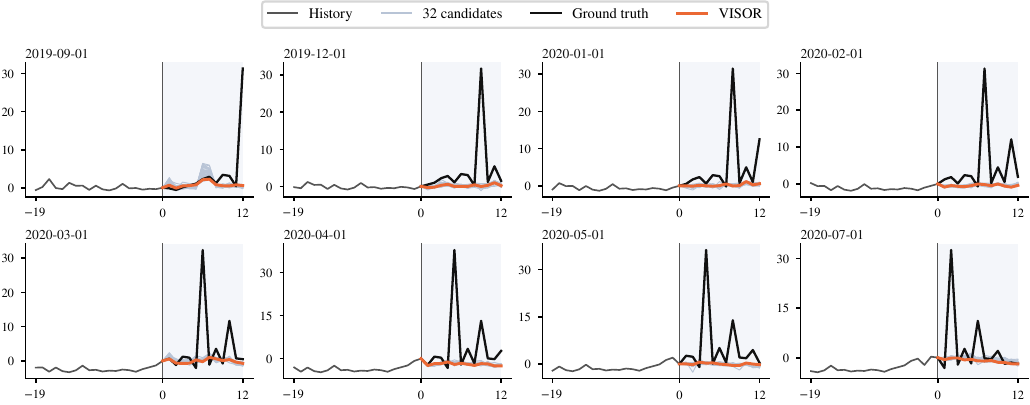}
\caption{\textbf{Qualitative analysis on Security.} Test rows whose truth holds a spike, with all candidates of the predictor, the truth, and the path chosen by \ours{}. \textit{Almost no candidate comes close to the timing and size of the spike}.}
\label{fig:security}
\end{figure*}%
 predictors that also read the prior of each horizon step of Time-MMD, as TaTS and MM-TSFlib do,%
\input{tables/tab_relerr_cfg} training candidates drawn out of fold, and the prior path added as one candidate. A reference path of the task can join the candidates as one more panel, and the selector learns from the training target when to choose it. As shown in Table~\ref{tab:tmmd_pick}, only the mixed pools, a lower sampling temperature, the out-of-fold training candidates (with the prior path on one dataset), and the prior-reading predictor lower the validation target beyond seed noise on five datasets once the number of tried changes is accounted for. Two more datasets keep the default selector and change only their predictor setting, which is chosen by cross-validation as described below. The remaining dataset keeps the default. The shape weight is left out of this search, since it changes the validation target itself.}

\noindent\textbf{Setting-agnostic performance.}
Table~\ref{tab:onesetting} runs \ours{} with one default setting on every dataset instead of the setting chosen on validation, as FNSPID and FinTexTS already do. Neither setting is better on every dataset, and the default is slightly ahead on average, as the validation splits of Time-MMD are small. Where a check on the most recent validation rows favors it, the test candidates come from the predictor refit on training and validation. The chart type, what the chart draws, the prompt, and the losses are the same on every dataset. With either setting, \ours{} has a lower average MAE and volatility error than every baseline. Table~\ref{tab:relerr_cfg} puts both settings next to the best baseline of each group under the scale-free average of Table~\ref{tab:relerr}. The two settings lead every group in the volatility error by a wide margin, and they differ far less from each other.

\noindent\textbf{Predictor setting by cross-validation.}
\pend{The validation splits of some Time-MMD datasets are too small to rank predictor settings, as some have only a few calm forecast windows. We therefore also choose the setting of each predictor, such as its width, dropout, and training length, from five-fold out-of-fold forecasts over the training and validation windows together. A setting replaces the default only if it lowers the medoid MAE over these windows under a paired one-sided $t$-test with a Bonferroni correction for the settings tried. Settings that add the extra columns of a domain are left out, since their folds share these columns. Two datasets change their predictor under this rule, and their selectors are then trained on the new candidates with the default settings of Appendix~\ref{app:impl} and no further tuning.}

\subsection{Failure Case: Spikes}
\label{app:security}

\noindent\textbf{Why spikes are hard.}
On Security, the MAE of every method lies in a narrow range, and \ours{} does not lead it (Table~\ref{tab:main_all}). The series counts disaster grants.%
\input{tables/tab_eff_infer} It stays calm for long stretches until a disaster sets off a burst of grants whose timing the past does not reveal. More of its test rows hold such a spike than on the other Time-MMD datasets, and on these rows almost no candidate reaches even half of the spike. The best candidate in hindsight is barely better than the no-change path, while on the other datasets it is far better. Figure~\ref{fig:security} shows the spike rows, where every candidate stays near the recent level while the truth jumps.

\noindent\textbf{Calm paths under spikes.}
With no candidate near the spike, the path with the lowest expected error is one that stays near the recent level or its mean. \ours{} picks one of the calmest candidates on most Security rows, more often than on the other datasets, and on the spike rows its path moves only a small fraction as much as the truth. This calm choice keeps its MAE at that of the no-change path, which places \ours{} among the baselines rather than ahead of them here.

\noindent\textbf{Limit of the predictor.}
The selector only chooses among the candidates and cannot recover a spike that none of them contains. The limit therefore \textit{lies in the predictor}, which learns from a history with few spikes, rather than in the VLM. Forecasting the timing and size of such rare events, for example by sampling candidates from a heavy-tailed model or tying them to event reports, remains future work.

\section{Efficiency Analysis}
\label{app:eff}

On the FNSPID test rows, Table~\ref{tab:eff_infer} reports the inference cost per forecast with the parameters, paths, and peak memory. Runs are batched as in our experiments with model loading excluded, and Paths is the number of paths each model returns per forecast. The selector time of \ours{} includes rendering the charts, and $^\dagger$ marks a whole zero-shot run divided by the test rows. Without batching, one forecast of \ours{} takes under two seconds. \ours{} takes longer than the TSFMs and about as long as the zero-shot LLM forecaster, and needs less GPU memory than Toto.

\section{Method Details}
\label{app:method}
\input{tables/tab_shapes}

\subsection{Training Order}
\label{app:order}

\ours{} is trained in two stages.
\begin{enumerate}[leftmargin=1.4em,itemsep=1pt]
\item \textbf{Predictor.} Trained on the history alone without text, then frozen. It samples the $K$ candidates of every window once.
\item \textbf{Selector.} The VLM with LoRA and a head learns to score these candidates from the chart and the text. No gradient reaches the predictor.
\end{enumerate}
At test time, the frozen predictor samples the candidates and the selector returns the one with the highest score.

\subsection{Input Shapes}
\label{app:shapes}

\noindent\textbf{Panel rendering.}
Each panel draws the history as a dark gray line and the candidate future as a blue line over a light shaded band, on a white background with a thin gray border and its number at the top left. The colors separate the history from the candidate, and the prompt names the shaded part as the future. The candidates are mapped back from the normalized space of the predictor to the level of the series divided by its last observed value, and all panels of one grid share one vertical axis, so the VLM compares the sizes of the moves directly. The panels form one grid image of four columns, and a larger $K$ adds rows of panels to it.

\noindent\textbf{From panels to tokens.}
Table~\ref{tab:shapes} follows one FNSPID row through the selector. The steps from the image to the tokens are the standard input pipeline of Qwen2.5-VL, which we keep unchanged.%
\input{tables/tab_pred_arch}%
\input{tables/tab_design_pick} It cuts the image into $14{\times}14$ patches in three color channels and merges each $2{\times}2$ block of patches into one token of the language model. One token thus covers $28{\times}28$ pixels, and each panel becomes the same number of tokens for any $K$ in training and evaluation.

\noindent\textbf{Repeated frame.}
The vision encoder of Qwen2.5-VL serves both images and videos, and its first layer reads two consecutive frames at once. A still image is therefore repeated as a second frame, which gives $3{\times}2{\times}14{\times}14$ values per patch. The copy adds no information, as the layer then acts as a single-frame layer whose weights are the sum of its two frame weights. Qwen2.5-VL also feeds every still image this way in pretraining.

\noindent\textbf{From tokens to scores.}
The prompt puts the text before the image and one marker token per candidate after it (Appendix~\ref{app:prompt}), and its length depends on the news of the row. Rows in a batch are padded on the left to a common length $T$. The last hidden states at the $K$ markers and at the final token are projected to the width of the head, which returns one score per candidate. The head has no positional encoding, and the same weights score any number of candidates.

\subsection{Predictor Architectures}
\label{app:pred_arch}

Table~\ref{tab:pred_arch} lists the networks of the two trained designs. TS-to-Chart samples numbers that the renderer draws as charts, and Chart-Native draws the future part of the chart itself. Chart-Native uses KiT on FNSPID, whose daily candles it reads and draws, and a small UNet on the other datasets, which have one value per step. For each dataset, validation chooses between the two trained designs (Table~\ref{tab:design_pick}). It picks Chart-Native on FNSPID and TS-to-Chart on every other dataset, so the UNet serves only the design comparison.

\subsection{Auxiliary Chart Reading}
\label{app:aux}

A linear layer maps each candidate state $\mathbf{h}_k$, the hidden state at the marker of panel $k$ (e.g., the token \texttt{<|box\_end|>} that follows ``Candidate~$k$'' in the prompt), to three values of its own panel, whose targets come from the candidate paths alone. With $c_{k,h} = \sqrt{h}\,y^{(k)}_h$ the path of candidate $k$ in units of $\sigma$ and $c_{k,0} = 0$, the three values are as follows.
\begin{itemize}
\item \textbf{Centrality.} The mean absolute gap of $y^{(k)}$ to all candidates of the window, standardized over these candidates.
\item \textbf{Volatility.} The standard deviation of \mbox{$c_{k,h} - c_{k,h-1}$}.
\item \textbf{Total change.} The last value $c_{k,H}$, divided by three.
\end{itemize}
With $g$ this layer and $\mathbf{a}_k$ the three targets, the auxiliary loss and the full training loss of the selector are
\begin{equation}
\mathcal{L}_{\text{aux}} = \frac{1}{3K}\sum_{k=1}^{K} \lVert g(\mathbf{h}_k) - \mathbf{a}_k \rVert^2, \quad
\mathcal{L}_{\text{total}} = \mathcal{L} + 0.5\,\mathcal{L}_{\text{aux}},
\end{equation}
where $\mathcal{L}$ is the regression loss of Eq.~\eqref{eq:loss}. The shape term thus changes what the selector regresses, whereas the auxiliary loss adds a reading task and leaves the target unchanged. The targets use no truth and thus \textit{cannot leak the future}. Their gradient reaches the LoRA of the vision tower through the language model, which pushes the panel tokens to encode how typical, how rough, and how far each path is. The layer serves only in training.

\subsection{Shape Properties}
\label{app:shapeprop}

The four shape properties of the selection target (Appendix~\ref{app:target}) are computed in units of $\sigma$. For a path $y$, let $c_h = \sqrt{h}\,y_h$ with $c_0 = 0$, let $r_h = c_h - c_{h-1}$ be its one-step changes, and let $u_h$ and $l_h$ be its upper and lower paths in the same units. With $h$ over $1, \dots, H$ and $h'$ over $0, \dots, h$,
\begin{itemize}
\item \textbf{Volatility.} $\phi_1 = \operatorname{std}_{h}\, r_h$.
\item \textbf{Largest drop.} $\phi_2 = \max_{h} \big(\max_{h' \le h} c_{h'} - c_h\big)$.
\item \textbf{Largest rise.} $\phi_3 = \max_{h} \big(c_h - \min_{h' \le h} c_{h'}\big)$.
\item \textbf{Range.} $\phi_4 = \frac{1}{H}\sum_{h=1}^{H} (u_h - l_h)$.
\end{itemize}
The fixed scales are $s = (0.4, 2.2, 1.65, 0.4)$, which bring the four terms to a similar size. Without upper and lower paths, $u_h = l_h = c_h$, so $\phi_4$ is zero for every path.

\subsection{Metric Definitions}
\label{app:metrics}

All metrics use the paths of Eq.~\eqref{eq:path}. For a chosen path $\hat{y}$ and the truth $y$, with $c_h$, $r_h$, $u_h$, and $l_h$ as in Appendix~\ref{app:shapeprop}, each test row gives the following metrics.
\begin{itemize}
\item \textbf{MAE and MSE.} $\frac{1}{H}\sum_{h} |\hat{y}_h - y_h|$ and $\frac{1}{H}\sum_{h} (\hat{y}_h - y_h)^2$ on the close.
\item \textbf{Endpoint.} $|\hat{y}_H - y_H|$, the error of the total change.
\item \textbf{HL.} The MAE on the upper and lower paths, averaged.
\item \textbf{Volatility.} $|\operatorname{std}_h \hat{r}_h - \operatorname{std}_h r_h|$.
\item \textbf{Range.} $|\phi_4(\hat{y}) - \phi_4(y)|$, the error of the mean gap between the upper and lower paths.
\end{itemize}
Each is averaged over the test rows. Volatility depends only on the set of the steps, so it ignores when the large moves occur, which MAE and the drop and rise features of FD capture. FD instead compares the chosen paths of a whole test set with the true paths. Each path gives seven features, each divided by its spread over the true paths, and FD fits one Gaussian $(\mu, \Sigma)$ to each of the two sets,
\begin{equation*}
\mathrm{FD} = \lVert \mu_{\text{pick}} - \mu_{\text{true}} \rVert^2 + \operatorname{Tr}\big(\Sigma_{\text{pick}} + \Sigma_{\text{true}} - 2(\Sigma_{\text{pick}}\Sigma_{\text{true}})^{1/2}\big).
\end{equation*}
The first term grows when the chosen paths are calmer on average, and the second when they vary less than the true paths. A set of averaged paths thus scores a large FD even when its MAE is low. FD does not check each row against its truth, so we read it together with MAE.

\noindent\textbf{Radar chart.}
\label{app:radar}
Each axis of Figure~\ref{fig:radar} shows the lowest error among the methods divided by the error of a method. The best method on an axis lies on the outer circle, and a larger error moves a method toward the center. The center is set a little below the lowest of these scores, not at zero.

\section{Selector Behavior}
\label{app:behavior}

\subsection{Number of Candidates in Training}
\label{app:train_k}
\input{tables/tab_train_k}

\noindent\textbf{Training and evaluation sizes.}
Section~\ref{sec:numk} changes the number of candidates only in evaluation. Here we also train the selector with 8, 16, or 32 candidates per row and let each one pick from the first 4 to 32 candidates of every FNSPID test row. As shown in Table~\ref{tab:train_k}, the MAE of every selector falls as the candidate set grows, while a random pick from the same candidates stays flat. The VLM thus \textit{finds a closer path in a larger set of candidates}. The training size matters less, and a selector trained with more candidates gains most when it also picks from more of them. The default of 16 is best or second best in every column, so it \textit{needs no retuning when the number of candidates in evaluation changes}.

\begin{figure*}[t]
\centering
\includegraphics[width=\textwidth]{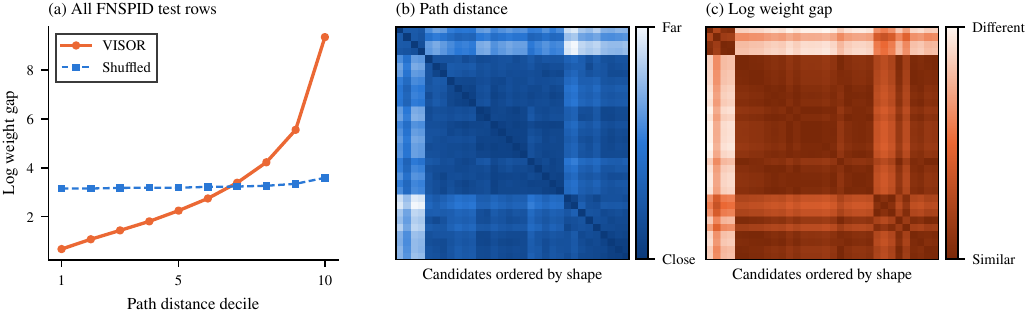}
\caption{\textbf{Weight gap by path distance.} (a) Gap of the log softmax weights of two candidates by the decile of their path distance on the FNSPID test rows. (b, c) Both quantities for the candidate pairs of one test row, ordered by shape.}
\label{fig:simscore}
\end{figure*}
\begin{figure*}[p]
\centering
\includegraphics[width=1.0\textwidth]{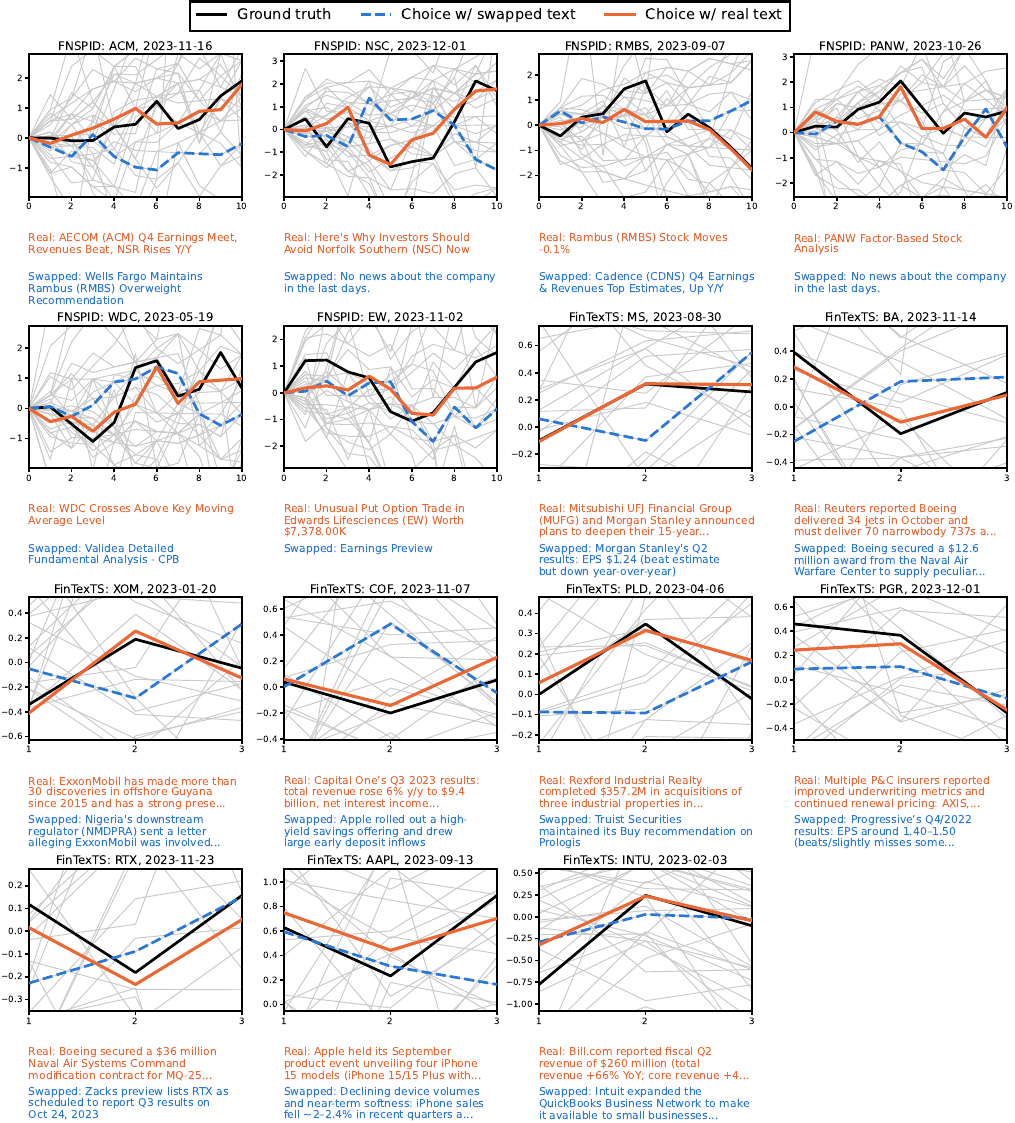}
\caption{\textbf{Real vs.\ swapped text.} The same selector reads the same chart and candidates with the real text or with the text of another test row. Gray lines are the candidates, and each text is cut to its first words.}
\label{fig:textswap_app}
\end{figure*}

\begin{figure*}[t]
\centering
\includegraphics[width=\textwidth]{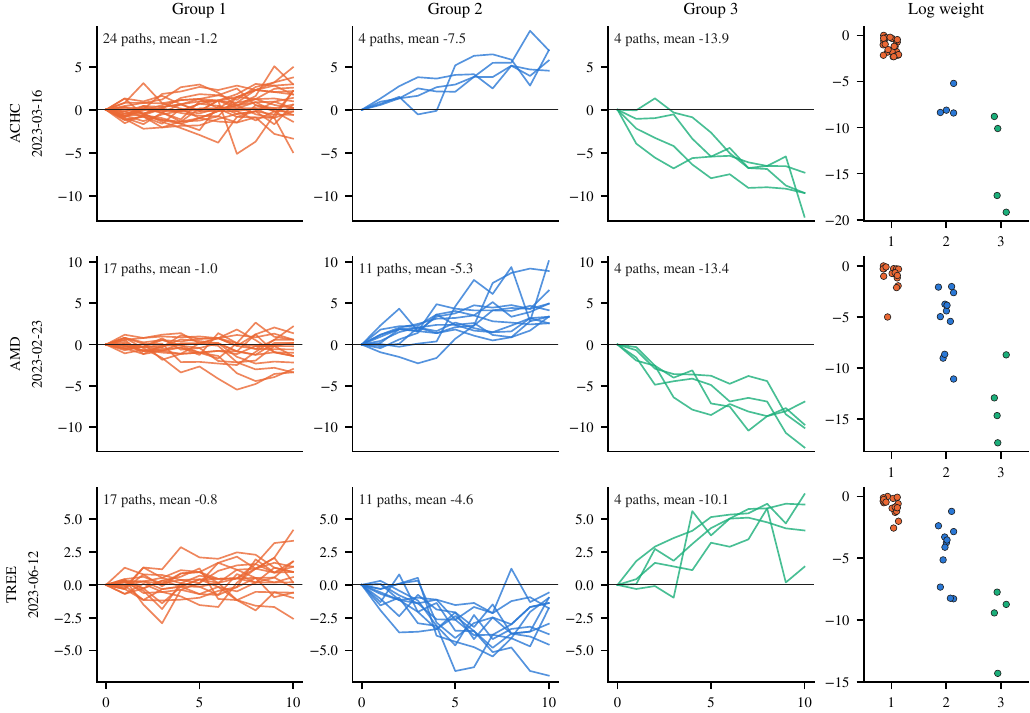}
\caption{\textbf{Candidates by shape group.} Candidates of three FNSPID test rows in three shape groups, with their log softmax weights.}
\label{fig:simscore_groups}
\end{figure*}
\begin{figure*}[tp]
\centering
\includegraphics[width=0.93\textwidth]{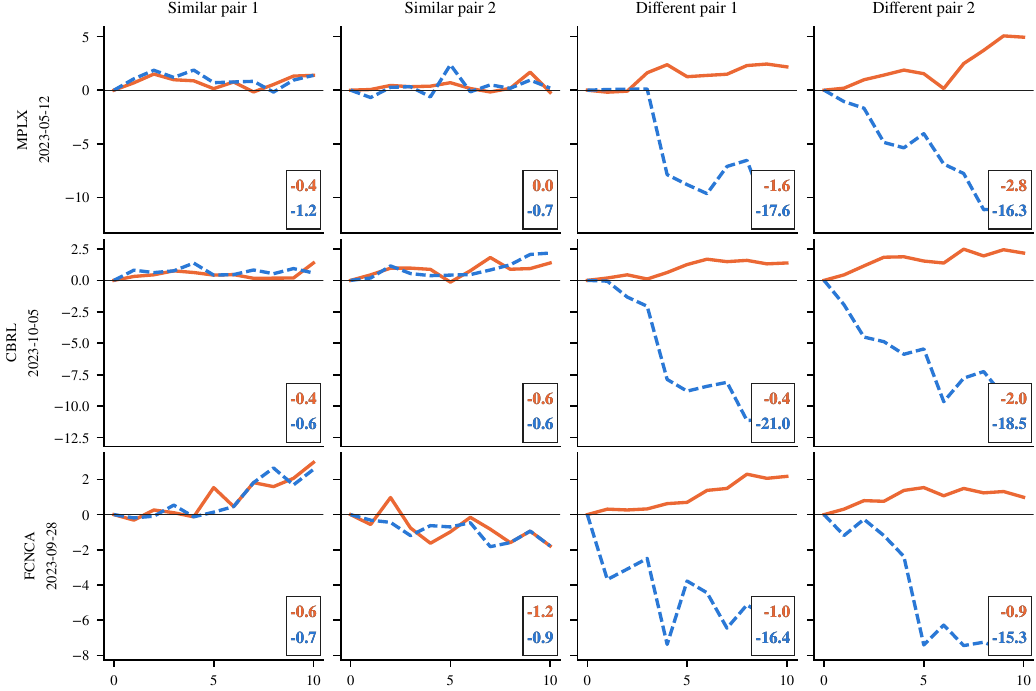}
\caption{\textbf{Similar and different pairs.} Closest and most different candidate pairs in three FNSPID test rows, with log softmax weights.}
\label{fig:simscore_pairs}
\end{figure*}
\begin{figure*}[tp]
\centering
\includegraphics[width=0.95\textwidth]{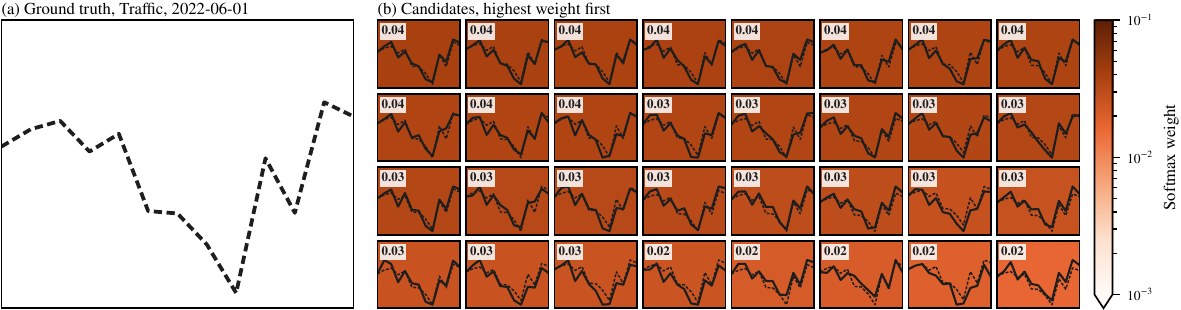}\\[4pt]
\includegraphics[width=0.95\textwidth]{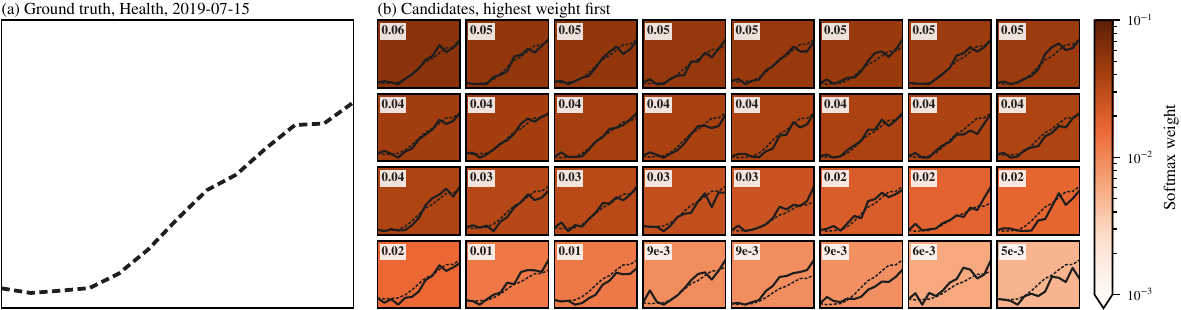}
\caption{\textbf{Candidates of one shape.} Rows that are easy to forecast, where all candidates share the shape of the truth. \textit{When the candidates are alike, they receive almost the same softmax weight}, as none of them is clearly better than the others.}
\label{fig:simscore_same}
\end{figure*}
\begin{figure*}[tp]
\centering
\includegraphics[width=0.95\textwidth]{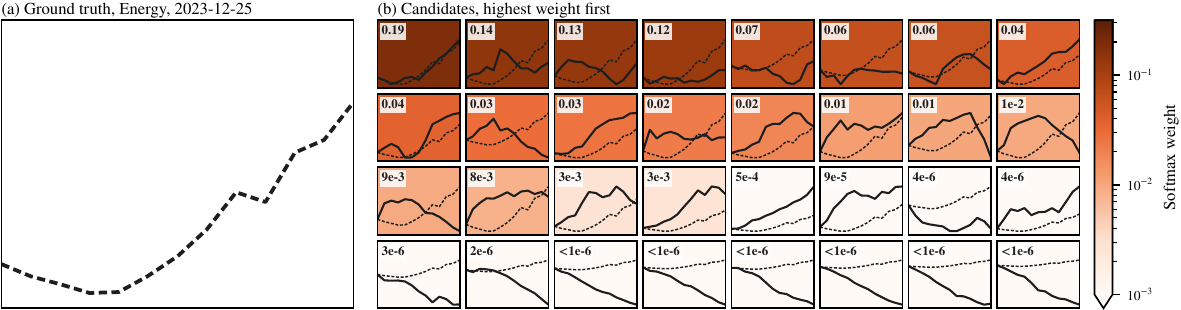}\\[3pt]
\includegraphics[width=0.95\textwidth]{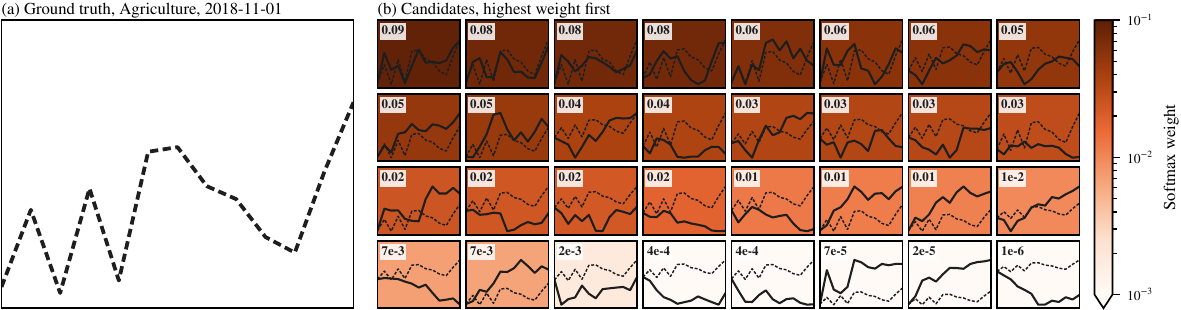}\\[3pt]
\includegraphics[width=0.95\textwidth]{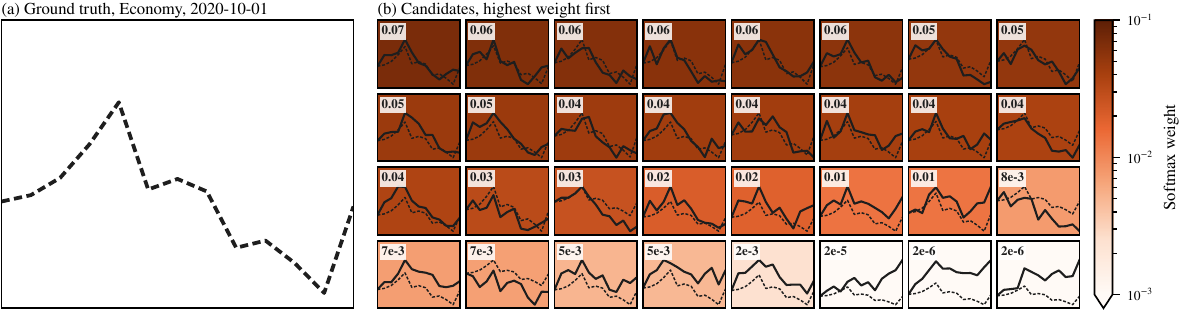}\\[3pt]
\includegraphics[width=0.95\textwidth]{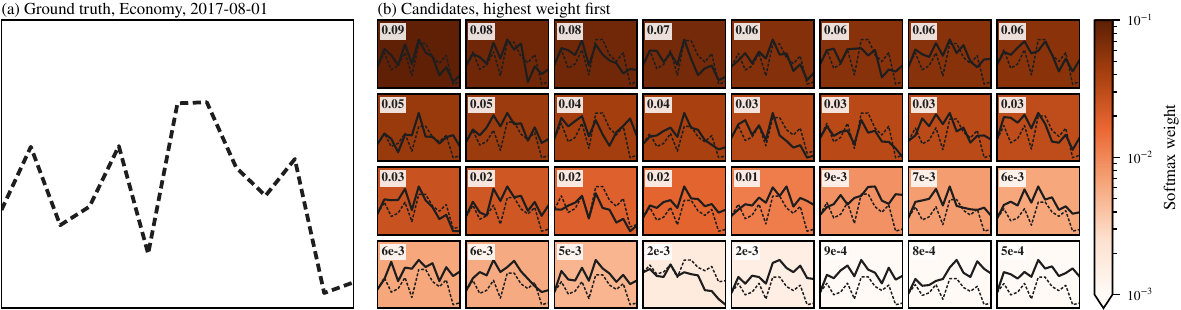}
\caption{\textbf{Candidates of many shapes.} The predictor returns diverse scenarios for each row, and \textit{the candidates closer to the truth receive higher softmax weights}. Each panel is scaled to its own range, with the truth dashed in every panel.}
\label{fig:simscore_ex}
\end{figure*}

\begin{figure*}[p]
\centering
\includegraphics[width=\textwidth]{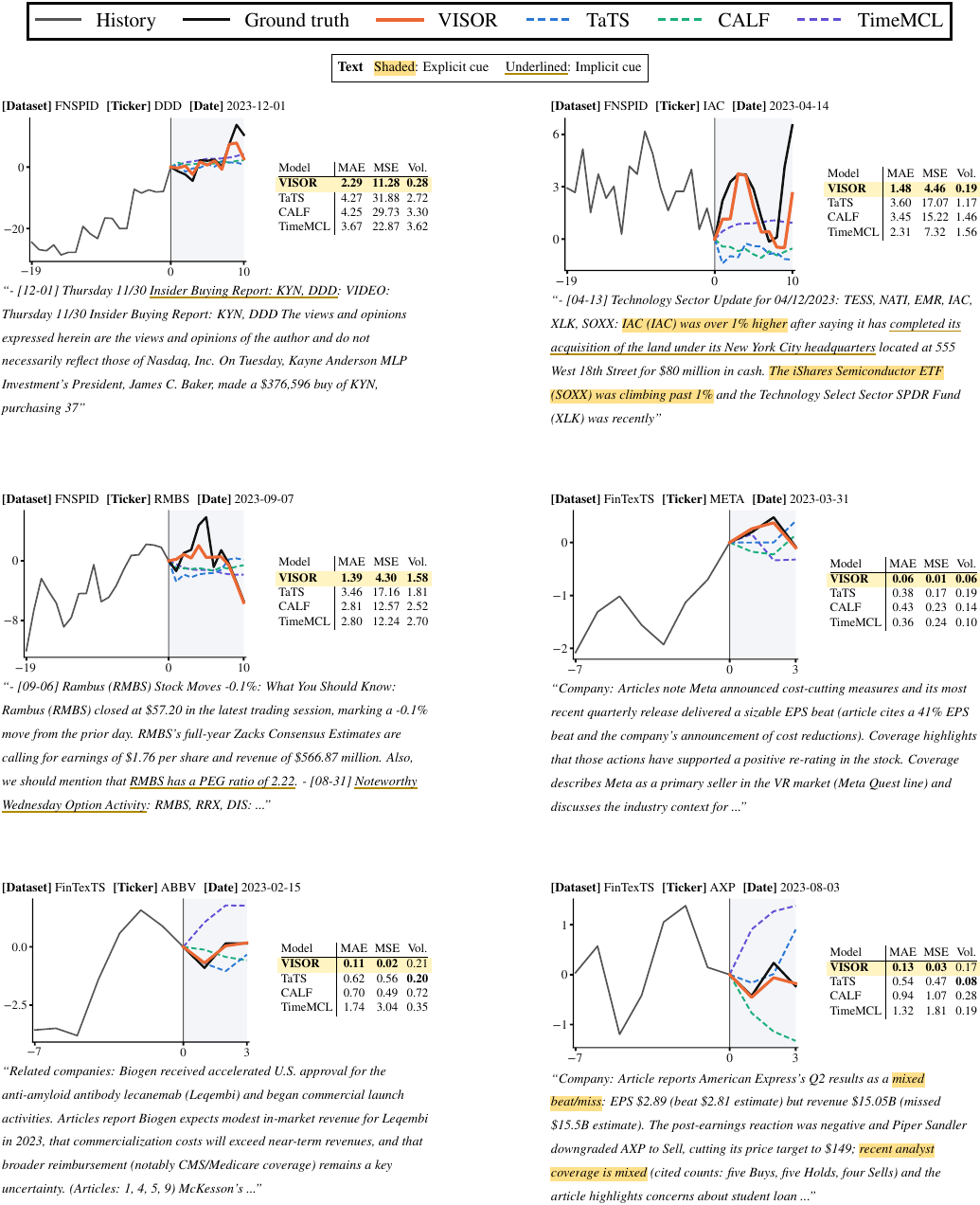}
\caption{\textbf{Forecast visualizations with their text (1/4).} Vol. is the gap in step volatility to the truth.}
\label{fig:grid1}
\end{figure*}

\begin{figure*}[p]
\centering
\includegraphics[width=\textwidth]{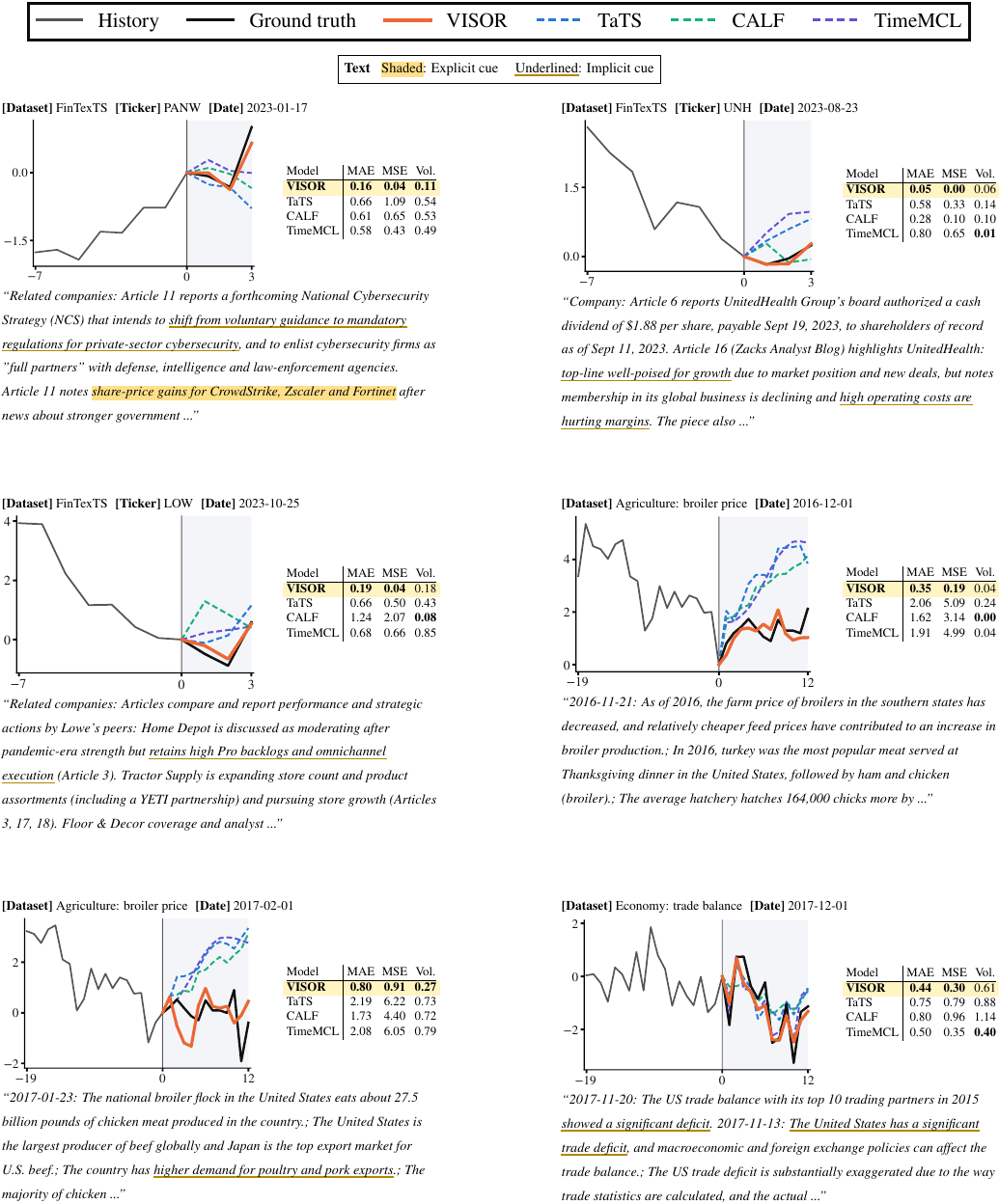}
\caption{\textbf{Forecast visualizations with their text (2/4).} Vol. is the gap in step volatility to the truth.}
\label{fig:grid2}
\end{figure*}

\begin{figure*}[p]
\centering
\includegraphics[width=\textwidth]{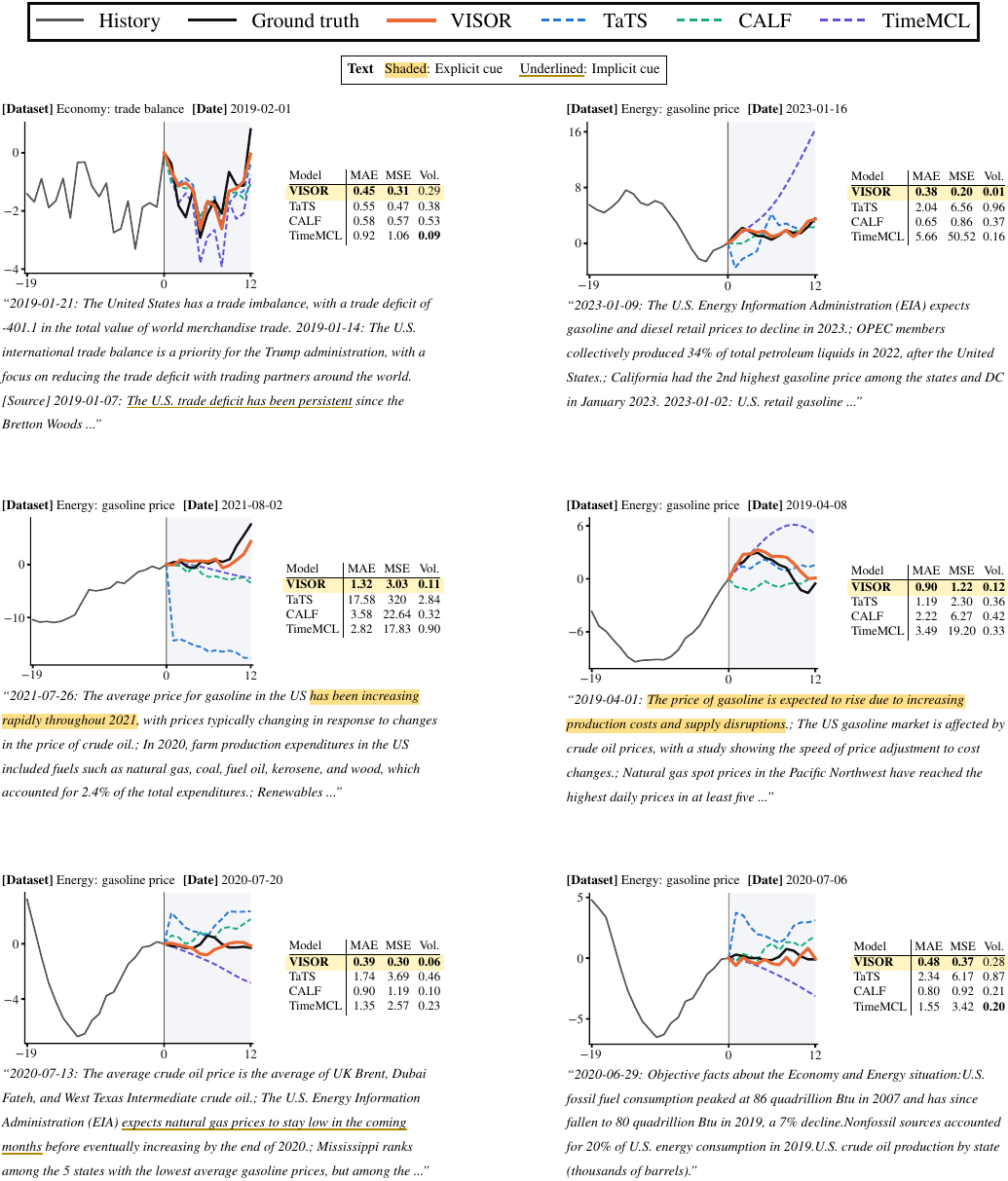}
\caption{\textbf{Forecast visualizations with their text (3/4).} Vol. is the gap in step volatility to the truth.}
\label{fig:grid3}
\end{figure*}

\begin{figure*}[p]
\centering
\includegraphics[width=\textwidth]{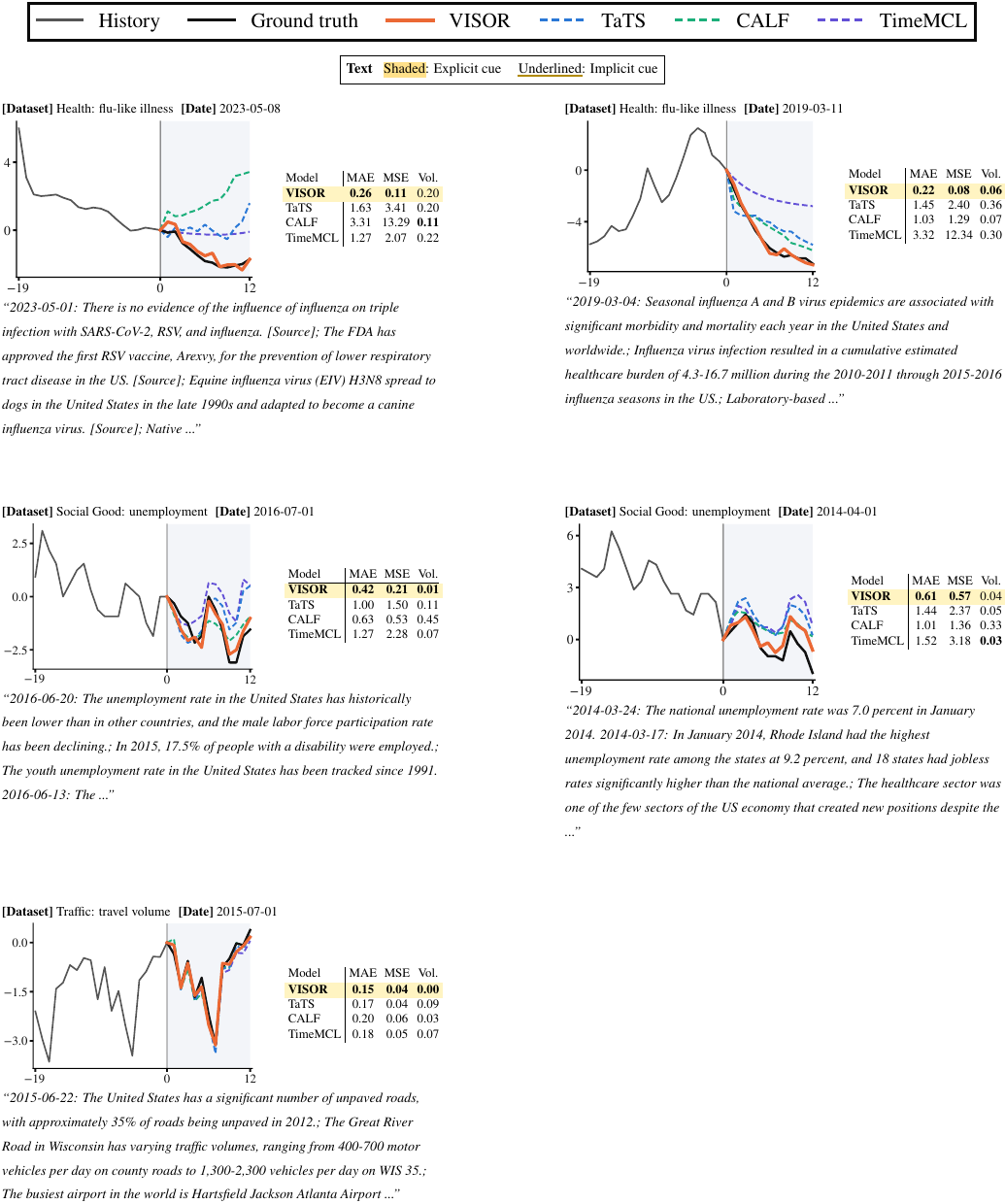}
\caption{\textbf{Forecast visualizations with their text (4/4).} Vol. is the gap in step volatility to the truth.}
\label{fig:grid4}
\end{figure*}

\subsection{Scores of Similar Candidates}
\label{app:simscore}
\input{tables/tab_simscore}

\noindent\textbf{Similar paths, similar weights.}
A selector that reads the shapes should give similar weights to similar candidates and different weights to candidates of different shapes. For every pair of candidates in a test row, we compare the MAE between their paths with the gap of their log softmax weights $\log\mathrm{softmax}(s_k/\tau)$ with $\tau{=}0.1$. We use the log, as the raw weights of two poor candidates are both near zero whatever their shapes. As shown in Table~\ref{tab:simscore}, the rank correlation between the two is positive on every dataset and vanishes when the scores are shuffled within each row.

\noindent\textbf{Examples.}
Figure~\ref{fig:simscore}a shows that the weight gap grows steadily with the path distance on FNSPID, and Figure~\ref{fig:simscore}b,c shows the same block pattern within one test row. Figure~\ref{fig:simscore_groups} splits the candidates of three rows into three shape groups, and each group receives a narrow range of weights far from those of the other groups. Figure~\ref{fig:simscore_pairs} compares the closest and the most different pairs, where close pairs receive almost the same weight. The selector thus \textit{scores the candidates as a smooth function of their shape}.

\noindent\textbf{Candidates with the truth.}
Figures~\ref{fig:simscore_same} and~\ref{fig:simscore_ex} list all candidates of six Time-MMD test rows next to the truth, ordered by their weight. When all candidates share one shape, as in Figure~\ref{fig:simscore_same}, the selector spreads its weight almost evenly over them. Otherwise, as in Figure~\ref{fig:simscore_ex}, the top candidates follow the shape of the truth, and their weight fades steadily as their distance from the truth grows.

\subsection{Training for Text Use}
\label{app:textuse}

\noindent\textbf{Counterfactual term.}
The selector of \ours{} is trained only to pick the best candidate, so nothing forces it to rely on the text. To ask how far the text can steer the choice, we add a counterfactual term to the training of the FNSPID selector. It requires the choice made with the real text of a row to lie closer to the truth than the choice made with the text of another stock on the same day. We score the trained selector on the test set with the real text of each row and with the shuffled text of another row.

\noindent\textbf{Effect of the text.}
With the shuffled text, the selector changes its choice on a large share of the rows. The real text gives a significantly lower volatility error than the shuffled text ($t$-test clustered by forecast time), and both stay below the medoid. Thus, \textit{the text carries information on the shape of the path} that a selector can learn to use.

\section{More Qualitative Examples}
\label{app:examples}

\pend{To show that the choice responds to the text, we give the selector the text of another test row and keep the chart and the candidates. Figure~\ref{fig:textswap_app} shows test rows of FNSPID and FinTexTS on which the choice changes, and in these examples \ours{} picks a path closer to the truth with the real text. For FinTexTS, the swapped text is the text of the same stock on another date of the test period.}

Extending Figure~\ref{fig:examples}, Figures~\ref{fig:grid1}--\ref{fig:grid4} show more test rows from FNSPID, FinTexTS, and Time-MMD. Each panel gives the forecast paths, the error of each model, and a quote from the text the selector saw, with its cues marked.

\section{Broader Impact}
\label{app:impact}

Forecasts of prices and public indicators can guide decisions with real costs. \ours{} returns one plausible path rather than a guaranteed outcome, and its forecasts are not investment advice. All datasets and texts are public.

%% file: tables/tab_baseline_variants.tex
\begin{table*}[t]
\centering
\caption{\textbf{Official vs.\ reproduced baselines.} TimeMCL with its official code and ViTime reading the full input window, against our runs of both in Table~\ref{tab:main_all}. Bold marks the lower error of each pair. \textit{Our runs are generally stronger}, especially in MAE.}
\label{tab:baseline_variants}
\small
\setlength{\tabcolsep}{1.0pt}
\begin{tabular}{l|cc|cc|cc|cc|cc|cc|cc|cc|cc|cc}
\toprule
\multirow{2.5}{*}{Method} & \multicolumn{2}{c|}{FNSPID} & \multicolumn{2}{c|}{FinTexTS} & \multicolumn{2}{c|}{Agriculture} & \multicolumn{2}{c|}{Climate} & \multicolumn{2}{c|}{Economy} & \multicolumn{2}{c|}{Energy} & \multicolumn{2}{c|}{Health} & \multicolumn{2}{c|}{Security} & \multicolumn{2}{c|}{Social Good} & \multicolumn{2}{c}{Traffic} \\
\cmidrule(lr){2-3} \cmidrule(lr){4-5} \cmidrule(lr){6-7} \cmidrule(lr){8-9} \cmidrule(lr){10-11} \cmidrule(lr){12-13} \cmidrule(lr){14-15} \cmidrule(lr){16-17} \cmidrule(lr){18-19} \cmidrule(lr){20-21}
 & MAE & Vol. & MAE & Vol. & MAE & Vol. & MAE & Vol. & MAE & Vol. & MAE & Vol. & MAE & Vol. & MAE & Vol. & MAE & Vol. & MAE & Vol. \\
\midrule
\rowcolor{gray!15}\multicolumn{21}{l}{\textit{TimeMCL}~\citep{cortes2025timemcl}} \\
As released & \textbf{1.075} & 0.959 & 5.921 & 3.142 & 1.208 & 0.762 & 0.314 & 0.757 & 1.235 & 0.371 & 1.828 & \textbf{0.744} & 3.055 & 1.127 & 1.495 & \textbf{3.550} & 1.275 & 0.782 & 0.278 & 0.187 \\
Our run & 1.131 & \textbf{0.921} & \textbf{0.918} & \textbf{0.490} & \textbf{1.094} & \textbf{0.640} & 0.314 & \textbf{0.752} & \textbf{0.388} & \textbf{0.293} & \textbf{1.679} & 0.799 & \textbf{1.712} & \textbf{1.055} & \textbf{0.714} & 3.742 & \textbf{0.804} & \textbf{0.748} & \textbf{0.210} & \textbf{0.184} \\
\rowcolor{gray!15}\multicolumn{21}{l}{\textit{ViTime}~\citep{yang2025vitime}} \\
As released & 1.665 & \textbf{0.556} & 4.016 & 1.586 & 1.676 & \textbf{0.554} & 0.370 & \textbf{0.555} & 0.556 & \textbf{0.672} & 2.711 & 0.859 & \textbf{2.290} & 1.217 & \textbf{0.779} & \textbf{3.628} & 1.297 & \textbf{1.007} & \textbf{0.397} & 0.482 \\
Our run & \textbf{0.982} & 0.576 & \textbf{0.910} & \textbf{0.515} & \textbf{1.268} & 0.655 & \textbf{0.363} & 0.644 & \textbf{0.481} & 0.756 & \textbf{2.343} & \textbf{0.700} & 2.291 & \textbf{1.213} & 0.805 & 3.712 & \textbf{0.946} & 1.074 & 0.398 & \textbf{0.480} \\
\bottomrule
\end{tabular}
\end{table*}

%% file: tables/tab_val_sweep.tex
% selected (paper_select.py); one column, Endpoint column dropped, End. in labels (user 11-23)
\begin{table}[t]
\centering
\caption{\textbf{Selection target on validation.} Each setting is compared with the medoid on 2022 validation. Notation follows Table~\ref{tab:robust}, and End. is the weight of the endpoint term.}
\label{tab:val_sweep}
\small
\setlength{\tabcolsep}{2pt}
\begin{tabular}{l|ccccc}
\toprule
Selection target & MAE & HL & Volatility & Range & FD \\
\midrule
MAE only & \textbf{\color{hred}0.850}$^\ast$ & \textbf{\color{hred}0.806}$^\ast$ & 0.466$^\dagger$ & 0.379$^\dagger$ & 7.875 \\
\rowcolor{yellow!18}+ Shape 0.1 & \underline{\color{hblue}0.852}$^\ast$ & \underline{\color{hblue}0.807}$^\ast$ & \textbf{\color{hred}0.405}$^\ast$ & \textbf{\color{hred}0.362}$^\ast$ & \textbf{\color{hred}6.591} \\
+ End. 0.5 & 0.854 & 0.809 & 0.440 & 0.371 & 7.507 \\
+ End. 1.0 & 0.855 & 0.810 & 0.446 & 0.373 & 7.495 \\
+ End. 0.5 + Shape 0.1 & 0.855 & 0.809$^\ast$ & \underline{\color{hblue}0.418}$^\ast$ & \underline{\color{hblue}0.364} & \underline{\color{hblue}7.029} \\
\bottomrule
\end{tabular}
\end{table}

%% file: tables/tab_anon.tex
\begin{table}[t]
\centering
\caption{\textbf{Named vs.\ anonymized prompt.} Errors of the selectors on FNSPID when the ticker and company name are hidden in prompt and news, with the named prompt of Table~\ref{tab:main_all} and the relative change between the two prompts (\%).}
\label{tab:anon}
\small
\setlength{\tabcolsep}{6pt}
\begin{tabular}{l|cccc}
\toprule
 Prompt & MAE & MSE & Volatility & FD \\
\midrule
Named & 0.787 & 1.163 & 0.355 & 5.253 \\
Hidden identity & 0.788 & 1.163 & 0.355 & 5.257 \\
\midrule
Change (\%) & +0.04 & +0.03 & $-$0.03 & +0.07 \\
\bottomrule
\end{tabular}
\end{table}

%% file: tables/tab_wtl.tex
\begin{table}[t]
\centering
\caption{\textbf{Paired significance tests.} Number of datasets on which \ours{} is significantly better (Win), not different (Tie), or significantly worse (Loss) than each baseline.}
\label{tab:wtl}
\small
\setlength{\tabcolsep}{2.4pt}
\begin{tabular}{l|ccc|ccc||ccc}
\toprule
\multirow{2.5}{*}{Baseline} & \multicolumn{3}{c|}{MAE} & \multicolumn{3}{c||}{Volatility} & \multicolumn{3}{c}{Total} \\
\cmidrule(lr){2-4} \cmidrule(lr){5-7} \cmidrule(lr){8-10}
 & Win & Tie & Loss & Win & Tie & Loss & Win & Tie & Loss \\
\midrule
\rowcolor{gray!15}\multicolumn{10}{l}{\textit{TSFMs}} \\
Chronos-T5 & \cellcolor{yellow!18}\textbf{\color{hred}7} & 3 & 0 & \cellcolor{yellow!18}\textbf{\color{hred}10} & 0 & 0 & \cellcolor{yellow!18}\textbf{\color{hred}17} & 3 & 0\\
Chronos-Bolt & \cellcolor{yellow!18}\textbf{\color{hred}8} & 2 & 0 & \cellcolor{yellow!18}\textbf{\color{hred}10} & 0 & 0 & \cellcolor{yellow!18}\textbf{\color{hred}18} & 2 & 0\\
Chronos-2 & \cellcolor{yellow!18}\textbf{\color{hred}7} & 3 & 0 & \cellcolor{yellow!18}\textbf{\color{hred}10} & 0 & 0 & \cellcolor{yellow!18}\textbf{\color{hred}17} & 3 & 0\\
TimesFM-2.5 & \cellcolor{yellow!18}4 & \textbf{\color{hred}5} & 1 & \cellcolor{yellow!18}\textbf{\color{hred}10} & 0 & 0 & \cellcolor{yellow!18}\textbf{\color{hred}14} & 5 & 1\\
Moirai-1.1 & \cellcolor{yellow!18}\textbf{\color{hred}8} & 2 & 0 & \cellcolor{yellow!18}\textbf{\color{hred}8} & 2 & 0 & \cellcolor{yellow!18}\textbf{\color{hred}16} & 4 & 0\\
Moirai-2.0 & \cellcolor{yellow!18}\textbf{\color{hred}6} & 3 & 1 & \cellcolor{yellow!18}\textbf{\color{hred}10} & 0 & 0 & \cellcolor{yellow!18}\textbf{\color{hred}16} & 3 & 1\\
Toto & \cellcolor{yellow!18}\textbf{\color{hred}7} & 2 & 1 & \cellcolor{yellow!18}\textbf{\color{hred}10} & 0 & 0 & \cellcolor{yellow!18}\textbf{\color{hred}17} & 2 & 1\\
VisionTS & \cellcolor{yellow!18}\textbf{\color{hred}7} & 3 & 0 & \cellcolor{yellow!18}\textbf{\color{hred}7} & 3 & 0 & \cellcolor{yellow!18}\textbf{\color{hred}14} & 6 & 0\\
ViTime & \cellcolor{yellow!18}\textbf{\color{hred}10} & 0 & 0 & \cellcolor{yellow!18}\textbf{\color{hred}10} & 0 & 0 & \cellcolor{yellow!18}\textbf{\color{hred}20} & 0 & 0\\
\midrule
\rowcolor{gray!15}\multicolumn{10}{l}{\textit{Task-specific TS models}} \\
DLinear & \cellcolor{yellow!18}2 & \textbf{\color{hred}7} & 1 & \cellcolor{yellow!18}\textbf{\color{hred}7} & 2 & 1 & \cellcolor{yellow!18}\textbf{\color{hred}9} & \textbf{\color{hred}9} & 2\\
PatchTST & \cellcolor{yellow!18}4 & \textbf{\color{hred}5} & 1 & \cellcolor{yellow!18}\textbf{\color{hred}8} & 1 & 1 & \cellcolor{yellow!18}\textbf{\color{hred}12} & 6 & 2\\
iTransformer & \cellcolor{yellow!18}\textbf{\color{hred}5} & \textbf{\color{hred}5} & 0 & \cellcolor{yellow!18}\textbf{\color{hred}7} & 2 & 1 & \cellcolor{yellow!18}\textbf{\color{hred}12} & 7 & 1\\
TimeMixer & \cellcolor{yellow!18}\textbf{\color{hred}6} & 4 & 0 & \cellcolor{yellow!18}\textbf{\color{hred}7} & 2 & 1 & \cellcolor{yellow!18}\textbf{\color{hred}13} & 6 & 1\\
TimeMCL & \cellcolor{yellow!18}\textbf{\color{hred}4} & 3 & 3 & \cellcolor{yellow!18}\textbf{\color{hred}7} & 2 & 1 & \cellcolor{yellow!18}\textbf{\color{hred}11} & 5 & 4\\
\midrule
\rowcolor{gray!15}\multicolumn{10}{l}{\textit{Multimodal TS models}} \\
MM-TSFlib & \cellcolor{yellow!18}\textbf{\color{hred}5} & \textbf{\color{hred}5} & 0 & \cellcolor{yellow!18}\textbf{\color{hred}6} & 3 & 1 & \cellcolor{yellow!18}\textbf{\color{hred}11} & 8 & 1\\
TaTS & \cellcolor{yellow!18}4 & \textbf{\color{hred}6} & 0 & \cellcolor{yellow!18}\textbf{\color{hred}6} & 1 & 3 & \cellcolor{yellow!18}\textbf{\color{hred}10} & 7 & 3\\
CFA & \cellcolor{yellow!18}3 & \textbf{\color{hred}6} & 1 & \cellcolor{yellow!18}\textbf{\color{hred}7} & 3 & 0 & \cellcolor{yellow!18}\textbf{\color{hred}10} & 9 & 1\\
ChatTime & \cellcolor{yellow!18}\textbf{\color{hred}9} & 1 & 0 & \cellcolor{yellow!18}\textbf{\color{hred}9} & 1 & 0 & \cellcolor{yellow!18}\textbf{\color{hred}18} & 2 & 0\\
Time-VLM & \cellcolor{yellow!18}4 & \textbf{\color{hred}5} & 1 & \cellcolor{yellow!18}\textbf{\color{hred}9} & 1 & 0 & \cellcolor{yellow!18}\textbf{\color{hred}13} & 6 & 1\\
GPT4MTS & \cellcolor{yellow!18}4 & \textbf{\color{hred}5} & 1 & \cellcolor{yellow!18}\textbf{\color{hred}5} & 4 & 1 & \cellcolor{yellow!18}\textbf{\color{hred}9} & \textbf{\color{hred}9} & 2\\
Time-LLM & \cellcolor{yellow!18}\textbf{\color{hred}8} & 2 & 0 & \cellcolor{yellow!18}\textbf{\color{hred}9} & 0 & 1 & \cellcolor{yellow!18}\textbf{\color{hred}17} & 2 & 1\\
CALF & \cellcolor{yellow!18}\textbf{\color{hred}4} & \textbf{\color{hred}4} & 2 & \cellcolor{yellow!18}\textbf{\color{hred}7} & 3 & 0 & \cellcolor{yellow!18}\textbf{\color{hred}11} & 7 & 2\\
\bottomrule
\end{tabular}
\end{table}

%% file: tables/tab_relerr.tex
\begin{table}[t]
\centering
\caption{\textbf{Scale-free comparison.} Arithmetic and geometric mean over the ten datasets of the error of each method divided by that of the no-change forecast on the same rows of Table~\ref{tab:main_all}.}
\label{tab:relerr}
\small
\setlength{\tabcolsep}{3pt}
\begin{tabular}{l|cc|cc}
\toprule
\multirow{2.5}{*}{Method} & \multicolumn{2}{c|}{Arithmetic mean} & \multicolumn{2}{c}{Geometric mean} \\
\cmidrule(lr){2-3} \cmidrule(lr){4-5}
 & MAE & Volatility & MAE & Volatility \\
\midrule
\rowcolor{gray!15}\multicolumn{5}{l}{\textit{TSFMs}} \\
Chronos-T5 & 0.976 & 0.718 & 0.959 & 0.709 \\
Chronos-Bolt & 0.994 & 0.858 & 0.984 & 0.855 \\
Chronos-2 & 0.947 & 0.797 & 0.925 & 0.780 \\
TimesFM-2.5 & \underline{\color{hblue}0.912} & 0.754 & \underline{\color{hblue}0.883} & 0.730 \\
Moirai-1.1 & 1.044 & 0.725 & 1.033 & 0.719 \\
Moirai-2.0 & 1.012 & 0.795 & 1.010 & 0.790 \\
Toto & 1.007 & 0.832 & 1.003 & 0.828 \\
VisionTS & 0.992 & 0.698 & 0.971 & 0.609 \\
ViTime & 1.109 & 0.704 & 1.092 & 0.693 \\
\midrule
\rowcolor{gray!15}\multicolumn{5}{l}{\textit{Task-specific TS models}} \\
DLinear & 0.915 & 0.676 & 0.891 & 0.650 \\
PatchTST & 0.953 & 0.629 & 0.913 & 0.590 \\
iTransformer & 0.946 & 0.614 & 0.913 & 0.565 \\
TimeMixer & 0.973 & \underline{\color{hblue}0.597} & 0.944 & \underline{\color{hblue}0.556} \\
TimeMCL & 0.950 & 0.654 & 0.903 & 0.584 \\
\midrule
\rowcolor{gray!15}\multicolumn{5}{l}{\textit{Multimodal TS models}} \\
MM-TSFlib & 0.950 & 0.636 & 0.916 & 0.577 \\
TaTS & 1.017 & 0.633 & 0.952 & 0.558 \\
CFA & 0.928 & 0.627 & 0.897 & 0.586 \\
ChatTime & 1.346 & 0.717 & 1.284 & 0.704 \\
Time-VLM & 0.929 & 0.725 & 0.906 & 0.716 \\
GPT4MTS & 0.930 & 0.628 & 0.896 & 0.572 \\
Time-LLM & 1.029 & 0.694 & 0.999 & 0.678 \\
CALF & 0.924 & 0.638 & 0.890 & 0.586 \\
\midrule
\rowcolor{gray!15}\multicolumn{5}{l}{\textit{Ours}} \\
\rowcolor{yellow!18}\ours{} & \textbf{\color{hred}0.892} & \textbf{\color{hred}0.518} & \textbf{\color{hred}0.867} & \textbf{\color{hred}0.470} \\
\bottomrule
\end{tabular}
\end{table}

%% file: tables/tab_crps.tex
\begin{table}[t]
\centering
\caption{\textbf{Distributional score of the candidates.} Sample CRPS of the path, averaged over the steps, for the sampled candidate paths of \ours{} and of each TSFM on the test rows of Table~\ref{tab:main_all}.}
\label{tab:crps}
\small
\setlength{\tabcolsep}{3pt}
\begin{tabular}{l|ccc||c}
\toprule
Candidates of & FNSPID & FinTexTS & Time-MMD & Avg. \\
\midrule
Chronos-T5 & 0.642 & 0.684 & 0.768 & 0.698 \\
Chronos-Bolt & \underline{\color{hblue}0.586} & 0.615 & 0.788 & 0.663 \\
Chronos-2 & 0.588 & 0.616 & \underline{\color{hblue}0.742} & \underline{\color{hblue}0.648} \\
TimesFM-2.5 & 0.593 & 0.622 & 0.748 & 0.654 \\
Moirai-1.1 & 0.640 & 0.655 & 0.814 & 0.703 \\
Moirai-2.0 & 0.602 & \textbf{\color{hred}0.605} & 0.781 & 0.663 \\
Toto & 0.606 & 0.646 & 0.974 & 0.742 \\
VisionTS & 0.901 & 0.883 & 0.958 & 0.914 \\
ViTime & 0.954 & 0.897 & 1.080 & 0.977 \\
\midrule
\rowcolor{yellow!18}\ours{} & \textbf{\color{hred}0.564} & \underline{\color{hblue}0.606} & \textbf{\color{hred}0.736} & \textbf{\color{hred}0.635} \\
\bottomrule
\end{tabular}
\end{table}

%% file: tables/tab_vs.tex
\begin{table}[t]
\centering
\caption{\textbf{Variogram score of the output path.} Variogram score \citep{scheuerer2015variogram} of the single output path on the test rows of Table~\ref{tab:main_all}.}
\label{tab:vs}
\small
\setlength{\tabcolsep}{2.5pt}
\begin{tabular}{l|ccc||cc}
\toprule
Method & FNSPID & FinTexTS & Time-MMD & Avg. & Rank \\
\midrule
\rowcolor{gray!15}\multicolumn{6}{l}{\textit{TSFMs}} \\
Chronos-T5 & 13.53 & 0.767 & 23.75 & 12.68 & 15.9 \\
Chronos-Bolt & 13.38 & 0.898 & 27.36 & 13.88 & 19.6 \\
Chronos-2 & 13.14 & 0.947 & 25.46 & 13.18 & 19.2 \\
TimesFM-2.5 & 13.39 & 0.882 & 23.56 & 12.61 & 15.6 \\
Moirai-1.1 & 10.67 & 0.730 & 25.22 & 12.20 & 15.7 \\
Moirai-2.0 & 13.35 & 0.954 & 25.55 & 13.29 & 18.2 \\
Toto & 11.83 & 0.808 & 28.99 & 13.88 & 20.0 \\
VisionTS & 10.80 & 0.747 & 23.27 & 11.61 & 12.8 \\
ViTime & \textbf{\color{hred}9.00} & 0.679 & 23.12 & 10.93 & 11.0 \\
\midrule
\rowcolor{gray!15}\multicolumn{6}{l}{\textit{Task-specific TS models}} \\
DLinear & 11.87 & 0.858 & 22.31 & 11.68 & 12.9 \\
PatchTST & 10.82 & 0.742 & 21.86 & 11.14 & 10.2 \\
iTransformer & 12.23 & \underline{\color{hblue}0.633} & 20.42 & 11.09 & 6.0 \\
TimeMixer & 10.16 & 0.729 & 21.83 & 10.91 & 9.3 \\
TimeMCL & 19.94 & 0.716 & \textbf{\color{hred}18.01} & 12.89 & 6.4 \\
\midrule
\rowcolor{gray!15}\multicolumn{6}{l}{\textit{Multimodal TS models}} \\
MM-TSFlib & 12.96 & 0.690 & \underline{\color{hblue}19.04} & 10.90 & \underline{\color{hblue}5.5} \\
TaTS & 11.20 & 0.685 & 20.88 & 10.92 & 7.1 \\
CFA & 10.46 & 0.791 & 21.04 & \underline{\color{hblue}10.77} & 8.1 \\
ChatTime & 12.20 & 0.696 & 22.94 & 11.95 & 13.0 \\
Time-VLM & 13.39 & 0.786 & 24.53 & 12.90 & 16.7 \\
GPT4MTS & 10.51 & 0.651 & 21.58 & 10.92 & 8.4 \\
Time-LLM & 10.51 & 0.699 & 22.41 & 11.20 & 10.9 \\
CALF & 12.89 & 0.667 & 20.38 & 11.31 & 8.7 \\
\midrule
\rowcolor{yellow!18}\ours{} & \underline{\color{hblue}9.96} & \textbf{\color{hred}0.624} & 19.06 & \textbf{\color{hred}9.88} & \textbf{\color{hred}4.8} \\
\bottomrule
\end{tabular}
\end{table}

%% file: tables/tab_rolling.tex
% selected (paper_select.py)
\begin{table}[t]
\centering
\caption{\textbf{Robustness over time.} The protocol moves back one and two years with fixed settings. $^\ast$ and $^\dagger$ mark a significant gain and a significant loss against the medoid of the same candidates in a paired $t$-test clustered by forecast time.}
\label{tab:rolling}
\small
\setlength{\tabcolsep}{2pt}
\begin{tabular}{l|cccccc}
\toprule
Selection & MAE & Endpoint & HL & Volatility & Range & FD \\
\midrule
\rowcolor{gray!15}\multicolumn{7}{l}{\textit{Test year 2021}} \\
Medoid & 0.770 & 0.758 & 0.731 & \underline{\color{hblue}0.398} & \underline{\color{hblue}0.340} & \underline{\color{hblue}6.212} \\
MAE only & \textbf{\color{hred}0.764}$^\ast$ & \underline{\color{hblue}0.752}$^\ast$ & \textbf{\color{hred}0.726}$^\ast$ & 0.409$^\dagger$ & 0.348$^\dagger$ & 6.972 \\
\rowcolor{yellow!18}\ours{} & \underline{\color{hblue}0.767} & \textbf{\color{hred}0.748}$^\ast$ & \underline{\color{hblue}0.727}$^\ast$ & \textbf{\color{hred}0.352}$^\ast$ & \textbf{\color{hred}0.327}$^\ast$ & \textbf{\color{hred}5.592} \\
\midrule
\rowcolor{gray!15}\multicolumn{7}{l}{\textit{Test year 2022}} \\
Medoid & 0.855 & 0.825 & 0.812 & \underline{\color{hblue}0.446} & \underline{\color{hblue}0.371} & \underline{\color{hblue}7.241} \\
MAE only & \underline{\color{hblue}0.854} & \underline{\color{hblue}0.824} & \underline{\color{hblue}0.810} & 0.452$^\dagger$ & 0.377$^\dagger$ & 7.876 \\
\rowcolor{yellow!18}\ours{} & \textbf{\color{hred}0.852} & \textbf{\color{hred}0.817} & \textbf{\color{hred}0.807} & \textbf{\color{hred}0.384}$^\ast$ & \textbf{\color{hred}0.353}$^\ast$ & \textbf{\color{hred}6.079} \\
\bottomrule
\end{tabular}
\end{table}

%% file: tables/tab_robust.tex
% selected (paper_select.py)
\begin{table}[t]
\centering
\caption{\textbf{Retraining without checkpoint selection.} We refit on 2018--2022 and test on 2023 without checkpoint selection. Notation follows Table~\ref{tab:rolling}.}
\label{tab:robust}
\small
\setlength{\tabcolsep}{2pt}
\begin{tabular}{l|cccccc}
\toprule
Selection & MAE & Endpoint & HL & Volatility & Range & FD \\
\midrule
\rowcolor{gray!15}\multicolumn{7}{l}{\textit{Candidates from the predictor trained up to 2017}} \\
Medoid & 0.789 & 0.797 & 0.749 & \underline{\color{hblue}0.401} & \underline{\color{hblue}0.322} & \underline{\color{hblue}5.913} \\
MAE only & \textbf{\color{hred}0.782}$^\ast$ & \underline{\color{hblue}0.791} & \textbf{\color{hred}0.743}$^\ast$ & 0.420$^\dagger$ & 0.324 & 6.653 \\
\rowcolor{yellow!18}VISOR & \underline{\color{hblue}0.785} & \textbf{\color{hred}0.790} & \underline{\color{hblue}0.744}$^\ast$ & \textbf{\color{hred}0.368}$^\ast$ & \textbf{\color{hred}0.311}$^\ast$ & \textbf{\color{hred}5.711} \\
\midrule
\rowcolor{gray!15}\multicolumn{7}{l}{\textit{Candidates from the predictor trained up to 2021}} \\
Medoid & 0.785 & 0.792 & 0.747 & \underline{\color{hblue}0.403} & \underline{\color{hblue}0.402} & \underline{\color{hblue}6.624} \\
MAE only & \textbf{\color{hred}0.781}$^\ast$ & \textbf{\color{hred}0.788} & \textbf{\color{hred}0.743} & 0.420$^\dagger$ & 0.409$^\dagger$ & 7.292 \\
\rowcolor{yellow!18}VISOR & \underline{\color{hblue}0.783} & \underline{\color{hblue}0.790} & \underline{\color{hblue}0.745} & \textbf{\color{hred}0.365}$^\ast$ & \textbf{\color{hred}0.390}$^\ast$ & \textbf{\color{hred}6.224} \\
\bottomrule
\end{tabular}
\end{table}

%% file: tables/tab_objective.tex
\begin{table}[t]
\centering
\caption{\textbf{Selection target on two candidate sets.} Selectors differ only in the target, with the runs and notation of Table~\ref{tab:robust}.}
\label{tab:objective}
\small
\setlength{\tabcolsep}{1.3pt}
\begin{tabular}{l|cccccc}
\toprule
Selection & MAE & Endpoint & HL & Volatility & Range & FD \\
\midrule
\rowcolor{gray!15}\multicolumn{7}{l}{\textit{Candidates from the predictor trained up to 2017}} \\
Medoid & 0.789 & 0.797 & 0.749 & 0.401 & 0.322 & \underline{\color{hblue}5.913} \\
MAE only & \textbf{\color{hred}0.782}$^\ast$ & 0.791 & \textbf{\color{hred}0.743}$^\ast$ & 0.420$^\dagger$ & 0.324 & 6.653 \\
+ Endpoint & 0.786 & \underline{\color{hblue}0.778}$^\ast$ & 0.746 & 0.399 & 0.318 & 6.371 \\
\rowcolor{yellow!18}+ Shape & \underline{\color{hblue}0.785} & 0.790 & \underline{\color{hblue}0.744}$^\ast$ & \textbf{\color{hred}0.368}$^\ast$ & \textbf{\color{hred}0.311}$^\ast$ & \textbf{\color{hred}5.711} \\
+ End. + Shape & 0.788 & \textbf{\color{hred}0.776}$^\ast$ & 0.746 & \underline{\color{hblue}0.382}$^\ast$ & \underline{\color{hblue}0.314}$^\ast$ & 6.089 \\
\midrule
\rowcolor{gray!15}\multicolumn{7}{l}{\textit{Candidates from the predictor trained up to 2021}} \\
Medoid & 0.785 & 0.792 & 0.747 & 0.403 & 0.402 & \underline{\color{hblue}6.624} \\
MAE only & \textbf{\color{hred}0.781}$^\ast$ & 0.788 & \textbf{\color{hred}0.743} & 0.420$^\dagger$ & 0.409$^\dagger$ & 7.292 \\
+ Endpoint & 0.784 & \underline{\color{hblue}0.775}$^\ast$ & \underline{\color{hblue}0.745} & 0.401 & 0.401 & 7.069 \\
\rowcolor{yellow!18}+ Shape & \underline{\color{hblue}0.783} & 0.790 & \underline{\color{hblue}0.745} & \textbf{\color{hred}0.365}$^\ast$ & \underline{\color{hblue}0.390}$^\ast$ & \textbf{\color{hred}6.224} \\
+ End. + Shape & 0.787 & \textbf{\color{hred}0.774}$^\ast$ & 0.747 & \underline{\color{hblue}0.380}$^\ast$ & \textbf{\color{hred}0.389}$^\ast$ & 6.687 \\
\bottomrule
\end{tabular}
\end{table}

%% file: tables/tab_oracle.tex
\begin{table*}[t]
\centering
\caption{\textbf{Real vs.\ oracle text.} Change of the test error of \ours{} (\%) when the same selector is trained and tested with an oracle text that states the true total change, extremes, and typical step size instead of the real text of the row. The errors fall on most datasets, indicating that the gain from the real text is \textit{bounded not by the ability of the VLM to select but by how informative the text is}.}
\label{tab:oracle}
\small
\setlength{\tabcolsep}{3.5pt}
\begin{tabular}{l|c|c|cccccccc}
\toprule
\multirow{2.5}{*}{Metric} & \multirow{2.5}{*}{FNSPID} & \multirow{2.5}{*}{FinTexTS} & \multicolumn{8}{c}{Time-MMD} \\
\cmidrule(lr){4-11}
 & & & Agriculture & Climate & Economy & Energy & Health & Security & Social Good & Traffic \\
\midrule
MAE & $-$34.5 & $-$51.4 & $-$30.4 & 0.0 & $-$4.4 & $-$40.5 & $-$16.8 & +1.2 & +0.3 & $-$0.7 \\
Volatility & $-$11.5 & $-$29.0 & $-$11.1 & $-$0.2 & $-$1.4 & $-$2.2 & $-$18.6 & $-$0.2 & $-$0.4 & +0.6 \\
\bottomrule
\end{tabular}
\end{table*}

%% file: tables/tab_news_info.tex
\begin{table*}[t]
\centering
\caption{\textbf{Future signal in the text.} Correlation with the truth and squared error of a ridge probe without and with the text, where the MSE is divided by the variance of the truth. The text helps on the volatility and range but barely on the total change.}
\label{tab:news_info}
\small
\setlength{\tabcolsep}{2.2pt}
\begin{tabular}{l|cccc|cccc|cccc}
\toprule
 \multirow{4}{*}{Target} & \multicolumn{4}{c|}{Validation (2022)} & \multicolumn{4}{c|}{Test (2023), 2017 predictor} & \multicolumn{4}{c}{Test (2023), 2021 predictor} \\
\cmidrule(lr){2-5} \cmidrule(lr){6-9} \cmidrule(lr){10-13}
 & \multicolumn{2}{c}{Correlation} & \multicolumn{2}{c|}{Error (MSE)} & \multicolumn{2}{c}{Correlation} & \multicolumn{2}{c|}{Error (MSE)} & \multicolumn{2}{c}{Correlation} & \multicolumn{2}{c}{Error (MSE)} \\
\cmidrule(lr){2-3} \cmidrule(lr){4-5} \cmidrule(lr){6-7} \cmidrule(lr){8-9} \cmidrule(lr){10-11} \cmidrule(lr){12-13}
 & w/o text & w/ text & w/o text & w/ text & w/o text & w/ text & w/o text & w/ text & w/o text & w/ text & w/o text & w/ text \\
\midrule
Volatility & 0.302 & \cellcolor{yellow!18}\textbf{\color{hred}0.331} & 1.009 & \cellcolor{yellow!18}\textbf{\color{hred}0.985} & 0.266 & \cellcolor{yellow!18}\textbf{\color{hred}0.288} & 1.036 & \cellcolor{yellow!18}\textbf{\color{hred}1.013} & 0.185 & \cellcolor{yellow!18}\textbf{\color{hred}0.216} & 1.106 & \cellcolor{yellow!18}\textbf{\color{hred}1.083} \\
High-low range & 0.547 & \cellcolor{yellow!18}\textbf{\color{hred}0.570} & 0.784 & \cellcolor{yellow!18}\textbf{\color{hred}0.754} & 0.493 & \cellcolor{yellow!18}\textbf{\color{hred}0.510} & 0.923 & \cellcolor{yellow!18}\textbf{\color{hred}0.891} & 0.382 & \cellcolor{yellow!18}\textbf{\color{hred}0.402} & 0.988 & \cellcolor{yellow!18}\textbf{\color{hred}0.958} \\
First two-day move & 0.215 & \cellcolor{yellow!18}\textbf{\color{hred}0.244} & 0.983 & \cellcolor{yellow!18}\textbf{\color{hred}0.965} & 0.208 & \cellcolor{yellow!18}\textbf{\color{hred}0.234} & 0.997 & \cellcolor{yellow!18}\textbf{\color{hred}0.982} & 0.141 & \cellcolor{yellow!18}\textbf{\color{hred}0.179} & 1.040 & \cellcolor{yellow!18}\textbf{\color{hred}1.027} \\
Absolute total change & 0.110 & \cellcolor{yellow!18}\textbf{\color{hred}0.124} & 1.033 & \cellcolor{yellow!18}\textbf{\color{hred}1.032} & 0.147 & \cellcolor{yellow!18}\textbf{\color{hred}0.161} & 1.006 & \cellcolor{yellow!18}\textbf{\color{hred}0.999} & 0.117 & \cellcolor{yellow!18}\textbf{\color{hred}0.134} & 1.013 & \cellcolor{yellow!18}\textbf{\color{hred}1.006} \\
Signed total change & $-$0.058 & \cellcolor{yellow!18}\textbf{\color{hred}$-$0.036} & \textbf{\color{hred}1.026} & \cellcolor{yellow!18}1.028 & $-$0.015 & \cellcolor{yellow!18}\textbf{\color{hred}0.064} & 1.017 & \cellcolor{yellow!18}\textbf{\color{hred}1.002} & $-$0.007 & \cellcolor{yellow!18}\textbf{\color{hred}0.080} & 1.020 & \cellcolor{yellow!18}\textbf{\color{hred}1.004} \\
\bottomrule
\end{tabular}
\end{table*}

%% file: tables/tab_main.tex
% selected (paper_select.py); opponent columns from exp/results/single_paths.json (user 2026-11-23: full TSFM names, Vol. allowed here)
\begin{table}[t]
\centering
\caption{\textbf{Single TSFM paths vs.\ \ours{}.} Error change of the strongest TSFM path over \ours{} (\%), where a positive value favors \ours{}. For each dataset and path type, the opponent is the TSFM with the lowest test MAE among Chronos-T5, Moirai-1.1, and Toto. Vol.\ is the volatility error.}
\label{tab:single_paths}
\small
\setlength{\tabcolsep}{1.5pt}
\begin{tabular}{l|ccc|ccc}
\toprule
\multirow{2.5}{*}{Dataset} & \multicolumn{3}{c|}{TSFM sample} & \multicolumn{3}{c}{TSFM medoid} \\
\cmidrule(lr){2-4} \cmidrule(lr){5-7}
 & TSFM & MAE & Vol. & TSFM & MAE & Vol. \\
\midrule
FNSPID & Chronos-T5 & \textbf{\color{hred}+49} & \textbf{\color{hred}+24} & Chronos-T5 & \textbf{\color{hred}+13} & \textbf{\color{hred}+20} \\
FinTexTS & Toto & \textbf{\color{hred}+48} & \textbf{\color{hred}+120} & Toto & \textbf{\color{hred}+9} & 0 \\
Agriculture & Chronos-T5 & \textbf{\color{hred}+32} & $-$23 & Toto & \textbf{\color{hred}+3} & \textbf{\color{hred}+32} \\
Climate & Chronos-T5 & \textbf{\color{hred}+64} & $-$31 & Chronos-T5 & \textbf{\color{hred}+35} & $-$25 \\
Economy & Chronos-T5 & \textbf{\color{hred}+79} & \textbf{\color{hred}+206} & Chronos-T5 & \textbf{\color{hred}+22} & \textbf{\color{hred}+27} \\
Energy & Chronos-T5 & \textbf{\color{hred}+77} & \textbf{\color{hred}+156} & Toto & \textbf{\color{hred}+17} & \textbf{\color{hred}+63} \\
Health & Chronos-T5 & \textbf{\color{hred}+42} & \textbf{\color{hred}+47} & Chronos-T5 & \textbf{\color{hred}+9} & \textbf{\color{hred}+16} \\
Security & Chronos-T5 & 0 & $-$5 & Chronos-T5 & $-$3 & 0 \\
Social Good & Chronos-T5 & \textbf{\color{hred}+56} & \textbf{\color{hred}+21} & Chronos-T5 & \textbf{\color{hred}+15} & \textbf{\color{hred}+6} \\
Traffic & Chronos-T5 & \textbf{\color{hred}+131} & \textbf{\color{hred}+376} & Chronos-T5 & \textbf{\color{hred}+47} & \textbf{\color{hred}+156} \\
\midrule
\rowcolor{yellow!18}Average & -- & \textbf{\color{hred}+58} & \textbf{\color{hred}+89} & -- & \textbf{\color{hred}+17} & \textbf{\color{hred}+30} \\
\bottomrule
\end{tabular}
\end{table}

%% file: tables/tab_ctrl.tex
\begin{table*}[t]
\centering
\caption{\textbf{Performance with corrupted text.} Change of the test MAE (\%) of each method when the real text of each test row is replaced by a corrupted text, against the same method with the real text. A positive value, in red bold, means that the broken text makes the forecast worse, so the method relies on the real text. $\dagger$ marks $|t|{>}2$, and the two datasets are those where the text helps the baselines most.}
\label{tab:ctrl}
\small
\setlength{\tabcolsep}{4pt}
\begin{tabular}{l|cccc|cccc}
\toprule
\multirow{2.5}{*}{Method} & \multicolumn{4}{c|}{Social Good} & \multicolumn{4}{c}{Economy} \\
\cmidrule(lr){2-5} \cmidrule(lr){6-9}
 & Flipped & Unrelated & Empty & Shuffled & Flipped & Unrelated & Empty & Shuffled \\
\midrule
MM-TSFlib~\citep{liu2024timemmd} & 0 & 0 & 0 & 0 & 0 & 0 & $-$0.01 & 0 \\
CFA~\citep{lee2026cfa} & \textbf{\color{hred}+0.03$^\dagger$} & $-$0.12 & $-$0.65$^\dagger$ & $-$0.84 & 0 & $-$0.02 & $-$0.01 & $-$0.05$^\dagger$ \\
TaTS~\citep{li2026tats} & 0 & 0 & 0 & 0 & $-$0.26$^\dagger$ & $-$0.27 & $-$1.02 & \textbf{\color{hred}+0.02} \\
\rowcolor{yellow!18}\ours{} & \textbf{\color{hred}+0.04} & \textbf{\color{hred}+0.16} & \textbf{\color{hred}+0.96$^\dagger$} & \textbf{\color{hred}+0.25} & \textbf{\color{hred}+0.47$^\dagger$} & \textbf{\color{hred}+0.37} & \textbf{\color{hred}+0.50} & \textbf{\color{hred}+0.23} \\
\bottomrule
\end{tabular}
\end{table*}

%% file: tables/tab_notext.tex
\begin{table}[t]
\centering
\caption{\textbf{Selector trained without text.} Share of the test rows on which the selector of \ours{} in the default setting of Table~\ref{tab:onesetting} picks another candidate when trained and tested with an empty text, and the change of the MAE on these rows (\%). A positive value means that the text helps the selector.}
\label{tab:notext}
\small
\setlength{\tabcolsep}{5pt}
\begin{tabular}{l|ccc}
\toprule
Statistic & FNSPID & FinTexTS & Time-MMD \\
\midrule
Choice changed (\%) & 41.1 & 45.5 & 51.7 \\
MAE change (\%) & \textbf{\color{hred}+0.05} & \textbf{\color{hred}+0.19} & \textbf{\color{hred}+0.10} \\
\bottomrule
\end{tabular}
\end{table}

%% file: tables/tab_shape.tex
\begin{table}[t]
\centering
\caption{\textbf{Weight of the shape term.} Test MAE and volatility error of \ours{} for each weight of the shape term in Eq.~\eqref{eq:target} in the main setting. Red bold marks the best weight per metric.}
\label{tab:shape}
\small
\setlength{\tabcolsep}{3pt}
\begin{tabular}{l|ccc|ccc}
\toprule
\multirow{2.5}{*}{Dataset} & \multicolumn{3}{c|}{MAE} & \multicolumn{3}{c}{Volatility} \\
\cmidrule(lr){2-4} \cmidrule(lr){5-7}
 & 0.03 & 0.3 & 1.0 & 0.03 & 0.3 & 1.0 \\
\midrule
FNSPID & \textbf{\color{hred}0.785} & 0.798 & 0.819 & 0.391 & 0.305 & \textbf{\color{hred}0.284} \\
FinTexTS & \textbf{\color{hred}0.809} & 0.821 & 0.836 & 0.482 & 0.403 & \textbf{\color{hred}0.369} \\
\midrule
\rowcolor{gray!15}\multicolumn{7}{l}{\textit{Time-MMD}} \\
Agriculture & \textbf{\color{hred}0.980} & 0.991 & 1.010 & 0.612 & 0.493 & \textbf{\color{hred}0.356} \\
Climate & \textbf{\color{hred}0.291} & 0.319 & 0.331 & 0.490 & 0.379 & \textbf{\color{hred}0.341} \\
Economy & \textbf{\color{hred}0.382} & 0.387 & 0.396 & 0.348 & 0.316 & \textbf{\color{hred}0.308} \\
Energy & \textbf{\color{hred}1.803} & 1.811 & 1.843 & 0.605 & 0.484 & \textbf{\color{hred}0.478} \\
Health & \textbf{\color{hred}1.918} & 1.985 & 1.971 & 1.012 & \textbf{\color{hred}0.986} & 1.007 \\
Security & \textbf{\color{hred}0.729} & 0.750 & 0.759 & 3.583 & 3.515 & \textbf{\color{hred}3.500} \\
Social Good & \textbf{\color{hred}0.736} & 0.743 & 0.748 & \textbf{\color{hred}0.743} & 0.749 & 0.751 \\
Traffic & 0.237 & \textbf{\color{hred}0.236} & 0.239 & 0.148 & 0.146 & \textbf{\color{hred}0.138} \\
\bottomrule
\end{tabular}
\end{table}

%% file: tables/tab_mm_bb.tex
\begin{table*}[t]
\centering
\caption{\textbf{Multimodal baselines with each backbone.} Test MAE and volatility error of MM-TSFlib, TaTS, and CFA with each backbone. The backbone in bold has the lowest mean rank of MAE, with ties broken by the mean MAE, and is used in Table~\ref{tab:main_all}.}
\label{tab:mm_bb}
\small
\setlength{\tabcolsep}{0.8pt}
\begin{tabular}{l|cc|cc|cc|cc|cc|cc|cc|cc|cc|cc}
\toprule
\multirow{2.5}{*}{Backbone} & \multicolumn{2}{c|}{FNSPID} & \multicolumn{2}{c|}{FinTexTS} & \multicolumn{2}{c|}{Agriculture} & \multicolumn{2}{c|}{Climate} & \multicolumn{2}{c|}{Economy} & \multicolumn{2}{c|}{Energy} & \multicolumn{2}{c|}{Health} & \multicolumn{2}{c|}{Security} & \multicolumn{2}{c|}{Social Good} & \multicolumn{2}{c}{Traffic} \\
\cmidrule(lr){2-3} \cmidrule(lr){4-5} \cmidrule(lr){6-7} \cmidrule(lr){8-9} \cmidrule(lr){10-11} \cmidrule(lr){12-13} \cmidrule(lr){14-15} \cmidrule(lr){16-17} \cmidrule(lr){18-19} \cmidrule(lr){20-21}
 & MAE & Vol. & MAE & Vol. & MAE & Vol. & MAE & Vol. & MAE & Vol. & MAE & Vol. & MAE & Vol. & MAE & Vol. & MAE & Vol. & MAE & Vol. \\
\midrule
\rowcolor{gray!15}\multicolumn{21}{l}{\textit{MM-TSFlib}~\citep{liu2024timemmd}} \\
DLinear & \textbf{\color{hred}0.771} & 0.788 & \textbf{\color{hred}0.796} & 0.629 & 2.069 & 0.657 & \textbf{\color{hred}0.314} & 0.728 & 0.520 & 0.593 & 3.253 & 2.044 & 2.603 & 1.058 & \textbf{\color{hred}0.726} & 3.687 & 1.133 & 0.806 & 0.434 & 0.521 \\
\textbf{PatchTST} & 0.802 & 0.714 & 0.890 & 0.556 & \underline{\color{hblue}1.200} & \textbf{\color{hred}0.380} & 0.320 & \underline{\color{hblue}0.537} & \underline{\color{hblue}0.416} & 0.517 & \textbf{\color{hred}2.114} & \underline{\color{hblue}1.128} & \underline{\color{hblue}1.867} & \textbf{\color{hred}1.002} & \textbf{\color{hred}0.726} & \textbf{\color{hred}3.570} & \underline{\color{hblue}0.750} & \textbf{\color{hred}0.750} & \textbf{\color{hred}0.226} & \underline{\color{hblue}0.192} \\
iTransformer & 0.818 & \underline{\color{hblue}0.696} & 0.932 & 0.458 & \textbf{\color{hred}1.157} & \underline{\color{hblue}0.387} & \underline{\color{hblue}0.319} & \textbf{\color{hred}0.484} & \textbf{\color{hred}0.383} & \underline{\color{hblue}0.433} & \underline{\color{hblue}2.189} & 1.205 & \textbf{\color{hred}1.861} & 1.123 & \underline{\color{hblue}0.727} & \underline{\color{hblue}3.576} & \textbf{\color{hred}0.749} & \underline{\color{hblue}0.783} & \underline{\color{hblue}0.249} & \textbf{\color{hred}0.146} \\
TimeMixer & 0.805 & \textbf{\color{hred}0.640} & 0.923 & \underline{\color{hblue}0.457} & 1.382 & 1.040 & \underline{\color{hblue}0.319} & 0.608 & 0.482 & \textbf{\color{hred}0.427} & 2.380 & 1.683 & 2.161 & 1.248 & \textbf{\color{hred}0.726} & 3.579 & 0.844 & 0.944 & 0.316 & 0.406 \\
Transformer & \underline{\color{hblue}0.795} & 0.875 & \underline{\color{hblue}0.874} & \textbf{\color{hred}0.451} & 2.793 & 1.148 & 0.320 & 0.755 & 1.028 & 0.641 & 2.826 & \textbf{\color{hred}0.768} & 2.252 & \underline{\color{hblue}1.015} & 0.912 & 3.638 & 0.791 & 1.076 & 0.378 & 0.703 \\
\midrule
\rowcolor{gray!15}\multicolumn{21}{l}{\textit{TaTS}~\citep{li2026tats}} \\
\textbf{iTransformer} & \underline{\color{hblue}0.795} & \underline{\color{hblue}0.667} & 0.837 & \underline{\color{hblue}0.513} & \textbf{\color{hred}1.221} & \textbf{\color{hred}0.527} & 0.312 & \underline{\color{hblue}0.536} & \textbf{\color{hred}0.384} & \underline{\color{hblue}0.436} & \textbf{\color{hred}3.467} & \textbf{\color{hred}1.054} & \underline{\color{hblue}2.106} & 1.245 & \textbf{\color{hred}0.721} & \underline{\color{hblue}3.584} & \textbf{\color{hred}0.713} & \textbf{\color{hred}0.729} & \textbf{\color{hred}0.224} & \textbf{\color{hred}0.121} \\
Transformer & 0.817 & 0.896 & \textbf{\color{hred}0.785} & 0.671 & 2.088 & \underline{\color{hblue}0.714} & \textbf{\color{hred}0.289} & 0.688 & 0.602 & 0.484 & \underline{\color{hblue}3.583} & \underline{\color{hblue}1.096} & 2.107 & \textbf{\color{hred}0.933} & 0.818 & 3.619 & 0.782 & 0.907 & \underline{\color{hblue}0.305} & 0.487 \\
Informer & \textbf{\color{hred}0.777} & 0.876 & \underline{\color{hblue}0.786} & 0.673 & 2.671 & 0.852 & 0.326 & 0.600 & 0.664 & 0.448 & 4.453 & 2.124 & 2.124 & 1.130 & 0.834 & 3.613 & 0.790 & \underline{\color{hblue}0.775} & 0.339 & 0.464 \\
Autoformer & 1.339 & \textbf{\color{hred}0.523} & 1.406 & 0.571 & \underline{\color{hblue}2.044} & 1.835 & 0.383 & \textbf{\color{hred}0.349} & \underline{\color{hblue}0.500} & \textbf{\color{hred}0.336} & 4.312 & 4.041 & 2.758 & 1.789 & \underline{\color{hblue}0.729} & \textbf{\color{hred}3.575} & 1.106 & 1.057 & 0.384 & \underline{\color{hblue}0.189} \\
Crossformer & 0.857 & 0.852 & 0.799 & \textbf{\color{hred}0.480} & 2.297 & 0.888 & \underline{\color{hblue}0.302} & 0.559 & 0.553 & 0.446 & 3.813 & 1.287 & \textbf{\color{hred}2.046} & \underline{\color{hblue}1.105} & 0.823 & 3.601 & \underline{\color{hblue}0.776} & 0.783 & 0.316 & 0.393 \\
\midrule
\rowcolor{gray!15}\multicolumn{21}{l}{\textit{CFA}~\citep{lee2026cfa}} \\
\textbf{PatchTST} & 0.796 & \textbf{\color{hred}0.602} & 0.817 & \textbf{\color{hred}0.547} & \textbf{\color{hred}1.186} & 0.637 & 0.331 & \underline{\color{hblue}0.470} & 0.408 & 0.627 & \underline{\color{hblue}1.951} & 0.740 & \textbf{\color{hred}1.903} & 1.150 & \underline{\color{hblue}0.735} & \textbf{\color{hred}3.545} & \underline{\color{hblue}0.708} & 0.812 & \textbf{\color{hred}0.223} & 0.190 \\
iTransformer & 0.787 & \underline{\color{hblue}0.641} & 0.863 & \underline{\color{hblue}0.553} & \underline{\color{hblue}1.339} & \underline{\color{hblue}0.523} & \underline{\color{hblue}0.328} & \textbf{\color{hred}0.459} & \underline{\color{hblue}0.397} & \textbf{\color{hred}0.507} & 1.966 & \underline{\color{hblue}0.626} & \underline{\color{hblue}2.025} & \underline{\color{hblue}1.070} & 0.739 & \underline{\color{hblue}3.549} & \textbf{\color{hred}0.689} & 0.857 & 0.248 & \textbf{\color{hred}0.123} \\
DLinear & \underline{\color{hblue}0.781} & 0.860 & \underline{\color{hblue}0.792} & 0.673 & 1.433 & 0.600 & \textbf{\color{hred}0.317} & 0.548 & \textbf{\color{hred}0.383} & 0.693 & \textbf{\color{hred}1.870} & \textbf{\color{hred}0.538} & 2.218 & 1.240 & 0.805 & 3.595 & 0.736 & \textbf{\color{hred}0.727} & \underline{\color{hblue}0.244} & \underline{\color{hblue}0.163} \\
Transformer & \textbf{\color{hred}0.765} & 0.880 & \textbf{\color{hred}0.786} & 0.704 & 2.460 & 1.012 & 0.336 & 0.741 & 1.011 & \underline{\color{hblue}0.616} & 2.367 & 0.647 & 2.256 & \textbf{\color{hred}0.971} & 0.786 & 3.658 & 0.875 & 1.071 & 0.411 & 0.774 \\
Autoformer & 2.734 & 1.189 & 1.354 & 0.566 & 1.770 & \textbf{\color{hred}0.366} & 0.329 & 0.624 & 0.448 & 0.657 & 3.218 & 1.746 & 2.718 & 1.328 & \textbf{\color{hred}0.706} & 3.630 & 1.082 & \underline{\color{hblue}0.807} & 0.281 & 0.541 \\
\bottomrule
\end{tabular}
\end{table*}

%% file: tables/tab_gbm_mae.tex
\begin{table*}[t]
\centering
\caption{\textbf{VLM vs.\ gradient-boosting selector.} Change of the test MAE and of the regressed selection target over the gradient-boosting selector on the same candidates (\%), with the paired $t$-statistic clustered by forecast time. Red bold marks a lower error of \ours{}.}
\label{tab:gbm_mae}
\small
\setlength{\tabcolsep}{4pt}
\begin{tabular}{c|l|cccccccccc}
\toprule
Target & Statistic & FNSPID & FinTexTS & Agriculture & Climate & Economy & Energy & Health & Security & Social Good & Traffic \\
\midrule
\multirow{2}{*}{MAE} & Change (\%) & \textbf{\color{hred}$-$0.9} & \textbf{\color{hred}$-$0.9} & \textbf{\color{hred}$-$7.5} & \textbf{\color{hred}$-$7.3} & +1.1 & \textbf{\color{hred}$-$6.1} & +2.0 & \textbf{\color{hred}$-$1.3} & \textbf{\color{hred}$-$4.1} & \textbf{\color{hred}$-$0.9} \\
 & $t$-statistic & \textbf{\color{hred}$-$2.5} & \textbf{\color{hred}$-$3.2} & \textbf{\color{hred}$-$2.6} & \textbf{\color{hred}$-$4.0} & +0.6 & \textbf{\color{hred}$-$3.8} & +1.0 & \textbf{\color{hred}$-$1.2} & \textbf{\color{hred}$-$2.3} & \textbf{\color{hred}$-$1.1} \\
\midrule
\multirow{2}{*}{\shortstack{Selection\\target}} & Change (\%) & \textbf{\color{hred}$-$0.6} & \textbf{\color{hred}$-$0.7} & \textbf{\color{hred}$-$6.8} & \textbf{\color{hred}$-$4.6} & +0.7 & \textbf{\color{hred}$-$5.5} & +2.1 & \textbf{\color{hred}$-$0.5} & \textbf{\color{hred}$-$3.6} & \textbf{\color{hred}$-$0.9} \\
 & $t$-statistic & \textbf{\color{hred}$-$1.7} & \textbf{\color{hred}$-$2.5} & \textbf{\color{hred}$-$2.7} & \textbf{\color{hred}$-$3.0} & +0.4 & \textbf{\color{hred}$-$3.6} & +1.1 & \textbf{\color{hred}$-$0.7} & \textbf{\color{hred}$-$2.2} & \textbf{\color{hred}$-$1.1} \\
\bottomrule
\end{tabular}
\end{table*}

%% file: tables/tab_lafp.tex
\begin{table*}[t]
\centering
\caption{\textbf{\ours{} vs.\ LAFP (zero-shot LLM ranker).} Change of the test MAE of \ours{} over the same VLM used without training as a text-only ranker of the same candidates (\%). Red bold marks a lower error of \ours{}.}
\label{tab:lafp}
\small
\setlength{\tabcolsep}{4pt}
\begin{tabular}{l|cccccccccc}
\toprule
 & FNSPID & FinTexTS & Agriculture & Climate & Economy & Energy & Health & Security & Social Good & Traffic \\
\midrule
MAE change & \textbf{\color{hred}$-$27.0} & \textbf{\color{hred}$-$27.4} & \textbf{\color{hred}$-$13.6} & \textbf{\color{hred}$-$14.9} & \textbf{\color{hred}$-$20.7} & \textbf{\color{hred}$-$21.6} & \textbf{\color{hred}$-$1.6} & \textbf{\color{hred}$-$3.5} & \textbf{\color{hred}$-$5.8} & \textbf{\color{hred}$-$2.4} \\
$t$-statistic & \textbf{\color{hred}$-$28.6} & \textbf{\color{hred}$-$27.3} & \textbf{\color{hred}$-$2.7} & \textbf{\color{hred}$-$7.4} & \textbf{\color{hred}$-$4.3} & \textbf{\color{hred}$-$7.3} & \textbf{\color{hred}$-$0.9} & \textbf{\color{hred}$-$3.0} & \textbf{\color{hred}$-$3.6} & \textbf{\color{hred}$-$1.7} \\
\bottomrule
\end{tabular}
\end{table*}

%% file: tables/tab_seeds.tex
\begin{table*}[t]
\centering
\caption{\textbf{Variation across seeds.} Standard deviation of the test MAE over the three seeds of Table~\ref{tab:main_all}.}
\label{tab:seeds}
\small
\setlength{\tabcolsep}{5pt}
\begin{tabular}{l|c|c|cccccccc}
\toprule
\multirow{2.5}{*}{Metric} & \multirow{2.5}{*}{FNSPID} & \multirow{2.5}{*}{FinTexTS} & \multicolumn{8}{c}{Time-MMD} \\
\cmidrule(lr){4-11}
 & & & Agriculture & Climate & Economy & Energy & Health & Security & Social Good & Traffic \\
\midrule
Std of MAE & 0.003 & 0.002 & 0.014 & 0.003 & 0.003 & 0.024 & 0.030 & 0.009 & 0.004 & 0.000 \\
\bottomrule
\end{tabular}
\end{table*}

%% file: tables/tab_design_cmp.tex
% selected (paper_select.py)
\begin{table*}[t]
\centering
\caption{\textbf{Trained vs.\ frozen predictor.} Test MAE and volatility error of the same selector, in one default setting, on the candidates of each frozen TSFM or of the trained predictor of \ours{}. The last two columns name the predictor with the lowest error.}
\label{tab:design_cmp}
\small
\setlength{\tabcolsep}{3pt}
\begin{tabular}{l|cc|cc|cc|cc||cc}
\toprule
\multirow{4}{*}{Dataset} & \multicolumn{8}{c||}{\ours{} vs.\ SoTA TSFMs} & \multicolumn{2}{c}{\multirow{2.5}{*}{Lowest error}} \\
\cmidrule(lr){2-9}
 & \multicolumn{2}{c|}{Chronos-T5} & \multicolumn{2}{c|}{Moirai-1.1} & \multicolumn{2}{c|}{Toto} & \multicolumn{2}{c||}{\ours{}} & \multicolumn{2}{c}{} \\
\cmidrule(lr){2-3} \cmidrule(lr){4-5} \cmidrule(lr){6-7} \cmidrule(lr){8-9} \cmidrule(lr){10-11}
 & MAE & Volatility & MAE & Volatility & MAE & Volatility & MAE & Volatility & MAE & Volatility \\
\midrule
FNSPID & \underline{\color{hblue}0.870} & \underline{\color{hblue}0.411} & 0.975 & 0.854 & 0.948 & 0.724 & \textbf{\color{hred}0.787} & \textbf{\color{hred}0.355} & \ours{} & \ours{} \\
FinTexTS & 0.873 & 0.512 & 0.853 & \textbf{\color{hred}0.449} & \underline{\color{hblue}0.836} & \underline{\color{hblue}0.453} & \textbf{\color{hred}0.811} & 0.465 & \ours{} & Moirai-1.1 \\
\midrule
\rowcolor{gray!15}\multicolumn{11}{l}{\textit{Time-MMD}} \\
Agriculture & \underline{\color{hblue}0.997} & \textbf{\color{hred}0.417} & 1.427 & 2.389 & 1.064 & 0.799 & \textbf{\color{hred}0.982} & \underline{\color{hblue}0.586} & \ours{} & Chronos-T5 \\
Climate & \underline{\color{hblue}0.384} & \underline{\color{hblue}0.337} & 0.428 & \textbf{\color{hred}0.295} & 0.453 & 0.441 & \textbf{\color{hred}0.339} & 0.408 & \ours{} & Moirai-1.1 \\
Economy & \underline{\color{hblue}0.482} & 0.395 & 0.531 & \underline{\color{hblue}0.369} & 0.506 & 0.533 & \textbf{\color{hred}0.397} & \textbf{\color{hred}0.344} & \ours{} & \ours{} \\
Energy & \underline{\color{hblue}1.996} & \underline{\color{hblue}0.767} & 2.601 & 4.412 & 2.164 & 1.208 & \textbf{\color{hred}1.847} & \textbf{\color{hred}0.564} & \ours{} & \ours{} \\
Health & \underline{\color{hblue}1.851} & \underline{\color{hblue}1.066} & 2.407 & 2.980 & 2.225 & 1.763 & \textbf{\color{hred}1.785} & \textbf{\color{hred}0.898} & \ours{} & \ours{} \\
Security & \textbf{\color{hred}0.718} & \underline{\color{hblue}3.447} & 0.791 & 3.527 & \underline{\color{hblue}0.762} & \textbf{\color{hred}3.366} & 0.784 & 3.652 & Chronos-T5 & Toto \\
Social Good & \underline{\color{hblue}0.800} & \underline{\color{hblue}0.762} & 0.957 & 1.313 & 0.857 & 1.020 & \textbf{\color{hred}0.733} & \textbf{\color{hred}0.731} & \ours{} & \ours{} \\
Traffic & \underline{\color{hblue}0.345} & 0.310 & 0.517 & \underline{\color{hblue}0.301} & 0.518 & 0.548 & \textbf{\color{hred}0.242} & \textbf{\color{hred}0.145} & \ours{} & \ours{} \\
\bottomrule
\end{tabular}
\end{table*}

%% file: tables/tab_diversity.tex
\begin{table}[t]
\centering
\caption{\textbf{Spread of the candidates.} Spread is the mean distance between two candidates and headroom the MAE of the best candidate in hindsight, both over the MAE of the medoid.}
\label{tab:diversity}
\small
\setlength{\tabcolsep}{3pt}
\begin{tabular}{l|ccc}
\toprule
Statistic & FNSPID & FinTexTS & Time-MMD \\
\midrule
Spread & 1.320 & 1.382 & 0.545 \\
Headroom & 0.498 & 0.384 & 0.697 \\
\bottomrule
\end{tabular}
\end{table}

%% file: tables/tab_cand.tex
\begin{table}[t]
\centering
\caption{\textbf{Quality of the candidate sets.} Errors of a random candidate, the medoid, and the best candidate in hindsight among the candidates of each test row of both sets.}
\label{tab:cand}
\small
\setlength{\tabcolsep}{3pt}
\begin{tabular}{l|ccccc}
\toprule
Choice & MAE & Endpoint & Volatility & Range & FD \\
\midrule
\rowcolor{gray!15}\multicolumn{6}{l}{\textit{Predictor trained up to 2017}} \\
Random candidate & 1.070 & 1.084 & 0.473 & 0.340 & 1.130 \\
Medoid & 0.789 & 0.797 & 0.401 & 0.322 & 5.913 \\
Oracle & 0.393 & 0.385 & 0.307 & 0.273 & 0.648 \\
\midrule
\rowcolor{gray!15}\multicolumn{6}{l}{\textit{Predictor trained up to 2021}} \\
Random candidate & 1.053 & 1.062 & 0.534 & 0.404 & 2.335 \\
Medoid & 0.785 & 0.792 & 0.403 & 0.402 & 6.624 \\
Oracle & 0.401 & 0.397 & 0.325 & 0.324 & 0.799 \\
\bottomrule
\end{tabular}
\end{table}

%% file: tables/tab_size.tex
\begin{table}[t]
\centering
\caption{\textbf{Effect of the VLM size.} MAE-only selectors are refit on 2018--2022 with a new seed and tested on 2023. Endpoint and HL are the errors of the total change and of the upper and lower paths, and $^\ast$ marks a significant gain against the medoid.}
\label{tab:size}
\small
\setlength{\tabcolsep}{2pt}
\begin{tabular}{l|ccc}
\toprule
Selection & MAE & Endpoint & HL \\
\midrule
\rowcolor{gray!15}\multicolumn{4}{l}{\textit{Candidates from the predictor trained up to 2017}} \\
Medoid & 0.789 & 0.797 & 0.749 \\
\rowcolor{yellow!18}Qwen2.5-VL-3B & \textbf{\color{hred}0.783}$^\ast$ & \textbf{\color{hred}0.789}$^\ast$ & \textbf{\color{hred}0.744}$^\ast$ \\
Qwen3-VL-4B & 0.786 & 0.796 & 0.746 \\
Qwen3-VL-8B & \underline{\color{hblue}0.785} & \underline{\color{hblue}0.794} & \underline{\color{hblue}0.745} \\
\midrule
\rowcolor{gray!15}\multicolumn{4}{l}{\textit{Candidates from the predictor trained up to 2021}} \\
Medoid & 0.785 & \underline{\color{hblue}0.792} & 0.747 \\
\rowcolor{yellow!18}Qwen2.5-VL-3B & \textbf{\color{hred}0.781} & \textbf{\color{hred}0.788} & \textbf{\color{hred}0.743} \\
Qwen3-VL-4B & \underline{\color{hblue}0.784} & 0.793 & \underline{\color{hblue}0.746} \\
Qwen3-VL-8B & \underline{\color{hblue}0.784} & \underline{\color{hblue}0.792} & \underline{\color{hblue}0.746} \\
\bottomrule
\end{tabular}
\end{table}

%% file: tables/tab_size_shape.tex
\begin{table}[t]
\centering
\caption{\textbf{VLM size with the shape-aware target.} Change of the test errors of \ours{} in one default setting when the Qwen2.5-VL-3B selector is replaced by a larger one (\%).}
\label{tab:size_shape}
\small
\setlength{\tabcolsep}{4pt}
\begin{tabular}{l|cc|cc}
\toprule
\multirow{2.5}{*}{Selector} & \multicolumn{2}{c|}{FinTexTS} & \multicolumn{2}{c}{Time-MMD} \\
\cmidrule(lr){2-3} \cmidrule(lr){4-5}
 & MAE & Volatility & MAE & Volatility \\
\midrule
Qwen3-VL-4B & +0.1 & $-$0.2 & $-$0.4 & $-$0.1 \\
Qwen3-VL-8B & $-$0.1 & +1.1 & $-$0.6 & +0.3 \\
\bottomrule
\end{tabular}
\end{table}

%% file: tables/tab_ablation_detail.tex
% selected (paper_select.py)
\begin{table}[t]
\centering
\caption{\textbf{Rendering on validation.} Each row changes one setting of \ours{} in the last row, with $224{\times}112$ panels (Appendix~\ref{app:impl}), compared with the medoid on the same validation rows. Band draws all candidates behind the path. Marks follow Table~\ref{tab:rolling}.}
\label{tab:ablation_detail}
\small
\setlength{\tabcolsep}{1.3pt}
\begin{tabular}{l|cccccc}
\toprule
Setting & MAE & Endpoint & HL & Volatility & Range & FD \\
\midrule
\rowcolor{gray!15}\multicolumn{7}{l}{\textit{Rendering}} \\
Panel $336{\times}168$ & \underline{\color{hblue}0.854} & \textbf{\color{hred}0.814} & \textbf{\color{hred}0.807}$^\ast$ & \underline{\color{hblue}0.413}$^\ast$ & \underline{\color{hblue}0.363}$^\ast$ & \underline{\color{hblue}6.706} \\
Band & 0.858 & 0.818 & 0.813 & 0.423$^\ast$ & 0.366 & 7.111 \\
\midrule
\rowcolor{yellow!18}\ours{} & \textbf{\color{hred}0.852}$^\ast$ & \textbf{\color{hred}0.814} & \textbf{\color{hred}0.807}$^\ast$ & \textbf{\color{hred}0.405}$^\ast$ & \textbf{\color{hred}0.362}$^\ast$ & \textbf{\color{hred}6.591} \\
\bottomrule
\end{tabular}
\end{table}

%% file: tables/tab_tmmd_pick.tex
\begin{table}[t]
\centering
\caption{\textbf{Hyperparameters per Time-MMD dataset.} Settings that replace the default on each dataset, chosen on validation. All other settings are shared across all datasets.}
\label{tab:tmmd_pick}
\small
\begin{tabularx}{\columnwidth}{l|>{\raggedright\arraybackslash}X}
\toprule
Dataset & Used \\
\midrule
Agriculture & Default \\
Climate & Out-of-fold training candidates, prior path as a candidate \\
Economy & Half the candidates at a lower and half at a higher temperature \\
Energy & Predictor reads the benchmark prior, out-of-fold training candidates \\
Health & Predictor reads the benchmark prior, out-of-fold training candidates \\
Security & Predictor setting by cross-validation \\
Social Good & Predictor setting by cross-validation \\
Traffic & Predictor temperature 0.6 \\
\bottomrule
\end{tabularx}
\end{table}

%% file: tables/tab_onesetting.tex
\begin{table*}[t]
\centering
\caption{\textbf{Setting-agnostic performance.} Test MAE and volatility error of \ours{} with the per-dataset setting chosen on validation (Dataset-specific) and with one setting for all datasets (Dataset-common). Average is over all ten datasets. Red bold marks the lower value.}
\label{tab:onesetting}
\small
\setlength{\tabcolsep}{1.0pt}
\begin{tabular}{l|cc|cc|cc|cc|cc|cc|cc|cc||cc}
\toprule
\multirow{2.5}{*}{Setting} & \multicolumn{2}{c|}{Agriculture} & \multicolumn{2}{c|}{Climate} & \multicolumn{2}{c|}{Economy} & \multicolumn{2}{c|}{Energy} & \multicolumn{2}{c|}{Health} & \multicolumn{2}{c|}{Security} & \multicolumn{2}{c|}{Social Good} & \multicolumn{2}{c||}{Traffic} & \multicolumn{2}{c}{Average} \\
\cmidrule(lr){2-3} \cmidrule(lr){4-5} \cmidrule(lr){6-7} \cmidrule(lr){8-9} \cmidrule(lr){10-11} \cmidrule(lr){12-13} \cmidrule(lr){14-15} \cmidrule(lr){16-17} \cmidrule(lr){18-19}
 & MAE & Vol. & MAE & Vol. & MAE & Vol. & MAE & Vol. & MAE & Vol. & MAE & Vol. & MAE & Vol. & MAE & Vol. & MAE & Vol. \\
\midrule
Dataset-specific & 0.982 & 0.586 & \textbf{\color{hred}0.294} & 0.487 & \textbf{\color{hred}0.387} & 0.344 & \textbf{\color{hred}1.809} & \textbf{\color{hred}0.522} & 1.946 & 0.968 & \textbf{\color{hred}0.734} & \textbf{\color{hred}3.590} & 0.736 & 0.753 & \textbf{\color{hred}0.237} & \textbf{\color{hred}0.142} & 0.872 & 0.821 \\
Dataset-common & 0.982 & 0.586 & 0.339 & \textbf{\color{hred}0.408} & 0.397 & 0.344 & 1.847 & 0.564 & \textbf{\color{hred}1.785} & \textbf{\color{hred}0.898} & 0.784 & 3.652 & \textbf{\color{hred}0.733} & \textbf{\color{hred}0.731} & 0.242 & 0.145 & \textbf{\color{hred}0.871} & \textbf{\color{hred}0.815} \\
\bottomrule
\end{tabular}
\end{table*}

%% file: tables/tab_relerr_cfg.tex
\begin{table}[t]
\centering
\caption{\textbf{Dataset-specific vs.\ dataset-common setting.} Scale-free averages of Table~\ref{tab:relerr}. Each baseline row takes the best value in its group for each column.}
\label{tab:relerr_cfg}
\small
\setlength{\tabcolsep}{3pt}
\begin{tabular}{l|cc|cc}
\toprule
\multirow{2.5}{*}{Method} & \multicolumn{2}{c|}{Arithmetic mean} & \multicolumn{2}{c}{Geometric mean} \\
\cmidrule(lr){2-3} \cmidrule(lr){4-5}
 & MAE & Volatility & MAE & Volatility \\
\midrule
\rowcolor{gray!15}\multicolumn{5}{l}{\textit{Best baseline of each group}} \\
Best TSFM & 0.912 & 0.698 & \underline{\color{hblue}0.883} & 0.609 \\
Best task-specific model & 0.915 & 0.597 & 0.891 & 0.556 \\
Best multimodal model & 0.924 & 0.627 & 0.890 & 0.558 \\
\midrule
\rowcolor{gray!15}\multicolumn{5}{l}{\textit{Ours}} \\
\rowcolor{yellow!18}\ours{} (Dataset-specific) & \textbf{\color{hred}0.892} & \underline{\color{hblue}0.518} & \textbf{\color{hred}0.867} & \underline{\color{hblue}0.470} \\
\rowcolor{yellow!18}\ours{} (Dataset-common) & \underline{\color{hblue}0.906} & \textbf{\color{hred}0.510} & 0.884 & \textbf{\color{hred}0.462} \\
\bottomrule
\end{tabular}
\end{table}

%% file: tables/tab_eff_infer.tex
\begin{table}[t]
\centering
\caption{\textbf{Inference cost per forecast.} Time on one L40S GPU and peak GPU memory on the FNSPID test rows.}
\label{tab:eff_infer}
\small
\setlength{\tabcolsep}{4pt}
\begin{tabular}{l|cccc}
\toprule
Model & Params & Paths & Time (ms) & Memory (GB) \\
\midrule
Chronos-T5 & 201M & 100 & 45.5 & 6.3 \\
Chronos-Bolt & 205M & 256 & 0.18 & 0.9 \\
Chronos-2 & 120M & 256 & 1.10 & 0.5 \\
TimesFM-2.5 & 231M & 256 & 639 & 1.7 \\
Moirai-1.1 & 91.4M & 100 & 12.1 & 1.0 \\
Moirai-2.0 & 11.4M & 256 & 0.70 & 0.1 \\
Toto & 151M & 256 & 31.1 & 11.4 \\
ViTime & 74.2M & 100 & 68.8 & 2.9 \\
VisionTS & 112M & 1 & 2.61 & 3.0 \\
\midrule
ChatTime$^\dagger$ & 6.7B & 1 & 1175 & -- \\
\midrule
\rowcolor{yellow!18}\ours{} predictor & 10.9M & 32 & 573 & 1.8 \\
\rowcolor{yellow!18}\ours{} selector & 3.8B & 1 & 829 & 8.9 \\
\rowcolor{yellow!18}\ours{} total & 3.8B & 1 & 1403 & 8.9 \\
\bottomrule
\end{tabular}
\end{table}

%% file: tables/tab_shapes.tex
% Input shapes of the selector (user 2026-11-20). Measured with the Qwen2.5-VL-3B processor of exp/vsel.py
% (--layout grid --img 224x112, 4 columns); news length over 400 sampled FNSPID rows with --news_clean 1.
\begin{table}[t]
\centering
\caption{\textbf{Tensor shapes of the selector.} One FNSPID row with $B$ rows per batch and $T$ tokens per row.}
\label{tab:shapes}
\small
\setlength{\tabcolsep}{3pt}
\begin{tabular}{l|cc}
\toprule
Tensor & Training ($K{=}16$) & Evaluation ($K{=}32$) \\
\midrule
\rowcolor{gray!15}\multicolumn{3}{l}{\textit{Image}} \\
Panel (pixels) & $224{\times}112$ & $224{\times}112$ \\
Grid image (pixels) & $896{\times}448$ & $896{\times}896$ \\
Patch grid $(t,h,w)$ & $(1, 32, 64)$ & $(1, 64, 64)$ \\
Pixel values & $[2048B, 1176]$ & $[4096B, 1176]$ \\
Image tokens & 512 & 1{,}024 \\
\midrule
\rowcolor{gray!15}\multicolumn{3}{l}{\textit{Text}} \\
Template and markers & 177 & 273 \\
News & 0--450 & 0--450 \\
Sequence length $T$ & 690--1{,}140 & 1{,}300--1{,}750 \\
\midrule
\rowcolor{gray!15}\multicolumn{3}{l}{\textit{Selector}} \\
Hidden states & $[B, T, 2048]$ & $[B, T, 2048]$ \\
Candidate states & $[B, 16, 2048]$ & $[B, 32, 2048]$ \\
Head input & $[B, 17, 512]$ & $[B, 33, 512]$ \\
Scores & $[B, 16]$ & $[B, 32]$ \\
Auxiliary outputs & $[B, 16, 3]$ & $[B, 32, 3]$ \\
\bottomrule
\end{tabular}
\end{table}

%% file: tables/tab_pred_arch.tex
% Predictor architectures (user 2026-11-20). Sources: exp/build_tmmd.py, exp/build_fintexts.py (Flow MLP; per-dataset
% configs in regen_temp.SRC), Section C.3 (KiT on FNSPID), exp/imgpred.py (flow UNet on Time-MMD).
\begin{table*}[t]
\centering
\caption{\textbf{Predictor architectures.} TS-to-Chart serves FinTexTS and Time-MMD, and Chart-Native serves FNSPID.}
\label{tab:pred_arch}
\small
\setlength{\tabcolsep}{5pt}
\begin{tabular}{l|l|l|l}
\toprule
 & \textbf{TS-to-Chart} & \textbf{Chart-Native (FNSPID)} & \textbf{Chart-Native (Time-MMD)} \\
\midrule
Network & Flow-matching MLP & KiT diffusion transformer & Flow-matching UNet \\
Input & Normalized history & Candle chart of 90 days & Line chart of the history \\
Output & Future values & Value bins of 10 future days & Future columns of the chart \\
Size & 3 layers, width 256--512 & 6 layers, width 256 & 4 levels, 32--128 channels \\
Sampling & 32 Euler steps & 32 flow steps & 32 Euler steps \\
Training & Batch 256, up to 20k steps & Batch 128, 30k steps & Batch 128, early stop on val. \\
Pretraining & Stock prices (six Time-MMD) & None & None \\
\bottomrule
\end{tabular}
\end{table*}

%% file: tables/tab_design_pick.tex
% generated by exp/tab_design_pick.py
\begin{table}[t]
\centering
\caption{\textbf{Choice of the predictor design.} Validation error of the selection target with the candidates of each trained design.}
\label{tab:design_pick}
\small
\setlength{\tabcolsep}{2.5pt}
\begin{tabular}{l|cc|l}
\toprule
Dataset & TS-to-Chart & Chart-Native & Chosen \\
\midrule
FNSPID & 1.022 & \textbf{\color{hred}0.940} & Chart-Native \\
FinTexTS & \textbf{\color{hred}0.914} & 0.944 & TS-to-Chart \\
\midrule
\rowcolor{gray!15}\multicolumn{4}{l}{\textit{Time-MMD}} \\
Agriculture & \textbf{\color{hred}0.568} & 0.634 & TS-to-Chart \\
Climate & \textbf{\color{hred}0.423} & 0.434 & TS-to-Chart \\
Economy & \textbf{\color{hred}0.264} & 0.335 & TS-to-Chart \\
Energy & \textbf{\color{hred}1.277} & 1.450 & TS-to-Chart \\
Health & \textbf{\color{hred}1.641} & 1.704 & TS-to-Chart \\
Security & \textbf{\color{hred}0.369} & 0.452 & TS-to-Chart \\
Social Good & \textbf{\color{hred}0.592} & 0.637 & TS-to-Chart \\
Traffic & \textbf{\color{hred}0.097} & 0.123 & TS-to-Chart \\
\bottomrule
\end{tabular}
\end{table}

%% file: tables/tab_train_k.tex
\begin{table}[t]
\centering
\caption{\textbf{Number of candidates in training and evaluation.} MAE on the FNSPID test rows when the selector trained with $K$ candidates per row picks from the first $K$ paths (columns).}
\label{tab:train_k}
\small
\setlength{\tabcolsep}{3pt}
\begin{tabular}{l|ccccc}
\toprule
\multirow{2.5}{*}{Selection} & \multicolumn{5}{c}{$K$ in evaluation} \\
\cmidrule(lr){2-6}
 & 4 & 8 & 16 & 24 & 32 \\
\midrule
Random pick & 1.063 & 1.065 & 1.072 & 1.065 & 1.064 \\
\midrule
Trained with $K{=}8$ & \underline{\color{hblue}0.842} & \underline{\color{hblue}0.811} & 0.796 & 0.792 & 0.790 \\
Trained with $K{=}16$ & \textbf{\color{hred}0.841} & \textbf{\color{hred}0.810} & \textbf{\color{hred}0.795} & \underline{\color{hblue}0.790} & \underline{\color{hblue}0.787} \\
Trained with $K{=}32$ & 0.842 & 0.811 & \underline{\color{hblue}0.795} & \textbf{\color{hred}0.790} & \textbf{\color{hred}0.786} \\
\bottomrule
\end{tabular}
\end{table}

%% file: tables/tab_simscore.tex
\begin{table}[t]
\centering
\caption{\textbf{Correlation of distance and weight gap.} Mean rank correlation between the two over the candidate pairs of each test row, with the learned or shuffled scores. Positive rows is the share of rows with a positive correlation of the learned scores.}
\label{tab:simscore}
\small
\setlength{\tabcolsep}{3pt}
\begin{tabular}{l|cc|c}
\toprule
\multirow{2.5}{*}{Dataset} & \multicolumn{2}{c|}{Rank correlation} & \multirow{2.5}{*}{\shortstack{Positive\\rows (\%)}} \\
\cmidrule(lr){2-3}
 & Learned scores & Shuffled scores & \\
\midrule
FNSPID & \textbf{\color{hred}0.62} & 0.00 & 100 \\
FinTexTS & \textbf{\color{hred}0.65} & 0.00 & 100 \\
\midrule
\rowcolor{gray!15}\multicolumn{4}{l}{\textit{Time-MMD}} \\
Agriculture & \textbf{\color{hred}0.61} & 0.00 & 100 \\
Climate & \textbf{\color{hred}0.30} & $-$0.01 & 100 \\
Economy & \textbf{\color{hred}0.62} & $-$0.01 & 100 \\
Energy & \textbf{\color{hred}0.62} & 0.00 & 100 \\
Health & \textbf{\color{hred}0.40} & $-$0.01 & 94 \\
Security & \textbf{\color{hred}0.40} & $-$0.01 & 100 \\
Social Good & \textbf{\color{hred}0.45} & 0.01 & 100 \\
Traffic & \textbf{\color{hred}0.41} & $-$0.02 & 100 \\
\bottomrule
\end{tabular}
\end{table}